%% file: main.tex
\documentclass[journal]{IEEEtran} 

\IEEEoverridecommandlockouts                             

\usepackage{CJK}
\usepackage{pifont}
\usepackage{graphicx}
\usepackage{multicol}
\usepackage{booktabs}
\usepackage{amsmath,amssymb}
\usepackage{amsthm}
\usepackage[utf8]{inputenc}
\usepackage[english]{babel}
\usepackage{multirow}
\usepackage[font=small]{caption}
\usepackage{float}
\usepackage[hidelinks]{hyperref}

\usepackage{lettrine}
\usepackage[]{algorithm2e}
\usepackage{algpseudocode}
\usepackage{cite}
\usepackage{filecontents}
\usepackage{lipsum}
\usepackage{color}
\usepackage{esdiff}
\usepackage{epstopdf}
\usepackage[normalem]{ulem}
\usepackage{soul}

\newcommand{\tabincell}[2]{\begin{tabular}{@{}#1@{}}#2\end{tabular}}
\usepackage{subcaption}
\usepackage{xcolor}
\usepackage{colortbl}
\usepackage{balance}
\definecolor{pink}{rgb}{1, 0, 1}
\definecolor{orange}{rgb}{1, 0.7529, 0}
\definecolor{darkgreen}{rgb}{0, 0.8, 0}
\begin{document}
\bstctlcite{IEEEexample:BSTcontrol}

\title{Coverage Path Planning: Classical Foundations, Recent Advances, and Future Directions}

\author{
{Zongyuan Shen$^1$}, {Shalabh Gupta$^2$}, {Shancheng Zhao$^1$}, {Dehua Zhou$^1$}, {Gao Wang$^1$}, {Zhongqiang Ren$^3$}, \\{Yaming Ou$^4$}, {Yikui Zhai$^5$}, and {C. L. Philip Chen$^6$, \textit{Life Fellow, IEEE}}
\vspace{-9pt}

\thanks {$^1$College of Information Science and Technology, Jinan University, Guangzhou 510632, China.}
\thanks {$^2$Department of Electrical and Computer Engineering, University of Connecticut, Storrs, CT 06269, USA.}
\thanks {$^3$Global College, Shanghai Jiao Tong University, Shanghai 200240, China.}
\thanks {$^4$School of Artificial Intelligence, University of Chinese Academy of Sciences, Beijing 100049, China.}
\thanks {$^5$School of Electronics and Information Engineering, Wuyi University, Jiangmen 529000, China.}
\thanks {$^6$School of Computer Science and Engineering, South China University of Technology, Guangzhou 510006, China.}
}
        
\maketitle
\thispagestyle{empty}

\input{sections/abstract.tex}
\input{sections/introduction.tex}
\input{sections/review.tex}
\input{sections/conclusions.tex}

\balance
\bibliographystyle{IEEEtran}
\bibliography{reference}

\end{document}

%% file: sections/abstract.tex
\begin{abstract}
Coverage path planning (CPP) is a fundamental problem in robot motion planning, whose aim is to produce robot trajectories that provide complete coverage of target workspaces while minimizing task-specific objectives such as path length, overlap, number of turns, and energy consumption. CPP has widespread applications in cleaning, inspection, mapping, agriculture, manufacturing, surveillance, demining, and environmental monitoring. Although classical CPP has been extensively studied, recent advances have extended CPP beyond single-robot settings to multi-robot systems, complex 3D environments, constrained platforms, learning-based coverage planning, and visual coverage tasks. This paper presents a comprehensive survey of 125 representative works published primarily between 2015 and 2026, while presenting the evolution of recent developments in light of the classical CPP methods published before 2015. The CPP methods are organized into six main categories: single-robot CPP, multi-robot CPP, 3D CPP, constrained CPP, learning-based CPP, and visual CPP. For each category, the review summarizes the main planning formulations, representative algorithms, strengths, and limitations. In addition, the review analyzes how environmental knowledge, workspace geometry, robot constraints, sensing objectives, and coordination requirements shape the CPP problem. The survey further discusses open challenges in scalable online planning, multi-robot coordination, 3D and visual coverage, unified platform-constrained and resource-aware coverage, and learning-enhanced coverage. Thus, the survey provides a structured overview of recent CPP developments and future research directions.
\end{abstract}

\vspace{-6pt}

%% file: sections/introduction.tex
\section{Introduction}

\begin{figure}[t]
\vspace{2pt}
    \centering
    \subfloat[Floor cleaning~\cite{ramesh2024}]{
    \includegraphics[width=0.23\textwidth]{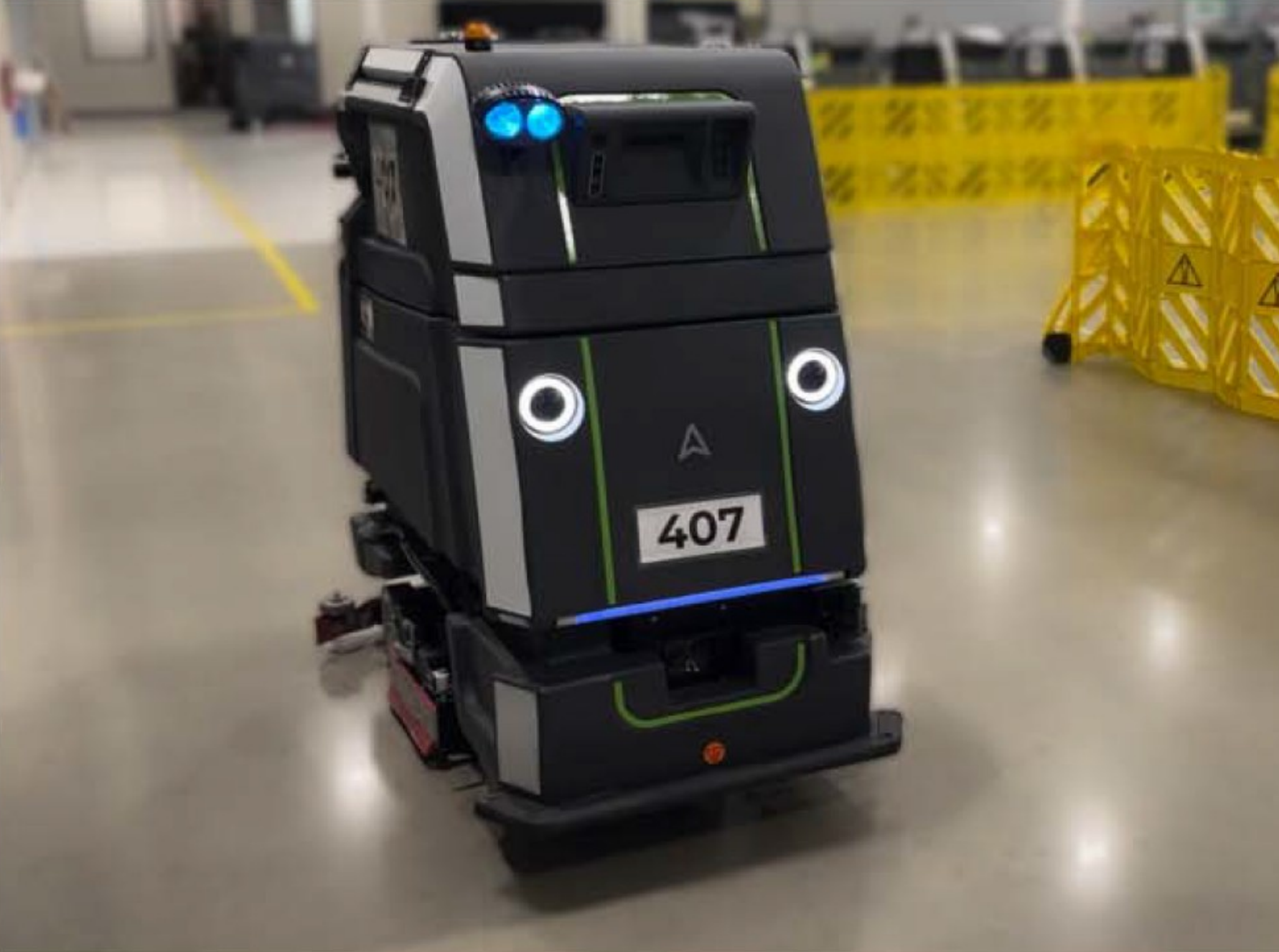}\label{fig:example_part1}}\hspace{-10pt}\quad
    \centering
    \subfloat[Underwater terrain mapping~\cite{yordanova2020coverage}]{
    \includegraphics[width=0.23\textwidth]{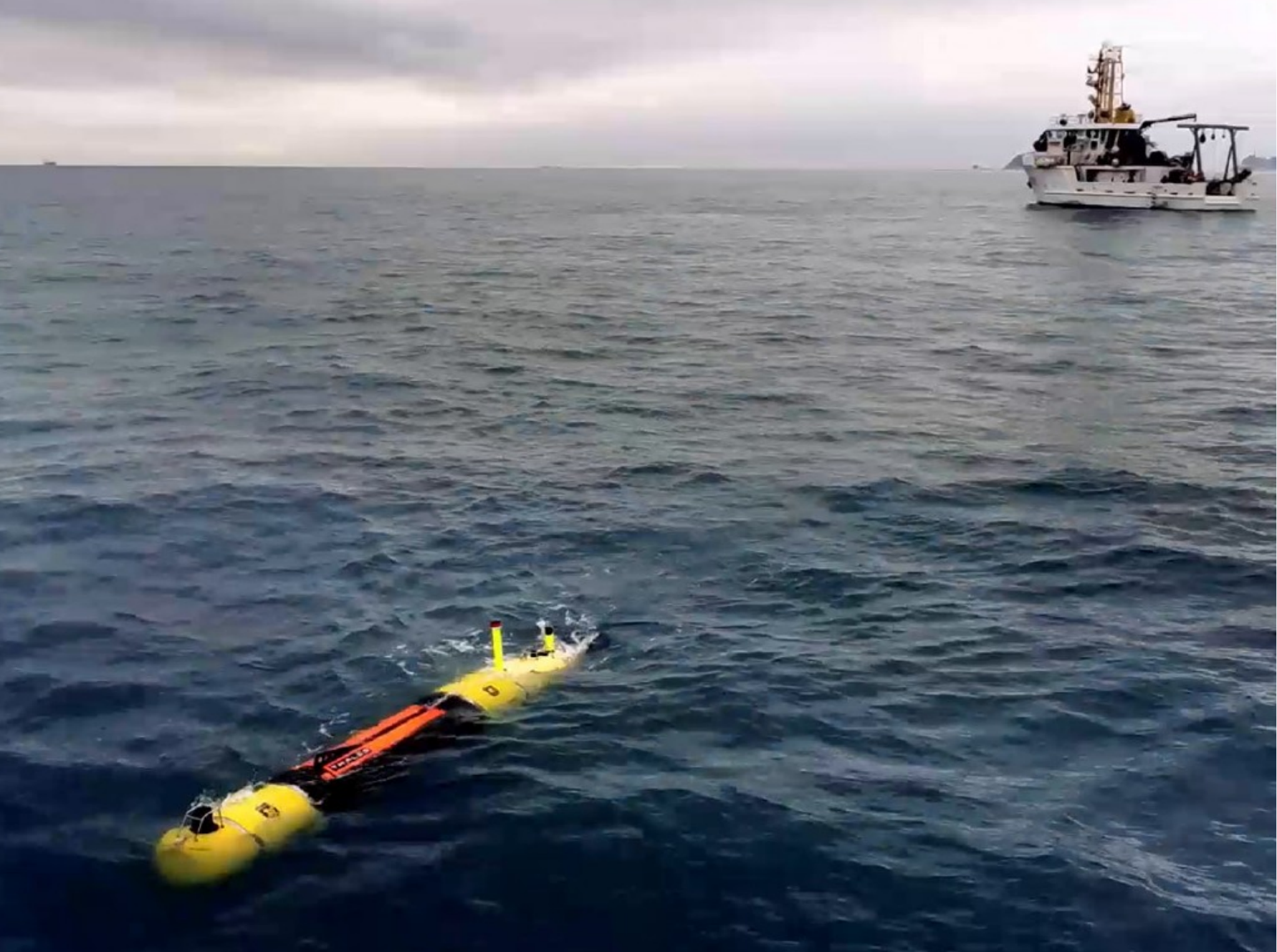}\label{fig:example_part2}}\vspace{0.5em}\\
    \centering
    \subfloat[Structural inspection~\cite{feng2024fc}]{
    \includegraphics[width=0.23\textwidth]{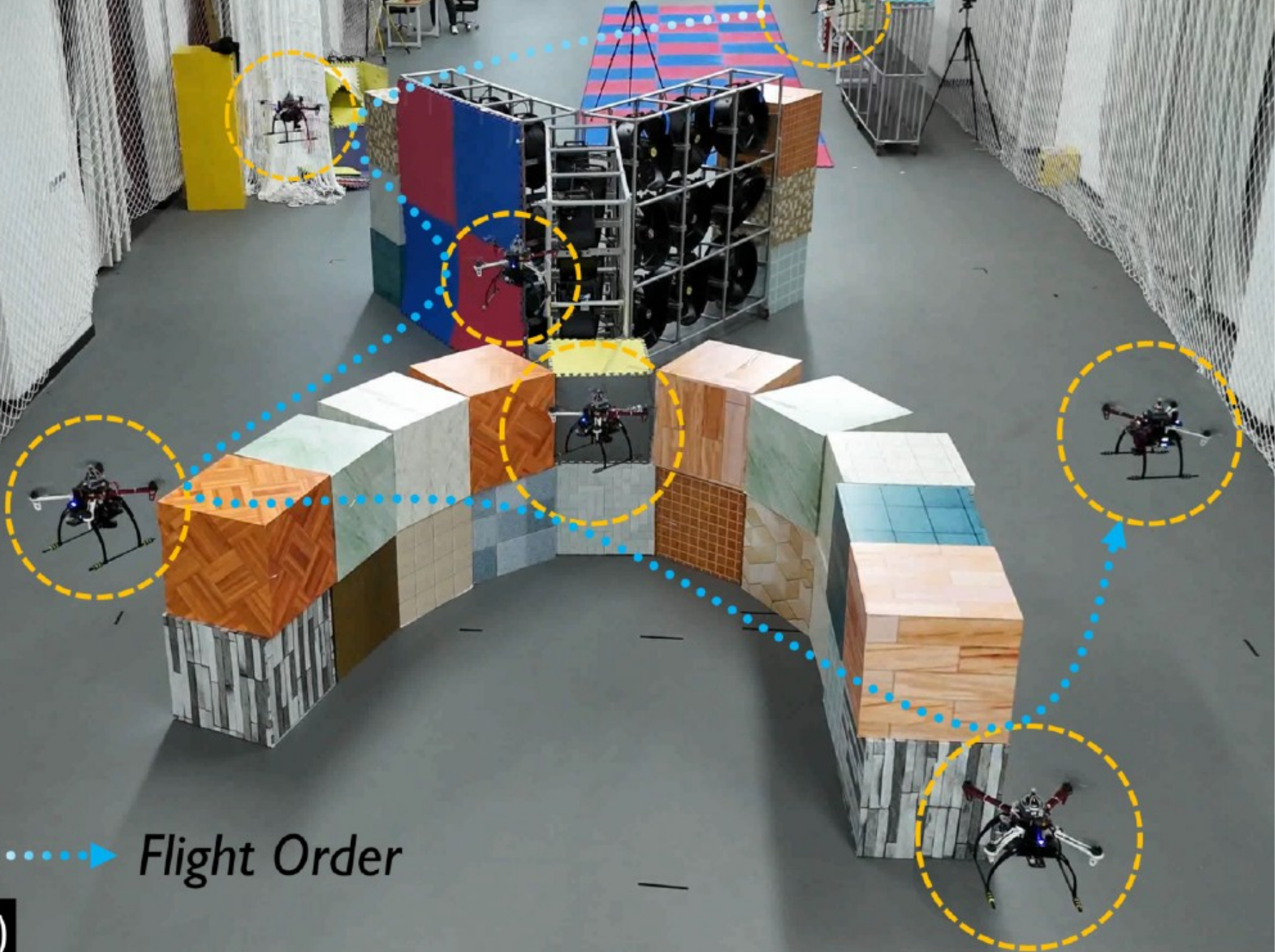}\label{fig:example_part3}}\hspace{-10pt}\quad
    \centering
    \subfloat[Surface cleaning~\cite{yang2020cellular}]{
    \includegraphics[width=0.23\textwidth]{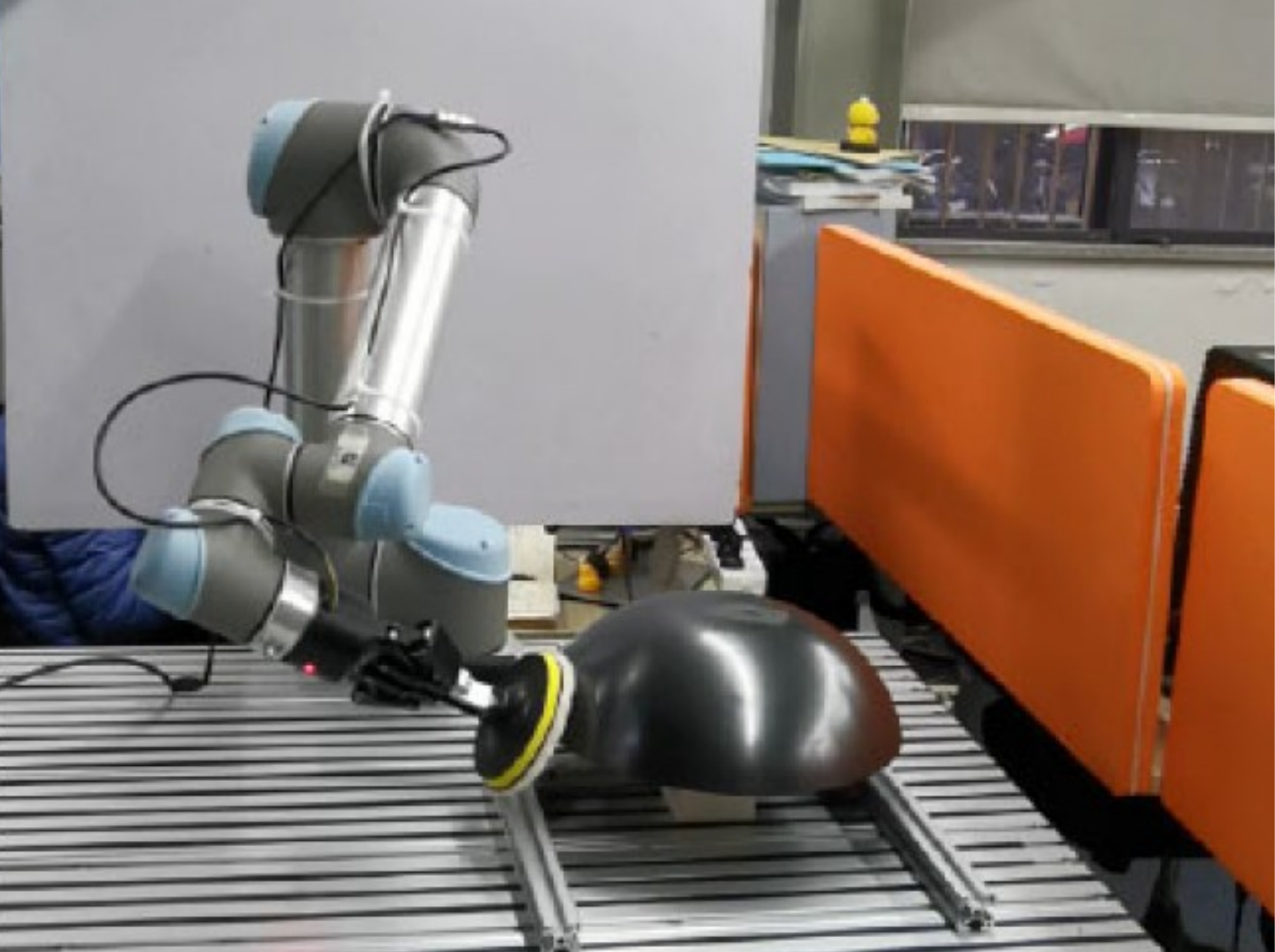}\label{fig:example_part4}}\\
          \caption{Application examples of CPP.}\label{fig:example}
          \vspace{-1em}
\end{figure}

Coverage path planning (CPP) is a fundamental problem in robot motion planning, whose aim is to produce robot trajectories that provide complete coverage of target workspaces while minimizing task-specific objectives such as path length, overlap, number of turns, and energy consumption. CPP plays an important role in a broad range of robotic applications, including household services (e.g., floor cleaning~\cite{ramesh2024} and lawn mowing~\cite{Song_sose2015}); environmental monitoring (e.g., terrain mapping~\cite{palomeras2018autonomous, yordanova2020coverage,Ou2025,shen2022ct}); industrial operations (e.g., structural quality inspection~\cite{feng2024fc, vidal2017}, surface cleaning~\cite{yang2020cellular}, crack filling~\cite{yu2019complete,veeraraghavan2024complete}, and spray-painting~\cite{vempati2018paintcopter,atkar2005}); agricultural tasks (e.g., weeding~\cite{maini2022online} and farming~\cite{jin2011coverage}); and hazardous operations (e.g., offshore oil spill cleaning~\cite{SGH13} and mine counter measures~\cite{mukherjee2011symbolic}). Fig.~\ref{fig:example} shows application examples of CPP using different robotic platforms, such as unmanned  ground vehicles (UGVs), unmanned underwater vehicles (UUVs), unmanned aerial vehicles (UAVs), and industrial manipulators.

\begin{figure*}[t]
        \centering        
        \includegraphics[width=0.86\textwidth]{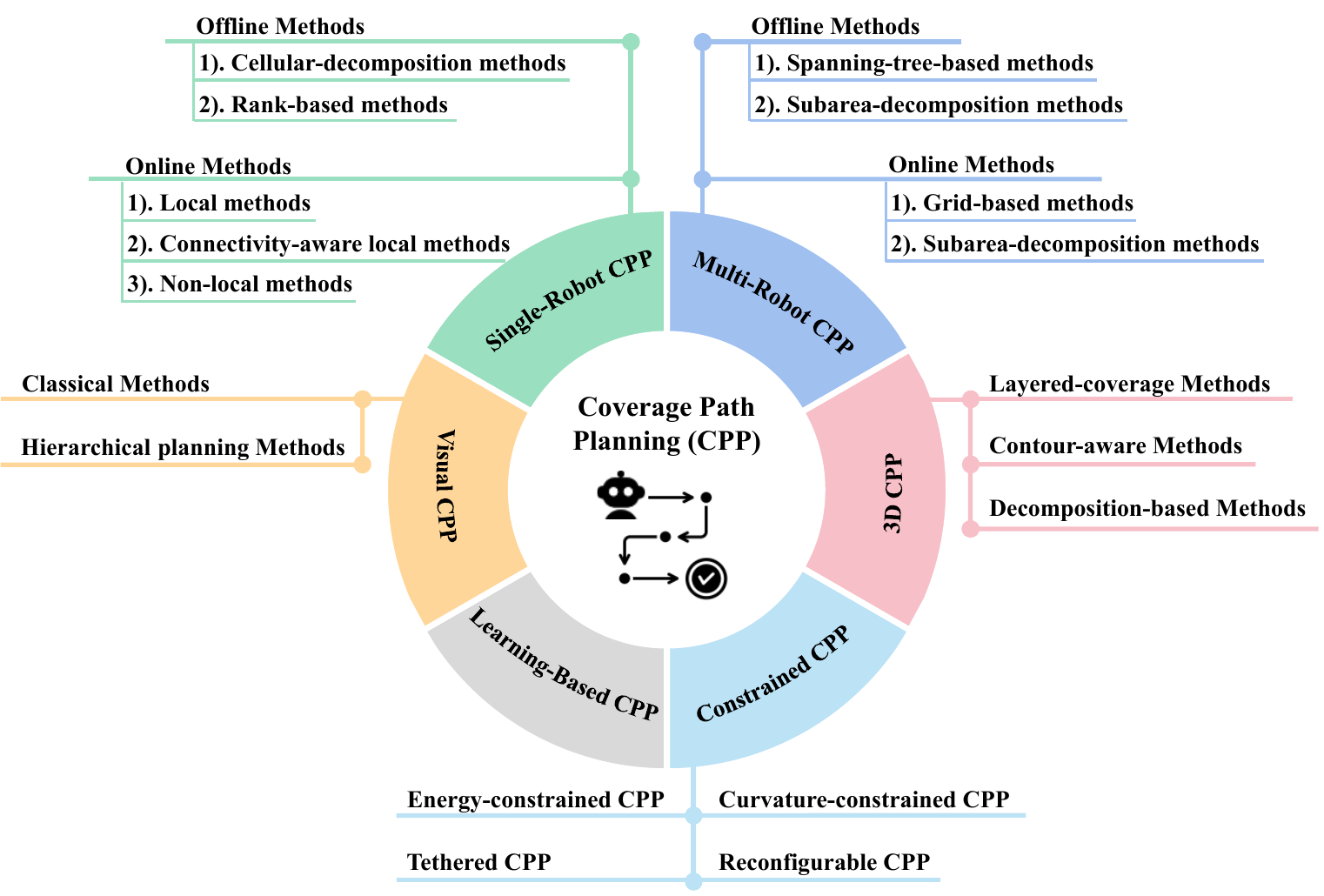}
    \caption{Taxonomy of CPP methods.}\label{fig:cppoverview} 
    \vspace{-1em}
 \end{figure*}

\subsection{CPP Classifications}
The CPP methods are classified in different ways, as shown in Fig.~\ref{fig:cppoverview}. The traditional CPP problem consists of a single robot performing coverage of 2D workspaces (e.g., floor cleaning in a warehouse). In recent years, CPP has evolved far beyond the classical single-robot 2D coverage to multi-robot CPP, 3D CPP, constrained CPP, learning-based CPP and visual CPP. Multi-robot CPP deals with problems such as task allocation, workload balancing, inter-robot conflict resolution, failure resilience, communication limitations, overlap reduction, and coordinated operation. Several mechanisms have been used for single-robot and multi-robot coverage, such as cellular decomposition, grid-based search, spanning-tree coverage, potential-field guidance, and graph-based sequencing.

The 3D CPP extension requires coverage of complex terrains and structures (e.g., high-rise buildings in aerial environments and uneven hilly surfaces in underwater environments). In these problems, surface reconstruction, elevation variation, and energy costs become critical. The constrained CPP problems require the planner to account for different platform-specific constraints, such as  battery capacities, bounded curvatures, tethered cable limits, and morphological restrictions. Recently, learning-based CPP has emerged as a novel methodology, 
where learning techniques are integrated into classical CPP frameworks to support specific planning components. Finally, visual CPP shifts the focus from physical area coverage to trajectory generation under viewpoint constraints for inspection or reconstruction of target surfaces.

Furthermore, all CPP methods are categorized as either offline or online, depending on the availability of environmental information. While offline methods assume the environment is known a priori to make plans before deployment, online methods operate in unknown or partially known environments and generate coverage trajectories incrementally based on real-time sensor data. Online CPP is challenging because it requires the robot to not only explore the unknown environment and cover the target regions but also avoid obstacles, escape dead ends, prevent coverage holes~\cite{shen2026}, and minimize overlaps.

These developments have significantly broadened the scope of CPP and created the need for a systematic review that connects classical methods with emerging problems.

\subsection{Existing Surveys}
Table~\ref{tab:surveycompare} summarizes the existing survey papers that have reviewed CPP with different scopes and emphases. Early surveys discussed classical methods that have served as the foundation for the development of advanced methods. The survey by Choset~\cite{choset2001coverage} focused mainly on single-robot CPP in 2D environments and organized representative methods according to the type of workspace decomposition, while also discussing the early multi-robot extensions. Based on Choset's taxonomy, Galceran and Carreras~\cite{galceran2013survey} systematically reviewed single-robot CPP methods up to 2013 and further discussed several representative 3D CPP and multi-robot CPP methods. 

More recent surveys have focused on specific application domains or problem settings. Bormann et al.~\cite{bormann2018indoor} implemented and quantitatively compared six offline CPP algorithms for indoor room-level coverage. Cabreira et al.~\cite{cabreira2019survey} reviewed UAV-related CPP methods considering specific factors during planning such as camera footprint, flight time, turning maneuvers, wind effects, and energy consumption. H{\"o}ffmann et al.~\cite{hoffmann2024optimal} provided an in-depth review of 2D offline CPP methods that are applicable to precision agriculture. 

 \begin{table*}[t]
\centering
\caption{Comparison of representative CPP survey papers.}\vspace{-3pt}
\label{tab:surveycompare}
\centering
\setlength\tabcolsep{3.5pt}
\begin{tabular}{c c l c c c c c c}
\toprule
\specialrule{0.1em}{1pt}{1pt} 
\tabincell{l}{\textbf{Paper}} 
&\tabincell{l}{\textbf{Year}} 
&\tabincell{l}{\textbf{Main Scope}} 
&\tabincell{l}{\textbf{Single-Robot CPP}} 
&\tabincell{l}{\textbf{Multi-Robot CPP}} 
&\tabincell{l}{\textbf{3D CPP}} 
&\tabincell{l}{\textbf{Constrained CPP}}  
&\tabincell{l}{\textbf{Visual CPP}}
&\tabincell{l}{\textbf{Learning-Based CPP}}\\
\toprule

\tabincell{l}{{\cite{choset2001coverage}}} 
&\tabincell{l}{2001} 
&\tabincell{l}{{2D CPP for}\\{mobile robots}} 
&\tabincell{c}{Full} 
&\tabincell{c}{Partial} 
&\tabincell{c}{None} 
&\tabincell{c}{None} 
&\tabincell{c}{None} 
&\tabincell{c}{None}\\

\specialrule{0em}{2pt}{2pt}
\tabincell{l}{\cite{galceran2013survey}} 
&\tabincell{l}{2013} 
&\tabincell{l}{{General CPP}\\{methods up to 2013}}
&\tabincell{l}{Full} 
&\tabincell{l}{Partial} 
&\tabincell{l}{Partial} 
&\tabincell{l}{None} 
&\tabincell{l}{Limited} 
&\tabincell{c}{None}\\

\specialrule{0em}{2pt}{2pt}
\tabincell{l}{{\cite{bormann2018indoor}}} 
&\tabincell{l}{2018} 
&\tabincell{l}{{Evaluation of six}\\{offline CPP methods}}
&\tabincell{l}{Limited} 
&\tabincell{l}{None} 
&\tabincell{l}{None} 
&\tabincell{l}{None} 
&\tabincell{l}{None} 
&\tabincell{c}{None}\\

\specialrule{0em}{2pt}{2pt}
\tabincell{l}{{\cite{cabreira2019survey}}} 
&\tabincell{l}{2019} 
&\tabincell{l}{{UAV-related CPP methods}} 
&\tabincell{l}{Partial} 
&\tabincell{l}{Partial} 
&\tabincell{l}{None} 
&\tabincell{l}{Partial} 
&\tabincell{l}{None} 
&\tabincell{c}{None}\\

\specialrule{0em}{2pt}{2pt}
\tabincell{l}{{\cite{hoffmann2024optimal}}} 
&\tabincell{l}{2024} 
&\tabincell{l}{{2D offline CPP}\\{for agriculture}} 
&\tabincell{l}{Partial} 
&\tabincell{l}{None} 
&\tabincell{l}{None} 
&\tabincell{l}{Limited} 
&\tabincell{l}{None} 
&\tabincell{c}{None}\\

\specialrule{0em}{2pt}{2pt}
\tabincell{l}{{Ours}} 
&\tabincell{l}{2026} 
&\tabincell{l}{Unified CPP survey from\\{classical to recent methods}}
&\tabincell{l}{Full} 
&\tabincell{l}{Full} 
&\tabincell{l}{Full} 
&\tabincell{l}{Full} 
&\tabincell{l}{Full} 
&\tabincell{c}{Full}\\
\bottomrule
\specialrule{0.1em}{1pt}{1pt} 
\end{tabular}

\vspace{2pt}
\begin{minipage}{\textwidth}
\footnotesize
For any category, ``Full''  indicates it is systematically reviewed with representative methods across a broad CPP context; ``Partial'' indicates it is covered within a narrow application- or problem-specific context; ``Limited'' indicates it is only briefly mentioned; and ``None'' indicates it is not reviewed.
\end{minipage}\vspace{-3pt}
\end{table*}

\subsection{Contributions}
Although existing works provide valuable insights, a unified and up-to-date survey that connects classical CPP with recent developments is still lacking. This survey addresses this gap by presenting a comprehensive and structured review of 125 representative works, primarily published between 2015 and 2026, while linking recent advances to classical CPP foundations. Rather than organizing the literature only by algorithmic families, we adopt a problem-driven taxonomy that captures the major directions along which CPP has evolved. Accordingly, CPP methods are organized into six main categories: single-robot CPP, multi-robot CPP, 3D CPP, constrained CPP, learning-based CPP and visual CPP. Among recent advances, the learning-based CPP is presented as a promising direction, where supervised learning and reinforcement learning are integrated with classical CPP components for path generation, subarea sequencing, multi-robot task allocation, and coverage coordination under robot-specific constraints. For each category, we summarize the main problem formulations, representative algorithms, key ideas, and strengths and limitations. Fig.~\ref{fig:cppoverview} shows these categories and the methodologies therein. 

\subsection{Organization}
The remainder of this paper is organized as follows. Section~\ref{sec:singlerobot} reviews single-robot CPP methods, including online and offline approaches. Section~\ref{sec:multirobot} discusses multi-robot CPP. Section~\ref{sec:threeDCPP} reviews 3D CPP methods according to layered coverage, contour-aware planning, and surface decomposition. Section~\ref{sec:robotconstrainedcpp} discusses constrained CPP, including energy-constrained CPP, curvature-constrained CPP, tethered CPP, and reconfigurable CPP. Section~\ref{sec:learningcpp} reviews learning-based CPP methods. Section~\ref{sec:visualcpp} summarizes visual CPP methods from the viewpoint of coverage planning. Finally, Section~\ref{sec:conclusions} concludes the survey and discusses future research directions.

%% file: sections/review.tex
\section{Single-Robot CPP} \label{sec:singlerobot}

Single-robot CPP forms the foundational basis of all CPP literature. Advanced CPP formulations (e.g., multi-robot CPP, 3D CPP, and constrained CPP) are extensions of the single-robot CPP under different requirements, workspace geometries, sensing objectives, and constraints (e.g., curvature and energy). The single-robot CPP methods can be broadly classified as offline or online, depending on the availability of environmental information. Offline methods assume prior knowledge of the environment and plan coverage paths before deployment. In contrast, online methods generate coverage trajectories incrementally based on sensor data collected during navigation, thus they are suitable for unknown environments.

\subsection{Offline Methods}
\label{sec:single_offline_related_work}

Typically, offline CPP problems can be solved using the traveling salesman problem (TSP) since the environment is known; however, TSP is NP-hard~\cite{galceran2013survey}. Thus, the search for an optimal solution becomes computationally intractable as the problem complexity increases. In this regard, the offline methods exploit prior map information to partition the workspace into obstacle-free subareas to reduce the complexity of coverage path generation. These subareas are simpler coverage units, such as cells or ranks. While a cell is a free region that can be covered using a simple coverage pattern (e.g., back and forth motion), a rank is a stripe-shaped cell that can be covered by a single straight line. Then, the coverage planner decides the subarea traversal order and the local coverage path within each subarea. These methods are classified into cellular decomposition and rank-based methods.

\begin{figure}[t]
        \centering        \includegraphics[width=0.45\textwidth]{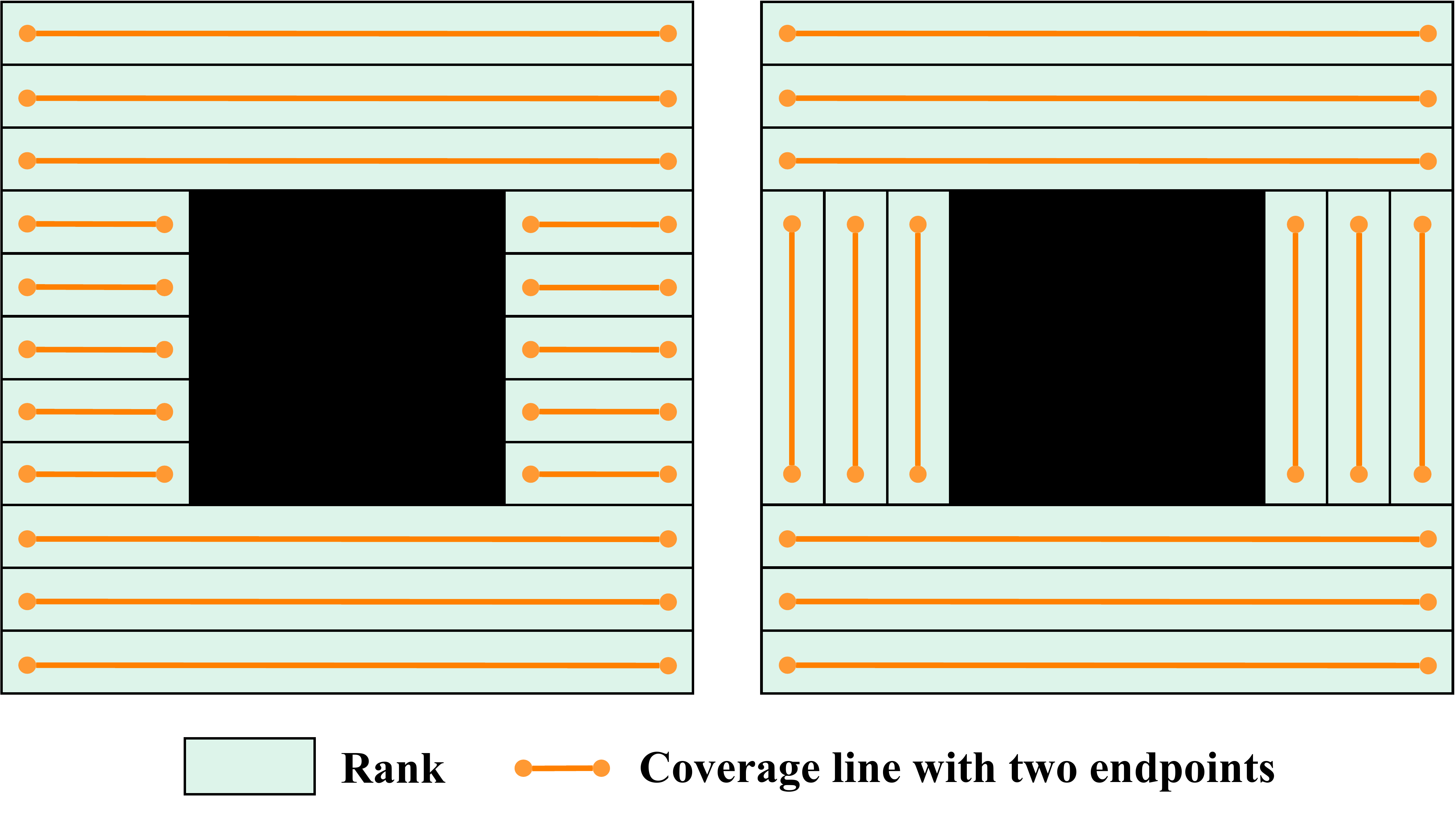}
    \caption{Illustration of the subarea-decomposition method in~\cite{ramesh2022optimal}. The free space is decomposed into ranks, each of which can be covered by a single straight line. The right decomposition uses fewer ranks than the left one, leading to less number of turns.}\label{fig:ramesh2022optimal} 
    \vspace{-1em}
 \end{figure}

 \begin{figure*}[t]
        \centering        \includegraphics[width=0.96\textwidth]{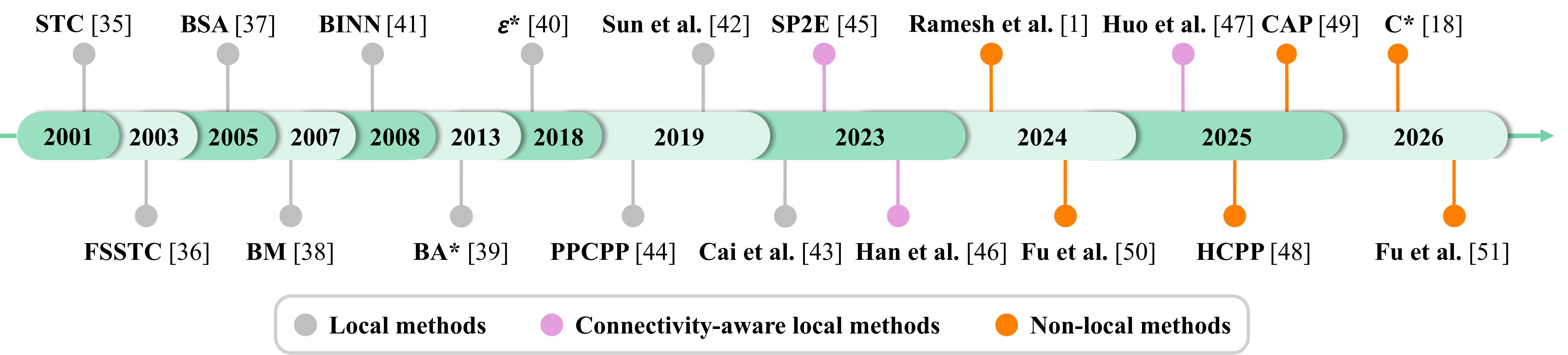}
    \caption{Timeline of the evolution of online single-robot CPP methods.}\label{fig:singleonlineTimeMap} 
    \vspace{-1.0em}
 \end{figure*}

\subsubsection{Cellular-decomposition methods} These methods partition the free space into non-overlapping cells and generate simple local coverage paths within each cell. Trapezoidal decomposition~\cite{latombe1991exact} is an early representative method that partitions the space with polygonal obstacles into non-overlapping cells. This partition, however, can produce many narrow cells in complex environments, which is undesirable for cell sequencing to avoid trajectory overlaps. Boustrophedon decomposition~\cite{choset2000coverage} improves Trapezoidal decomposition by merging adjacent cells according to connectivity changes to reduce the total number of cells. Morse-based decomposition~\cite{acar2002morse} further generalizes this idea by formulating the partitioning process in terms of critical points of Morse functions~\cite{milnor1963morse}. It is applicable to environments containing complex obstacles with smooth geometries. This line of research has been extended to online coverage in unknown environments~\cite{acar2002sensor}, optimization of the subspace traversal order~\cite{xu2014efficient}, uncertainty-aware coverage planning in belief space~\cite{schirmer2019coverage}, and constriction-based decomposition for indoor environments~\cite{brown2016constriction}. 

\subsubsection{Rank-based methods} These methods aim to reduce the number of turns in the coverage path through decomposition, making them useful for applications where turning is expensive. Bochkarev and Smith~\cite{bochkarev2016minimizing} partitioned the environment to reduce the total number of coverage lines (i.e., straight-line sweep paths), and then sequenced these lines by solving a Generalized TSP. Ramesh et al.~\cite{ramesh2022optimal} extended this by proposing an optimal partitioning method that computes the minimum number of ranks in polynomial time, thereby minimizing the total number of turns. Each rank is a rectangular subarea that can be covered by a single coverage line. Fig.~\ref{fig:ramesh2022optimal} shows an example of such partitioning. Ramesh et al.~\cite{ramesh2025} further extended this method to achieve minimum-length coverage while treating turn reduction as a secondary objective.

\textit{Strengths and Limitations:} Subarea-decomposition methods simplify the coverage planning process by partitioning the workspace into obstacle-free subareas. Cellular decomposition methods provide a systematic way to handle obstacle-induced workspace topology, but their performance depends on the detection of connectivity changes or critical points, which can be difficult in complex environments. Rank-based methods reduce the number of turns, but they require optimization to determine a compact set of ranks and their traversal order. Therefore, decomposition-based methods are effective in simple and structured environments, however, their performance might degrade in complex obstacle-dense environments.

\subsection{Online Methods}
\label{sec:single_online_related_work}

Online methods generate coverage paths incrementally in unknown environments based on sensor observations. At each planning step, the robot uses onboard sensor data to update an evolving map that records the occupancy state of the workspace, such as free, occupied, and unknown cells, as well as the coverage state of traversable cells. The updated map is then used to determine the next coverage decision.

Fig.~\ref{fig:singleonlineTimeMap} shows the timeline of major developments in single-robot online CPP. The methods in this category are further grouped into local methods, connectivity-aware local methods, and non-local methods. Table~\ref{tab:single_online_table} summarizes the key features of representative single-robot online CPP methods.

\subsubsection{Local methods}

These methods generate the coverage trajectory by selecting the next waypoint in the robot's local neighborhood using lightweight decision rules. These methods can be grouped into deterministic-rule based, potential-field based, and reward-function based methods.

a) \textit{Deterministic-rule-based methods}: These methods use deterministic rules to generate systematic coverage patterns. Gabriely and Rimon~\cite{gabriely2001spanning} proposed the spanning tree covering (STC) algorithm based on a two-resolution grid, where each coarse cell is divided into four fine cells that match the robot's footprint. Then a spanning tree is expanded by selecting unvisited neighboring coarse cells. The covering path is generated by circumnavigating the tree; however, it leaves the partially occupied cells uncovered. Full spiral STC (FSSTC)~\cite{gabriely2003competitive} extended STC by including partially occupied cells in an augmented spanning tree enabling coverage at the robot's footprint level. Gonzalez et al.~\cite{gonzalez2005bsa} proposed the backtracking spiral algorithm (BSA), which follows the boundary of the uncovered area and obstacles on a fixed lateral side to generate a spiral coverage path. Ferranti et al.~\cite{ferranti2007brick} proposed the brick-and-mortar (BM) algorithm, which gradually expands blocks of inaccessible cells while avoiding decisions that disconnect accessible cells, thus allowing the robot to escape dead-ends via previously explored cells. Viet et al.~\cite{viet2013ba} proposed the BA$^*$ algorithm, which follows a predefined directional priority order to produce back-and-forth motions. 

\begin{figure}[t]
        \centering        \includegraphics[width=0.49\textwidth]{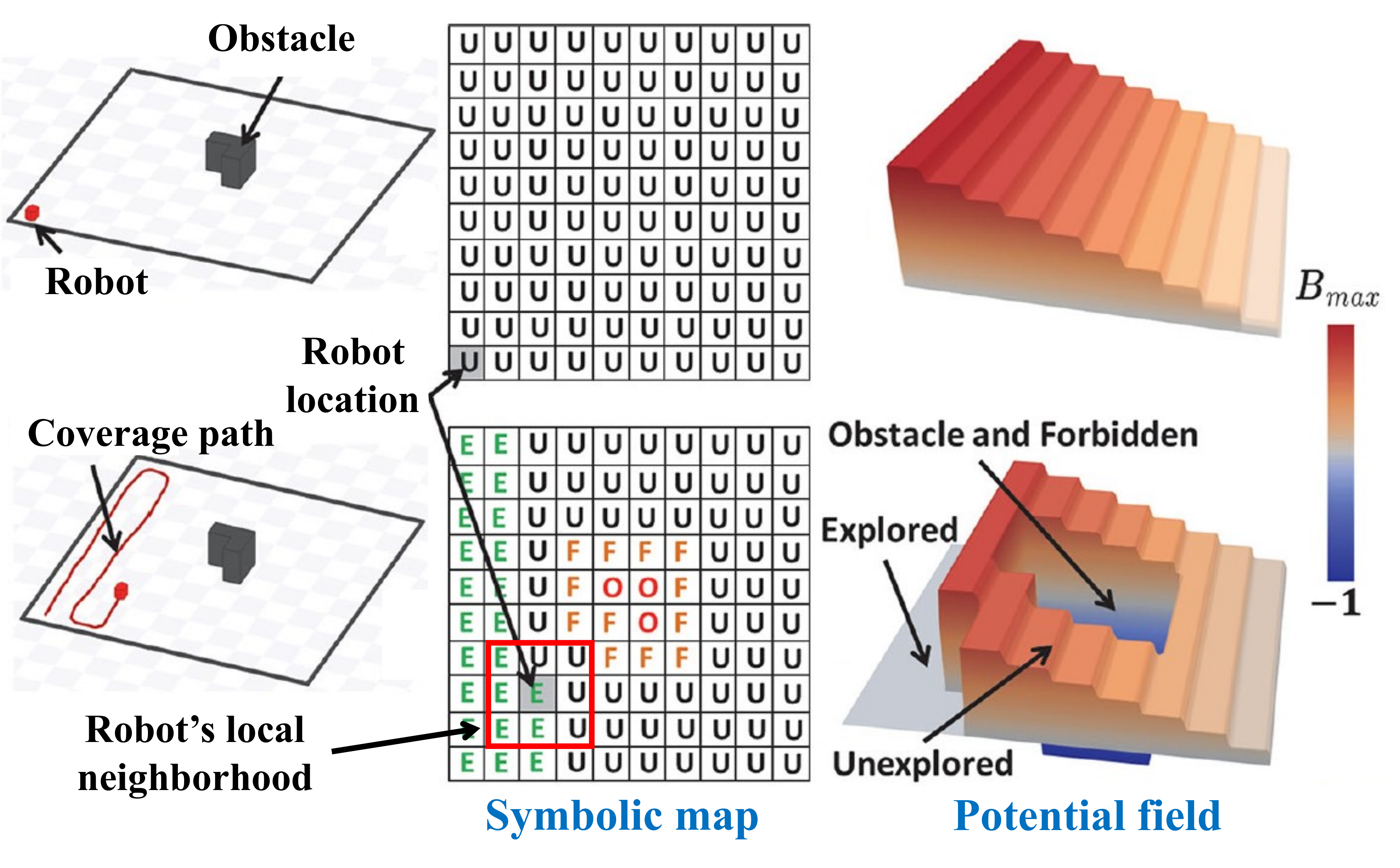}
    \caption{Illustration of the $\varepsilon^*$~\cite{song2018} operation at level 0 of MAPS, where it selects the target cell from the robot's local neighborhood.}\label{fig:song2018} 
    \vspace{-1em}
 \end{figure}

b) \textit{Potential-field-based methods}: These methods construct potential-field representations on grid maps to guide waypoint selection. Luo and Yang~\cite{luo2008bioinspired} proposed the bio-inspired neural network (BINN) algorithm, where each grid cell corresponds to a neural activity such that uncovered cells provide excitatory inputs, and obstacle cells provide inhibitory inputs. The robot moves to the neighboring cell with the highest neural activity. The activity propagation allows the robot to move toward distant uncovered regions when local uncovered cells are not available. Later variants~\cite{sun2019complete,cai2023} simplified the neural activity update to reduce computational cost. Song and Gupta~\cite{song2018} proposed the $\varepsilon^*$ algorithm, which uses multiscale adaptive potential surfaces (MAPS) to guide online coverage. By default, the algorithm operates at the lowest level of MAPS, where the robot selects the neighboring cell with the highest potential, as shown in Fig.~\ref{fig:song2018}. The higher-level potential surfaces of MAPS provide non-local information to escape from dead-ends.

c) \textit{Reward-function-based methods}: These methods use reward functions to guide waypoint
selection. They usually combine multiple objectives into a unified decision rule. Hassan and Liu~\cite{hassan2019ppcpp} proposed the predator-prey CPP (PPCPP) algorithm, in which each neighboring cell is evaluated by a reward function consisting of three terms: a predation-avoidance reward that drives the robot away from a virtual stationary predator; a smoothness reward that favors straight motion; and a boundary reward that prioritizes covering the boundary of the uncovered area. The robot then selects the neighboring cell with the highest reward.

\textit{Strengths and Limitations:} Local methods are computationally efficient and easy to implement, thus suitable for real-time coverage in unknown environments. However, their decisions are based on local information, without considering the global structure of the scenario. As such, they may generate myopic waypoints, repeated traversal, dead ends, and coverage holes in complex environments. These limitations motivate later methods that incorporate connectivity preservation and global guidance in coverage decisions. 

  \begin{figure}[t]
    \centering        
    \includegraphics[width=0.47\textwidth]{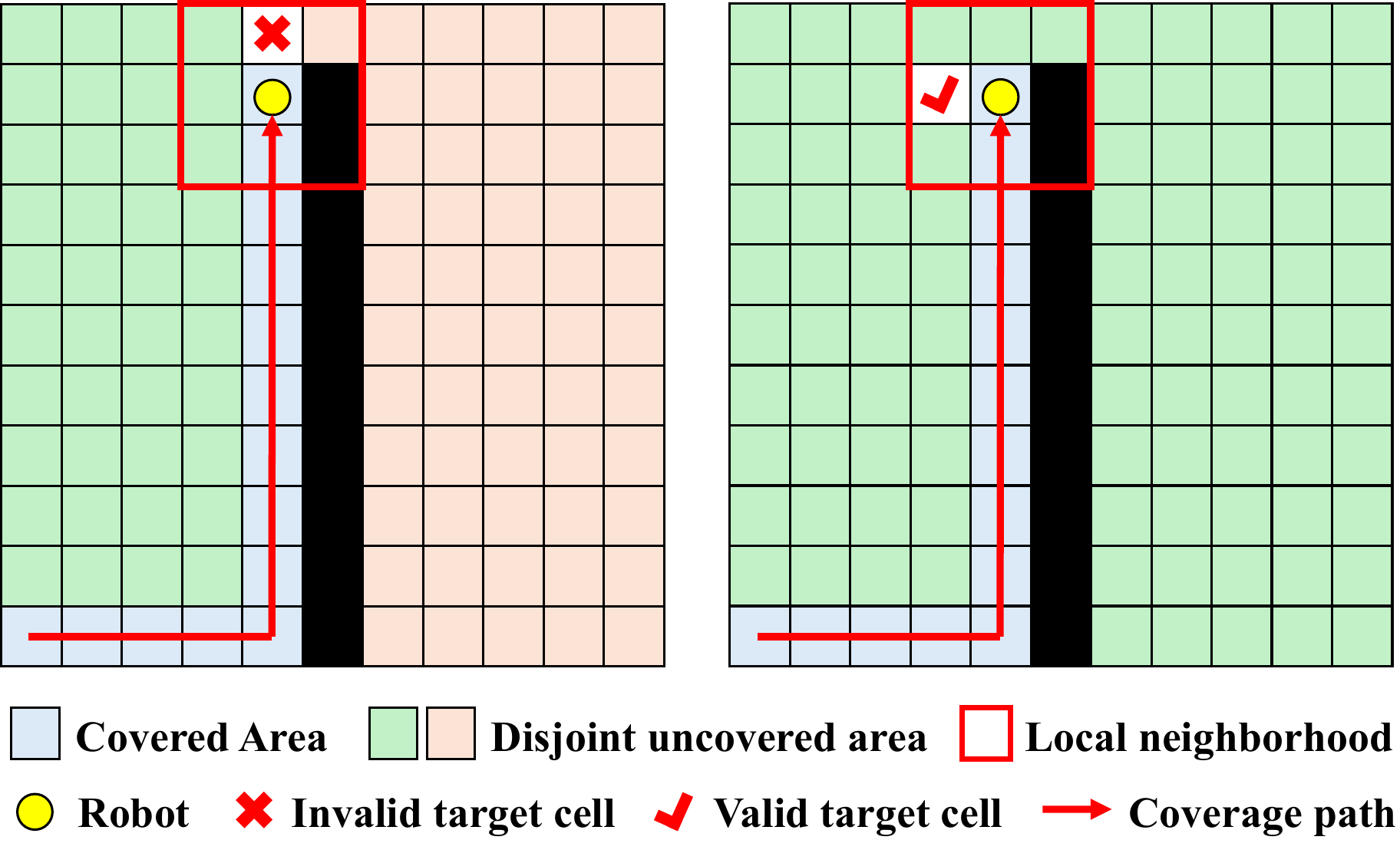}
    \caption{Illustration of the connectivity-aware local method SP2E~\cite{li2023sp2e}. It adapts the local spiral direction to select a target cell that attempts to preserve the connectivity of the uncovered area.}\label{fig:li2023sp2e} 
    \vspace{-1em}
 \end{figure}

\subsubsection{Connectivity-aware local methods}

These methods improve target-cell selection by considering whether visiting a neighboring cell would disconnect the remaining uncovered area. Li et al.~\cite{li2023sp2e} extended the BSA algorithm~\cite{gonzalez2005bsa} by developing the SP2E algorithm, which adapts the local spiral direction so that the selected target cell preserves the connectivity of the uncovered area, as shown in Fig.~\ref{fig:li2023sp2e}. Several works~\cite{han2023,huo2025} improved the BINN algorithm~\cite{luo2008bioinspired} by avoiding target cells whose visitation would disconnect the uncovered area. 

\textit{Strengths and Limitations:} By preventing locally selected cells from disconnecting the uncovered area, these methods reduce fragmentation-induced redundant travel. However, these methods still rely on local feasibility checks. In situations when all candidate target cells disconnect the uncovered area, these methods can still produce fragmented residual regions (i.e., coverage holes) leading to considerable redundant travel.

\subsubsection{Non-local methods}

These methods use higher-level decision structures to guide the selection of non-local waypoints for coverage trajectory generation.

a) \textit{Subarea-decomposition methods}:
These methods use online map updates to organize the uncovered area into non-uniform subareas and reason over their connectivity or traversal order. Li et al.~\cite{li2025} proposed the hierarchical coverage path planning (HCPP) algorithm, which aims to maintain the connectivity of the uncovered area through global and local aspects. Globally, the algorithm partitions the uncovered area into stripe-shaped subareas and represents their adjacency relationships as a graph. It then selects subareas whose coverage does not disconnect the uncovered area to form a global tour. Locally, if the current subarea is connected to the target subarea, the robot follows the boundary of the uncovered area to reach it; otherwise, a shortest collision-free path is used.

Another line of work explicitly sequences disconnected uncovered subareas to guide local coverage. Shen et al.~\cite{shen2025_cap} proposed the connectivity-aware hierarchical coverage path planning (CAP) algorithm, which combines global sequencing of disconnected uncovered subareas with an adaptive local coverage strategy. The algorithm incrementally constructs a graph in which disconnected uncovered subareas are represented as nodes, and the edges encode the shortest collision-free connections between them. A subarea traversal tour is then computed by solving a Traveling Salesman Problem. During local execution, it uses a local strategy in exploring subareas, i.e., subareas that contain unknown cells, and a shortest coverage path in explored subareas.

\begin{figure}[t]
    \centering
    \subfloat[Generation of frontier samples]{
    \includegraphics[width=0.49\columnwidth]{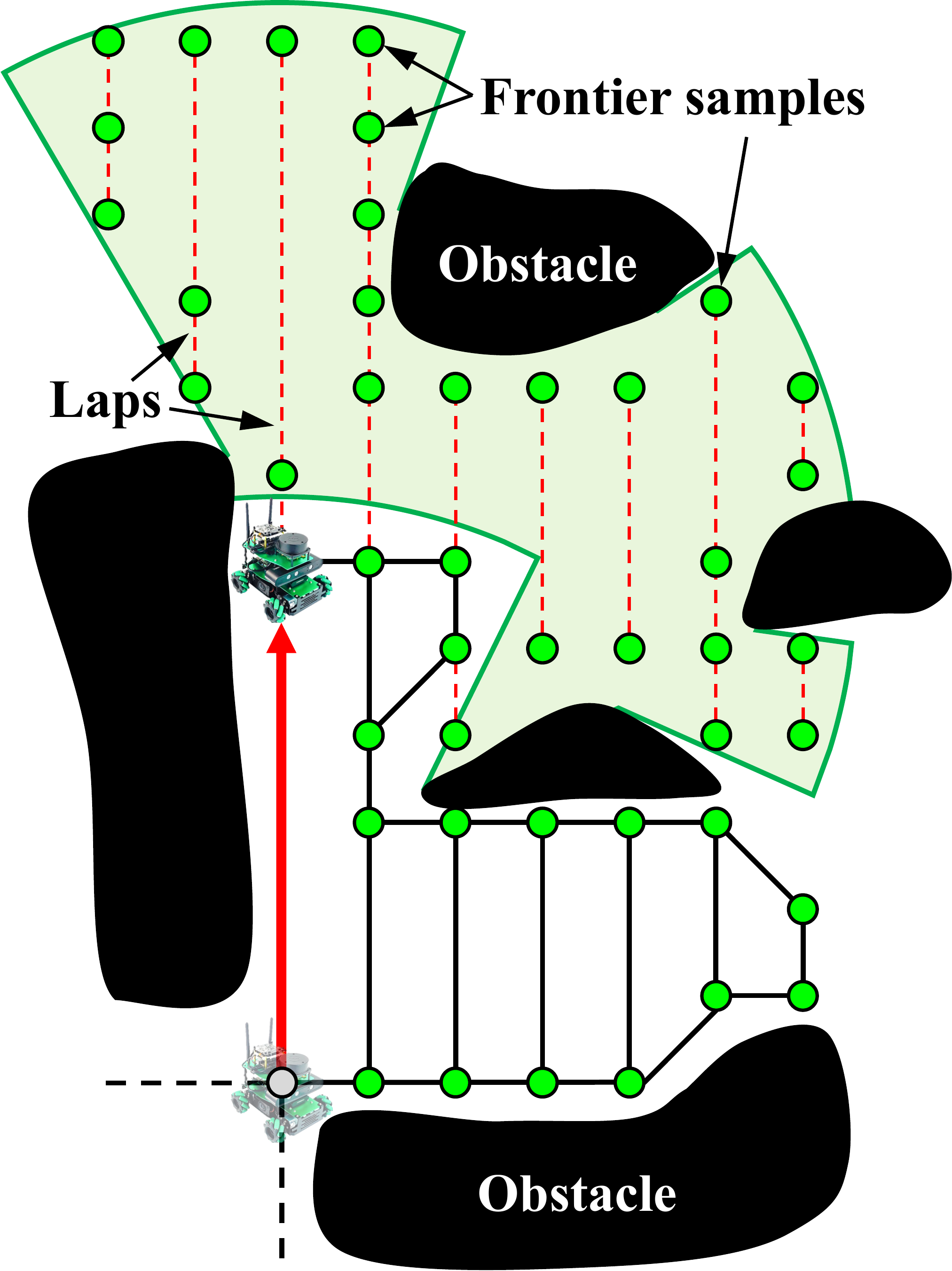}\label{fig:shen2026_part1}}\hspace{-12pt}\quad
    \centering
    \subfloat[Updated RCG]{
    \includegraphics[width=0.49\columnwidth]{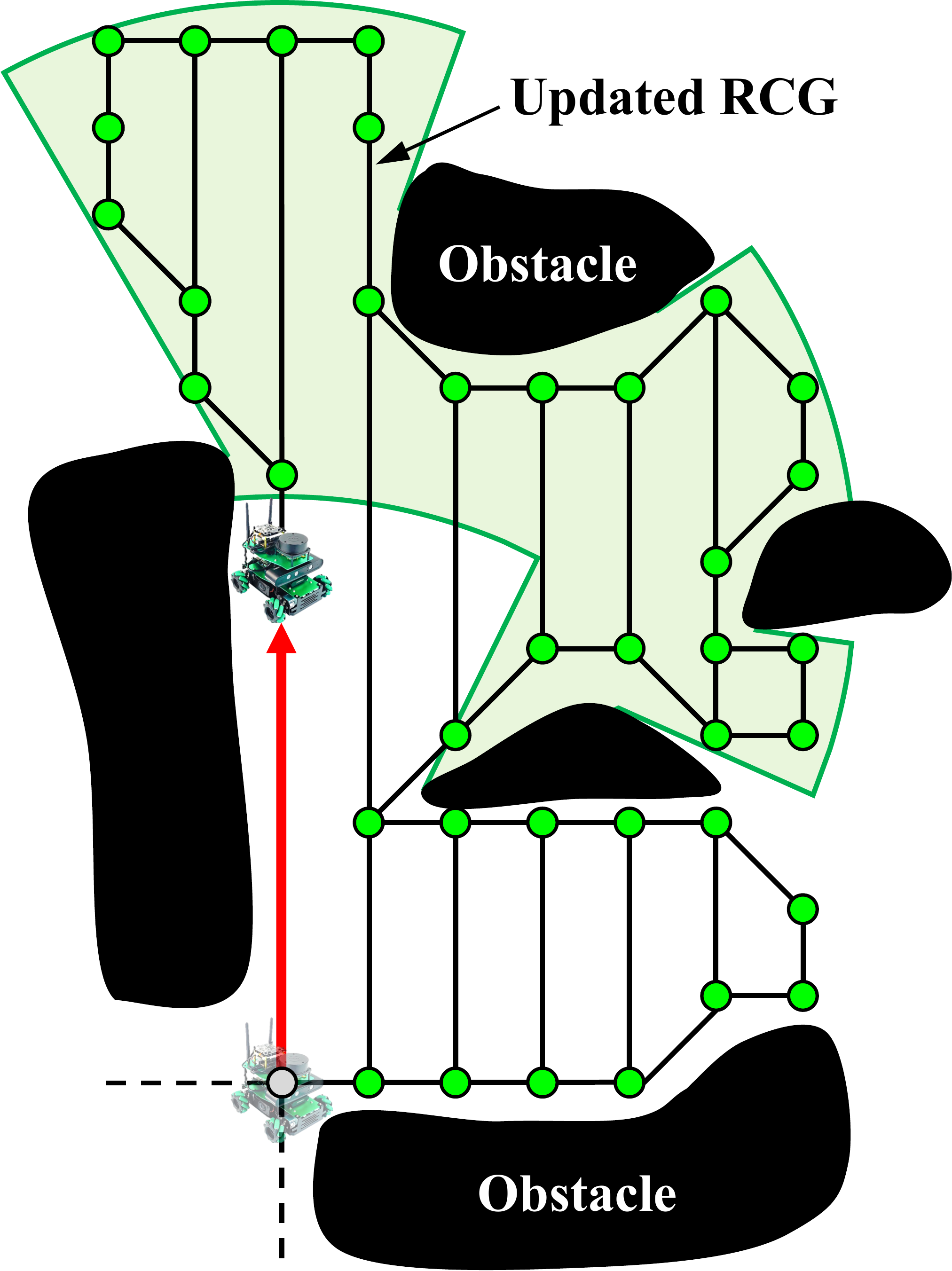}\label{fig:shen2026_part2}}\\
          \caption{Illustration of the sampling-based method C$^*$~\cite{shen2026} . During navigation, C$^*$ generates frontier samples within the newly-discovered area by sensors. Then it updates the RCG using these samples to guide the formation of subsequent coverage trajectory.}\label{fig:shen2026}
          \vspace{-1em}
\end{figure}

\begin{table*}[t]{}
\footnotesize
\caption {Comparison of representative online single-robot CPP methods.}\label{tab:single_online_table}\vspace{-3pt}
\centering
\setlength\tabcolsep{3.0pt}
\begin{tabular}{l l l l l l l l} 
 \toprule
\specialrule{0.1em}{1pt}{1pt} 
\tabincell{l}{\textbf{Category}}
&\multicolumn{1}{l}{\textbf{Method}}
&\tabincell{c}{\textbf{Year}}
&\tabincell{c}{\textbf{Main Idea}}
&\tabincell{c}{\tabincell{c}{{\textbf{Coverage Guidance}}}}
&\tabincell{c}{{\textbf{Decision Domain}}}  
&\tabincell{c}{\textbf{Strength}} 
&\tabincell{c}{\textbf{Limitation}}\\ 
\toprule

\specialrule{0em}{1pt}{0pt}
\multirow{14.7}{*}{\tabincell{l}{\textbf{{Local}}}}
& \tabincell{l}{\textbf{BSA}~\cite{gonzalez2005bsa}}       & 2005     &\tabincell{l}{{Follows boundary of}\\{uncovered area on a}\\{certain lateral side}} & \tabincell{l}{{Deterministic}\\{rule}}  & \multirow{14.5}{*}{\tabincell{l}{{Robot's local}\\{neighborhood}}} & \multirow{14.5}{*}{\tabincell{l}{{Computationally}\\{efficient and easy}\\{to implement}}} & \multirow{14.5}{*}{\tabincell{l}{{Prone to myopic}\\{waypoints in}\\{complex scenarios}}}\vspace{0.5em}\\

& \tabincell{l}{\textbf{BINN}~\cite{luo2008bioinspired}}      & 2008     &\tabincell{l}{{Constructs a potential}\\{field to guide target}\\{cell selection}} & \tabincell{l}{{Potential}\\{field}}  & & & \vspace{0.5em}\\

& \tabincell{l}{\textbf{BA$^*$}~\cite{viet2013ba}}      & 2013     &\tabincell{l}{{Follows a priority of}\\{north-south-east-west}\\{to select target cells}} & \tabincell{l}{{Deterministic}\\{rule}}  & & & \vspace{0.5em}\\

& \tabincell{l}{\textbf{$\varepsilon^*$}~\cite{song2018}}     & 2018     &\tabincell{l}{{Uses multiscale potential}\\{field to pick target cell}\\ and escape dead-ends} & \tabincell{l}{{Multiscale}\\{potential}\\{field}}  & & & \vspace{0.5em}\\

& \tabincell{l}{\textbf{PPCPP}~\cite{hassan2019ppcpp}}     & 2019     &\tabincell{l}{{Uses a multi-objective}\\{reward function to guide}\\{target cell selection}} & \tabincell{l}{{Reward}\\{function}}  & & & \\

\specialrule{0.05em}{2pt}{2pt}
\multirow{4.5}{*}{\tabincell{l}{\textbf{{Conne-}}\\\textbf{{ctivity-}}\\\textbf{{aware}}\\\textbf{{local}}}}
& \tabincell{l}{\textbf{SP2E}~\cite{li2023sp2e}}     & 2023     &\tabincell{l}{{Adapts direction of spiral}\\{path to keep connectivity}\\{of uncovered area}} & \multirow{4.5}{*}{\tabincell{l}{{Local conne-}\\{ctivity checks}}} &\multirow{4.5}{*}{\tabincell{l}{{Robot's local}\\{neighborhood}}}  & \multirow{4.5}{*}{\tabincell{l}{{Reduces}\\{fragmentation-}\\{induced redundant}\\{travel}}} & \multirow{4.5}{*}{\tabincell{l}{{May still fragment}\\{uncovered area}\\{in complex scenarios}}}\vspace{0.5em}\\

& \tabincell{l}{\textbf{Huo et al.}~\cite{huo2025}}     & 2025     &\tabincell{l}{{Avoids target cells}\\{that would disconnect}\\{the uncovered area}} &   & & & \\

\specialrule{0.05em}{2pt}{2pt}
\multirow{11}{*}{\tabincell{l}{\textbf{{Non}}\\\textbf{{-local}}}}
& \tabincell{l}{\textbf{Ramesh et al.}~\cite{ramesh2024}}     & 2024     &\tabincell{l}{{Repairs the interrupted}\\{path portion using a}\\{rank-based method}} & \tabincell{l}{{Current full}\\{coverage path}} &\multirow{11}{*}{\tabincell{l}{{Higher-level}\\{representation}\\{of environments}}} & \multirow{11}{*}{\tabincell{l}{{Improves overall}\\{coverage efficiency}\\{through higher-level}\\{structural guidance}}} & \multirow{11}{*}{\tabincell{l}{{Performance depends}\\{on the efficiency of}\\{higher-level decision}\\{structures}}}\vspace{0.5em}\\

& \tabincell{l}{\textbf{HCPP}~\cite{li2025}}     & 2025     &\tabincell{l}{{Prioritizes subareas whose}\\{coverage keeps connectivity}\\{of uncovered area}}  &\tabincell{l}{{Connectivity-}\\{aware subarea}\\{tour}}  & & & \vspace{0.5em}\\

& \tabincell{l}{\textbf{CAP}~\cite{shen2025_cap}}     & 2025     &\tabincell{l}{{Computes shortest traversal}\\{tour of uncovered subareas}\\{to guide local coverage}}  &\tabincell{l}{{Shortest}\\{subarea tour}}  & & & \vspace{0.5em}\\

& \tabincell{l}{\textbf{C$^*$}~\cite{shen2026}}     & 2026     &\tabincell{l}{{Creates a rapidly covering}\\{graph incrementally to}\\{guide target cell selection}}  &\tabincell{l}{{Rapidly}\\{covering}\\{graph}}  & & & \\

\bottomrule
\specialrule{0.1em}{1pt}{1pt}
\end{tabular}
\vspace{-1.0em}
\end{table*}

b) \textit{Sampling-based methods}: These methods use non-uniform sampling representations to provide non-local guidance for online coverage. Shen et al.~\cite{shen2026} proposed a sampling-based algorithm, called C$^*$, which provides near-optimal coverage in unknown environments. During navigation, C$^*$ builds a rapidly covering graph (RCG), which is a minimum-sufficient graphical representation of the obstacle-free space. RCG is incrementally constructed via progressive sampling and efficient graph pruning, as shown in Fig.~\ref{fig:shen2026}. Since RCG is sparse, C$^*$ selects nonmyopic nodes as waypoints to generate the back-and-forth coverage pattern and escape from dead-end situations. In addition, it proactively detects potential coverage holes and adapts to the TSP-based optimal trajectories to cover such regions, thus improving the coverage performance.

c) \textit{Path-repair methods}: These methods assume that an initial coverage path is available from the current map and then treat online CPP as a path-repair problem to update the portions of the path that become invalid when new obstacles are detected. Ramesh et al.~\cite{ramesh2024} partitioned the affected uncovered region into a minimum number of ranks, where each rank is a horizontal or vertical sweep over a strip of grid cells. A tour of these ranks is then computed to repair the path. Fu et al.~\cite{fu2024full,fu2026online} instead split the interrupted path into contiguous segments and reconnect successive segments through collision-free transitions. These methods are efficient when obstacle-induced interruptions remain local. However, their advantage may diminish when newly detected obstacles cause frequent or extensive fragmentation of the original coverage path.

\textit{Strengths and Limitations:} Non-local methods incorporate higher-level decision structures into online decision-making, which helps reduce myopic waypoint selection, dead ends, repeated backtracking, and fragmentation-induced redundant travel.
However, their performance depends on the efficient construction, updating, and utilization of higher-level decision structures, such as subarea graphs, sampling-based representations, or repaired coverage paths.

\section{Multi-Robot CPP}
\label{sec:multirobot}

Multi-robot CPP extends single-robot CPP by using multiple robots to cover a target workspace collaboratively. In comparison to single-robot CPP, the multi-robot CPP problem is not limited to generating a single coverage trajectory, but also requires distributing coverage responsibilities among robots, balancing workloads, avoiding inter-robot conflicts, and coordinating local or global replanning. By exploiting multiple robots, multi-robot CPP can accelerate large-scale coverage, provide resilience to robot failures, and support distributed sensing and execution. 

Existing multi-robot CPP methods can be broadly classified as offline or online, depending on whether the environment is known before deployment. Offline methods exploit prior map information to partition the workspace, allocate robot-specific coverage tasks, and optimize coverage paths before execution. Online methods, in contrast, operate in unknown environments,  update the map, reallocate coverage tasks, replan coverage paths, and avoid conflicts during execution.

\subsection{Offline Methods}
\label{sec:multi_offline_related_work}

The offline methods can be broadly classified into spanning-tree-based and subarea-decomposition methods. Table~\ref{tab:multi_offline_table} summarizes the key features of offline multi-robot CPP methods.

 \begin{table*}[t]{}
\footnotesize
\caption {Comparison of representative offline multi-robot CPP methods.}\label{tab:multi_offline_table}\vspace{-3pt}
\centering
\setlength\tabcolsep{5pt}
\begin{tabular}{l l l l l l l l} 
 \toprule
\specialrule{0.1em}{1pt}{1pt} 
\tabincell{l}{\textbf{Category}}
&\multicolumn{1}{l}{\textbf{Method}}
&\tabincell{c}{\textbf{Year}}
&\tabincell{l}{\textbf{Main Idea}} 
&\tabincell{l}{{\textbf{Planning}}\\{\textbf{Representation}}} 
&\tabincell{l}{{\textbf{Allocation}}\\{\textbf{Unit}}} 
&\tabincell{l}{{\textbf{Optimization}}\\{\textbf{Focus}}} \\ 
\toprule

\specialrule{0em}{1pt}{0pt}
\multirow{12}{*}{\tabincell{l}{\textbf{{Spanning-}}\\\textbf{{tree-based}}}}
& \tabincell{l}{\textbf{AWSTC}~\cite{dong2020artificially}}       & 2020     &\tabincell{l}{{Grows robot-specific trees}\\{using weight-based grid}\\{cell selection strategy}} &\tabincell{l}{{Robot-specific}\\{spanning tree}} &\tabincell{l}{{Grid cell}} &\tabincell{l}{{Workload balancing}}  \vspace{0.5em}\\

& \tabincell{l}{\textbf{MSTC$^*$}~\cite{tang2021mstc}}  & 2021     &\tabincell{l}{{Partitions a global STC-}\\{based path into balanced}\\{robot-specific subpaths}}  &Global STC path &\tabincell{l}{Path segment} &\tabincell{l}{{Workload balancing}}  \vspace{0.5em}\\ 

& \tabincell{l}{\textbf{TMSTC$^*$}~\cite{lu2023tmstc}} & 2023     &\tabincell{l}{{Generates a turn-reduced}\\{global STC path and}\\{partitions it using MSTC$^*$}}  &Global STC path &\tabincell{l}{Path segment} &\tabincell{l}{{Turn reduction}}  \vspace{0.5em}\\

& \tabincell{l}{\textbf{PAMCPP}~\cite{lee2026}}      & 2026     &\tabincell{l}{{Covers high-priority zones}\\{through zone-wise spanning}\\{trees, followed by Steiner-}\\{tree-based residual coverage}} &\tabincell{l}{{Priority-zone-}\\{wise spanning}\\{tree, Steiner tree}\\{in residual area}} &\tabincell{l}{{Priority-zones}\\{and residual}\\{path segment}} &\tabincell{l}{{Priority-weighted latency}}  \\

\specialrule{0.05em}{2pt}{2pt}
\multirow{14}{*}{\tabincell{l}{\textbf{{Subarea-}}\\\textbf{decomposition}}}
& \tabincell{l}{\textbf{Tang et al.}~\cite{tangyuan2020}}     & 2020     &\tabincell{l}{{Constructs a disk-based}\\{representation of the space}\\{for task assignment and}\\{coverage path planning}} &\tabincell{l}{{A set of disk-}\\{shaped regions}} &\tabincell{l}{{Disk-shaped}\\{regions}} &\tabincell{l}{{Workload balancing}} \vspace{0.5em}\\

& \tabincell{l}{\textbf{Agarwal et al.}~\cite{agarwal2022area}}     & 2022     &\tabincell{l}{{Partitions the environment}\\{to reduce turns, generates}\\{coverage lines for task}\\{allocation and sequencing}}   &\tabincell{l}{{A set of subareas}\\{with coverage lines}} &Coverage line &\tabincell{l}{{Turn reduction}} \vspace{0.5em}\\

& \tabincell{l}{\textbf{Cao et al.}~\cite{cao2026multi}}     & 2026     &\tabincell{l}{{Partitions a subarea-based}\\{graph into connectivity-}\\{preserving subgraphs for}\\{coverage path generation}}    &\tabincell{l}{{Subarea-based}\\{adjacency graph}} &\tabincell{l}{{Subgraph}} &\tabincell{l}{{Connectivity preservation}} \vspace{0.5em}\\

& \tabincell{l}{\textbf{Huang et al.}~\cite{huang2026comaea}}     & 2026     &\tabincell{l}{{Jointly optimizes subarea}\\{visiting orders and local}\\{coverage modes}}    &\tabincell{l}{{Trapezoidal}\\{subareas}} &\tabincell{l}{{Subarea and}\\{its coverage}\\{mode}} &\tabincell{l}{{Coverage path length}}\\

\bottomrule
\specialrule{0.1em}{1pt}{1pt}
\end{tabular}
\vspace{-1.0em}
\end{table*}

\subsubsection{Spanning-tree-based methods}
These methods apply the STC algorithm~\cite{gabriely2001spanning} originally developed for single-robot CPP to multi-robot settings. 

a) \textit{Global STC path partitioning}: In this line of work, a global STC path is first generated and then partitioned into segments to be distributed among multiple robots. Hazon and Kaminka extended STC to multi-robot STC (MSTC)~\cite{hazon2005redundancy},  where the STC path is partitioned using the robots' initial positions. Although this provides a simple partitioning rule, the resulting workload depends on the robots' initial distribution. In general, optimizing multi-robot coverage using the STC framework requires determining robot-specific constraints and balancing their coverage times, which is NP-hard~\cite{zheng2010multirobot}. Therefore, later methods mainly rely on approximations and heuristics for workload balancing, turn reduction, completion-time minimization, priority handling, and conflict avoidance.

\begin{figure}[t]
        \centering        \includegraphics[width=0.47\textwidth]{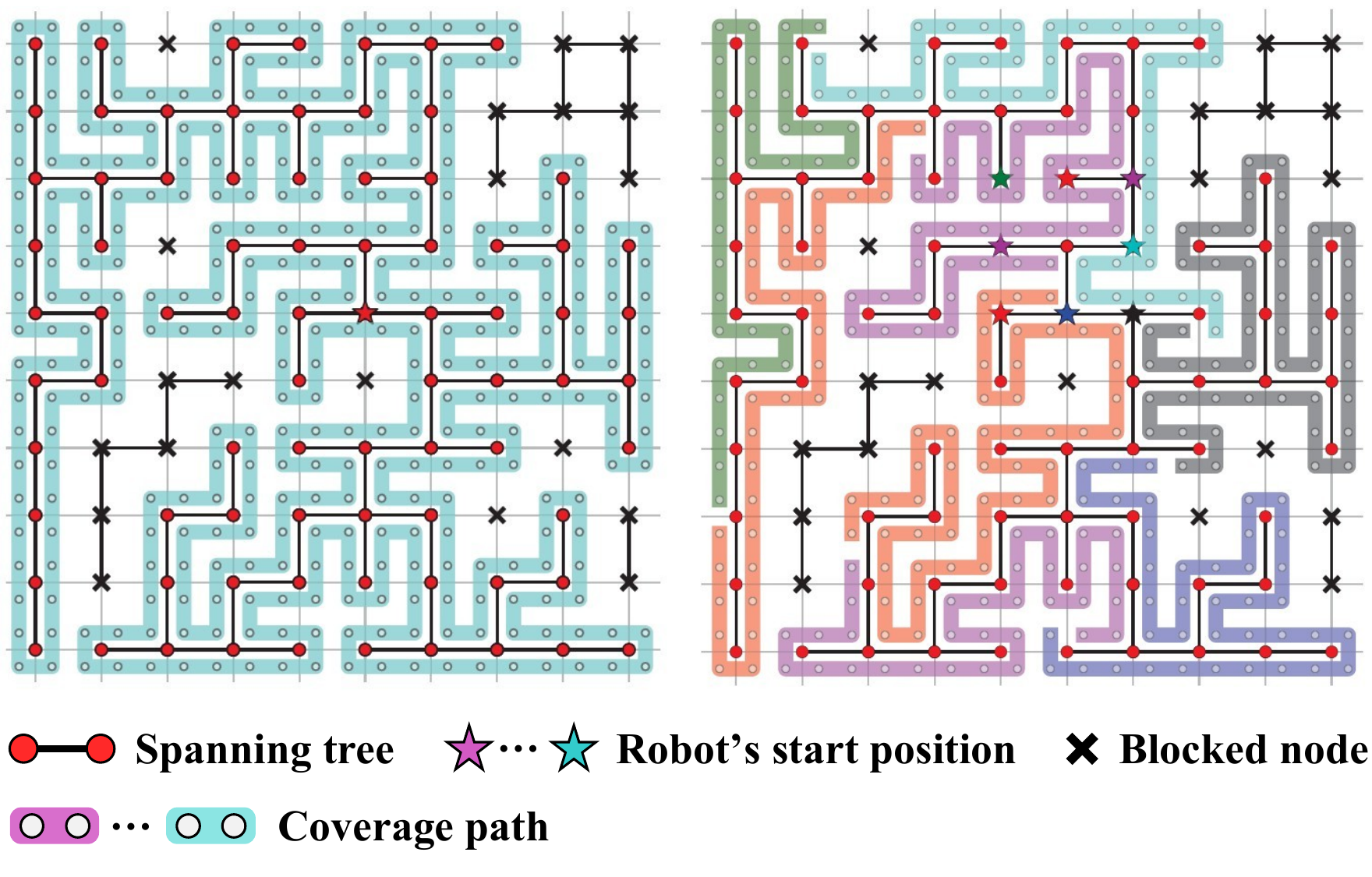}
    \caption{Illustration of the spanning-tree-based method MSTC$^*$~\cite{tang2021mstc}. It partitions a global STC-based coverage path into robot-specific subpaths considering  traversability constraints and workload capacity.}\label{fig:tang2021mstc} 
    \vspace{-1em}
 \end{figure}

Tang et al.~\cite{tang2021mstc} proposed MSTC$^*$ to improve the workload balance of MSTC considering the physical constraints of traversability and limited workload capacity. The method treats the global STC-based coverage path as a topological loop and partitions it into robot-specific subpaths, as shown in Fig.~\ref{fig:tang2021mstc}. It first generates a naive uniform partition of the STC path and then iterates it to produce balanced partitions. Lu et al.~\cite{lu2023tmstc} further proposed TMSTC$^*$ to reduce turns by constructing a spanning tree using a minimum number of rectangular bricks before applying MSTC$^*$ for partitioning.

b) \textit{Robot-specific tree construction}: This line of work directly constructs robot-specific spanning trees instead of partitioning a single global STC path. Dong et al.~\cite{dong2020artificially} proposed the artificially weighted STC (AWSTC) algorithm, where each robot grows a spanning tree from its initial position using artificial weights that encourage expansion toward uncovered regions, separation from other robots, and reduction of repeated coverage. Tang and Ma~\cite{tang2023mixed} formulated time-optimal multi-robot CPP as a mixed-integer programming problem that computes robot-specific subtrees while minimizing the maximum subtree cost, and introduced scalable heuristics to reduce the search space. Tang et al.~\cite{tang2025} further improved robot-specific subtrees through local neighborhood operators and post-processing for path deconfliction. Chen et al.~\cite{chen2026dynamic} combined work-area allocation with turn-reduced spanning tree construction by assigning smooth work areas to robots and then merging tree branches according to turn cost.

c) \textit{Priority-aware planning}: Beyond workload balancing and path-length objectives, the priority-aware methods consider regional priorities across the workspace. Lee et al.~\cite{lee2026} proposed PAMCPP, which assigns prioritized zones to robots, constructs zone-wise spanning trees according to the assigned visiting order, and then covers the residual region using a Steiner-tree-based workload-balancing scheme. Thus, the high priority zones are covered before the remaining area, while maintaining complete coverage.

\textit{Strengths and Limitations:} Spanning-tree-based methods provide systematic coverage; however, their performance depends strongly on how the global tree or robot-specific subtrees are constructed. Balancing workload, reducing turns, and avoiding inter-robot conflicts often require additional optimization or heuristic refinement.

\subsubsection{Subarea-decomposition methods}
These methods partition the free space into subareas and then plan the traversal order between them and the local coverage path within each subarea. These methods provide a flexible framework for balancing the workload and improving local coverage quality. 

a) \textit{Workload-aware partitioning}: This line of work focuses on balancing the coverage workload across robots. Karapetyan et al.~\cite{karapetyan2017efficient} grouped Boustrophedon cells~\cite{choset2000coverage} into clusters and planned back-and-forth coverage for a robot within each cluster. Palacios-Gas{\'o}s et al.~\cite{palacios2019equitable} used geodesic power diagrams to compute equitable partitions according to importance-weighted workload. Tang et al.~\cite{tangyuan2020} represented the environment with disk-shaped coverage regions, clustered them using K-means, and assigned them to robots. A coverage path is then generated for each cluster. Collins et al.~\cite{collins2021scalable} partitioned the environment into square cells and clustered them using Lloyd's algorithm~\cite{lloyd1982least}. An auction process reallocates contested boundary cells using a bias factor based on the distance between each robot's initial position and its closest assigned cell. Choton et al.~\cite{choton2023optimal} decomposed the environment into trapezoidal subareas and used dynamic programming to merge and assign contiguous subareas for balanced coverage.

   \begin{figure}[t]
        \centering        \includegraphics[width=0.47\textwidth]{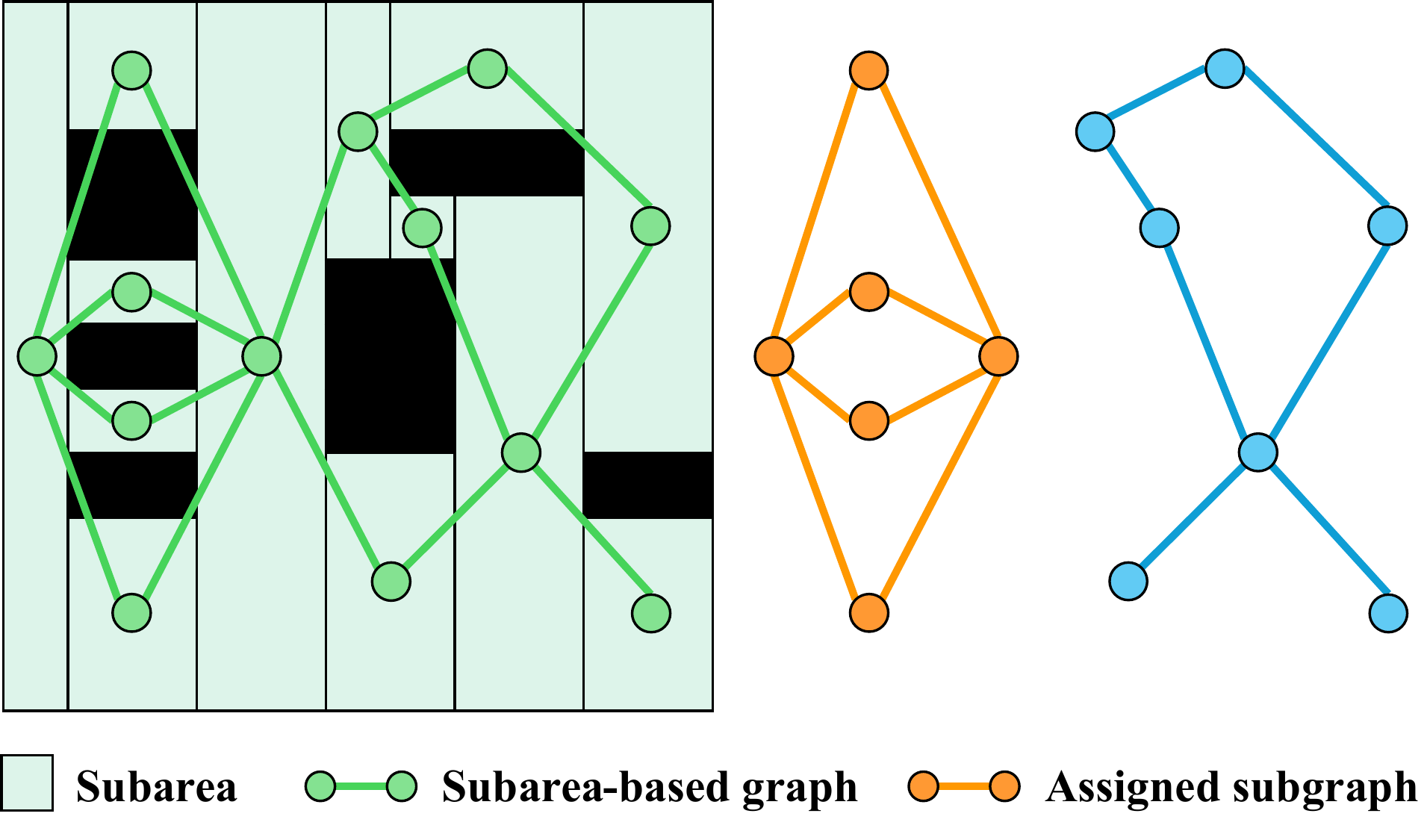}
    \caption{Illustration of the subarea-decomposition method in~\cite{cao2026multi}. The environment is decomposed into subareas and represented as an adjacency graph, which is then partitioned into connected subgraphs assigned to individual robots for coverage path generation.}\label{fig:cao2026multi} 
    \vspace{-1em}
 \end{figure}

b) \textit{Coverage-quality-aware partitioning}: This line of work improves coverage quality via decomposition design. Modares et al.~\cite{modares2017ub} assigned grid cells to robots based on an energy cost to form robot-specific subareas, and then generated a minimum-energy path within each subarea. Vandermeulen et al.~\cite{vandermeulen2019turn} proposed turn-minimizing rank decomposition, where each rank is covered by a straight-line sweep and ranks are assigned via a minmax m-TSP formulation~\cite{wang2017memetic}. Agarwal and Akella~\cite{agarwal2022area} converted area coverage into a capacity-constrained line coverage problem by generating coverage lines from turn-reduced decomposition and routing multiple robots over them~\cite{agarwal2020line}. Cao et al.~\cite{cao2026multi} built an adjacency graph over sweep-line-generated subareas and refined connected robot-specific graph partitions via node migration and swap operations, as shown in Fig.~\ref{fig:cao2026multi}. A spanning tree is then built within assigned subgraph to generate the final coverage path.

c) \textit{Joint task allocation and path generation}: A more integrated approach couples subarea allocation with local coverage path generation. Datsko et al.~\cite{datsko2024energy} generated multiple candidate back-and-forth paths for each Boustrophedon cell~\cite{choset2000coverage} and formulated path allocation and ordering as a Multiple Set Traveling Salesman Problem. Huang et al.~\cite{huang2026comaea} considered conflict-aware multi-robot coverage by jointly optimizing subarea visiting orders and intra-area coverage modes using a cooperative evolutionary algorithm, followed by spatio-temporal conflict detection and resolution.

The above methods consider a single contiguous workspace. In many applications, however, robots must cover multiple physically separated areas, such as separated fields, inspection zones, or disaster-response regions. Xie and Chen~\cite{xie2022multiregional} addressed this problem for multiple energy-constrained UAVs by jointly optimizing region visiting order and intra-region coverage paths. Zhang et al.~\cite{zhang2025multi} further allowed multiple UAVs to jointly cover the same region by optimizing region sequences and fractional coverage proportions. Later extension considered path repair after UAV failures~\cite{luo2026real} 

\textit{Strengths and Limitations:} Subarea-decomposition methods provide greater flexibility by incorporating workload, energy, turn cost, and local coverage quality into the partitioning process. However, their performance depends strongly on decomposition quality, subarea assignment, and inter-subarea transition planning. Furthermore, the joint optimization of these coupled factors remains computationally challenging.

\subsection{Online Methods}
\label{sec:multi_online_related_work}
Online methods generate coverage decisions on the fly while the robots explore the environment, where each robot uses its sensor observations to update the occupancy map and coverage states of the workspace. The coverage decisions are made in a coordinated manner between multiple robots using shared information such as robot positions, local maps, task states, robot states (e.g., battery life), or other information. Existing methods can be classified into grid-based methods and subarea-decomposition methods. Table~\ref{tab:multi_online_table} summarizes the key features of representative multi-robot online CPP methods.

 \begin{figure}[t]
        \centering        \includegraphics[width=0.47\textwidth]{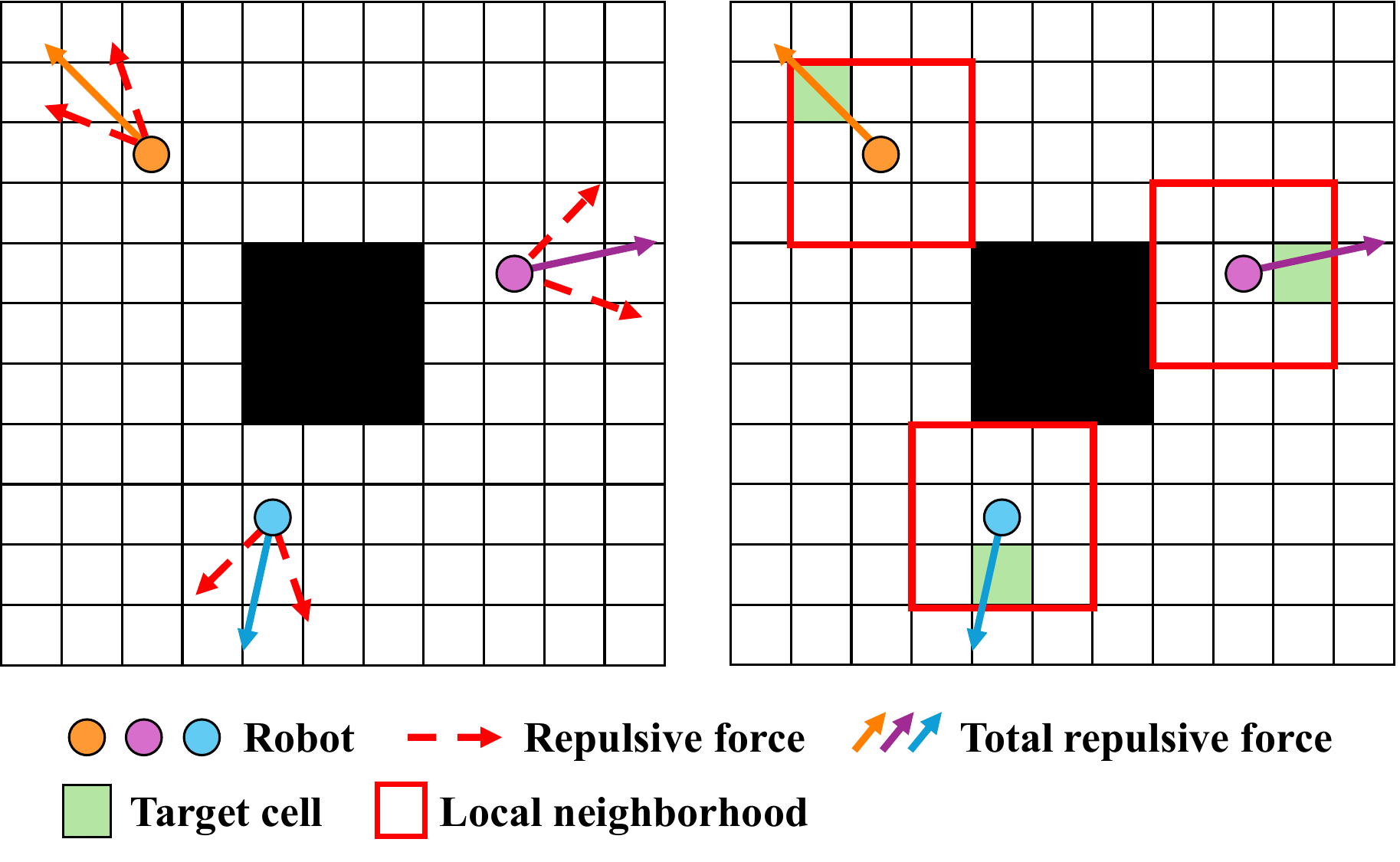}
    \caption{Illustration of the grid-based method APFCPP~\cite{wang2024apf}. It computes total repulsive force for each robot and selects the nearest uncovered cell that is most aligned with this force as target cell.}\label{fig:wang2024apf} 
    \vspace{-1em}
 \end{figure}

\subsubsection{Grid-based methods}
These methods utilize a grid map to make coverage decisions, where each robot usually selects its next target grid cell within a local neighborhood, while inter-robot interference is handled by treating other robots as obstacles or introducing repulsive effects.

Early grid-based methods extended single-robot CPP algorithms with inter-robot avoidance mechanisms. Viet et al.~\cite{viet2015bob} proposed the BoB algorithm as a multi-robot extension of BA$^*$~\cite{viet2013ba}, where each robot performs systematic back-and-forth coverage under a predefined directional priority order and treats directions blocked by other robots as unavailable. Luo et al.~\cite{luo2017neural} extended the BINN algorithm~\cite{luo2008bioinspired} to multi-robot coverage through a neural dynamics (ND) model, where uncovered cells provide positive excitation, while obstacles and other robots provide inhibitory effects. Each robot selects the neighboring cell with the highest neural activity.

Another line of grid-based methods promotes spatial separation through inter-robot distance terms or repulsive forces. Hassan et al.~\cite{hassan2020dec} proposed the decentralized predator-prey CPP (DecPPCPP) algorithm, which builds on PPCPP~\cite{hassan2019ppcpp}. Each robot treats the other robots as dynamic predators and evaluates neighboring uncovered cells using an aggregated distance term, which is incorporated into a multi-objective cost function for target-cell selection. Zhang et al.~\cite{zhang2024herd} extended this idea to the dual-environmental herd-foraging-based CPP (DHCPP) algorithm, which prioritizes cells near each robot’s starting position and gradually expands the covered region outward. Wang et al.~\cite{wang2024apf} proposed the artificial potential field based CPP (APFCPP) algorithm, as shown in Fig.~\ref{fig:wang2024apf}. It computes the total repulsive force from the other robots and selects, among the nearest uncovered cells in the local neighborhood, the one most aligned with this force.

  \begin{figure}[t]
        \centering        \includegraphics[width=0.47\textwidth]{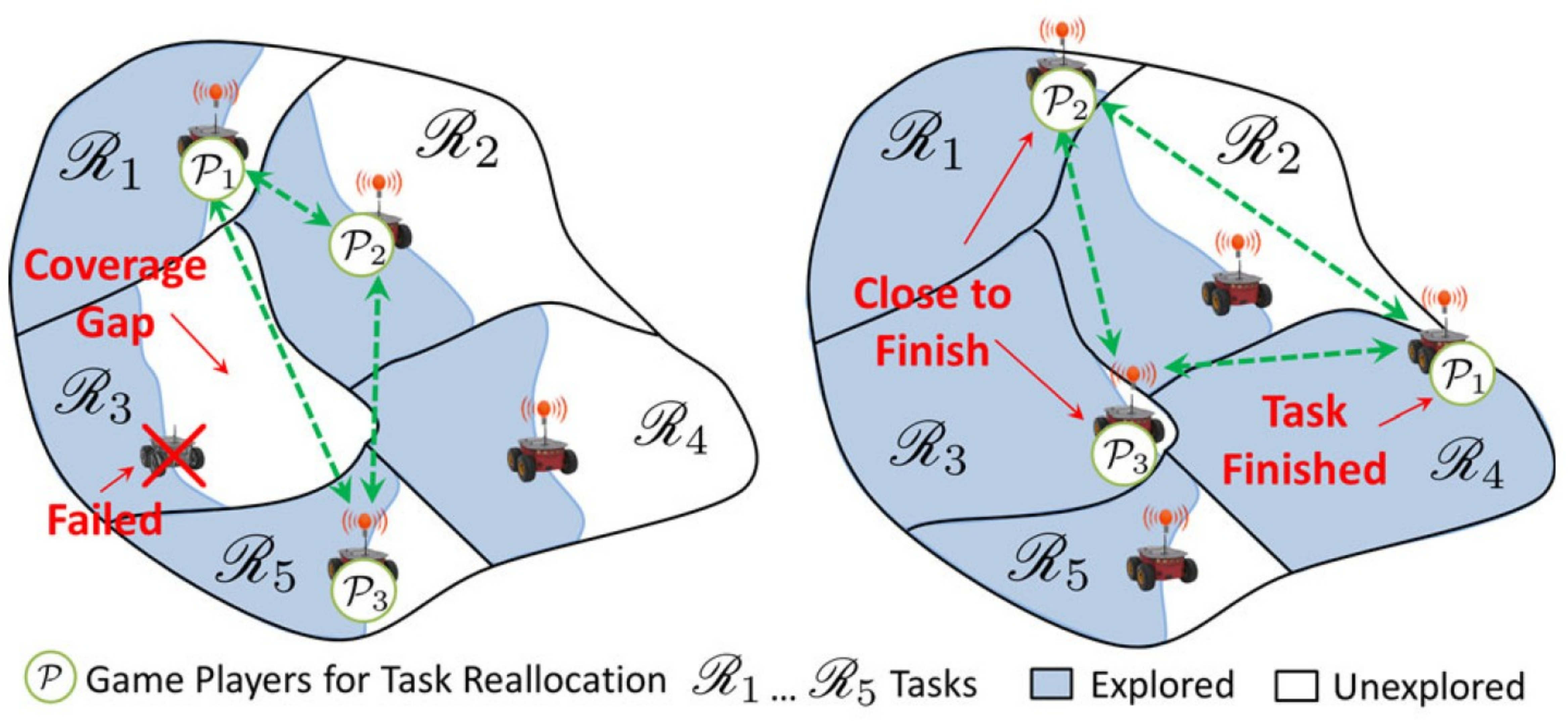}
    \caption{Concepts of resilience and efficiency in the CARE algorithm~\cite{song2020care}, a subarea-decomposition method. It reallocates unfinished coverage tasks to a team of robots when a robot fails or becomes idle.}\label{fig:song2020care} 
    \vspace{-1em}
 \end{figure}

 \begin{table*}[t]{}
\footnotesize
\caption {Comparison of representative online multi-robot CPP methods.}\label{tab:multi_online_table}\vspace{-3pt}
\centering
\setlength\tabcolsep{3.5pt}
\begin{tabular}{l l l l l l l l} 
 \toprule
\specialrule{0.1em}{1pt}{1pt} 
\tabincell{l}{\textbf{Category}}
&\multicolumn{1}{l}{\textbf{Method}}
&\tabincell{c}{\textbf{Year}}
&\tabincell{l}{\textbf{Main Idea}} 
&\tabincell{l}{{\textbf{Subarea}}\\{\textbf{Generation}}} 
&\tabincell{l}{{\textbf{Inter-robot Conflict}}\\{\textbf{Handling}}}
&\tabincell{l}{\textbf{{Strength}}}
&\tabincell{l}{\textbf{Limitation}}\\ 
\toprule

\specialrule{0em}{1pt}{0pt}
\multirow{12}{*}{\tabincell{l}{\textbf{{Grid-based}}}}
& \tabincell{l}{\textbf{BoB}~\cite{viet2015bob}}  & 2015     &\tabincell{l}{{Selects target cells using}\\{a directional-priority rule}}  &\multirow{12}{*}{None} &\tabincell{l}{{Labels other robots'}\\{cells as obstacles}} &\multirow{12}{*}{\tabincell{l}{{Computationally}\\{efficient and}\\{easy to deploy}}} &\multirow{12}{*}{\tabincell{l}{{Prone to inter-}\\{robot conflicts}\\{and overlapping}\\{paths in cluttered}\\{environments}}} \vspace{0.5em}\\ 

& \tabincell{l}{\textbf{ND}~\cite{luo2017neural}} & 2017     &\tabincell{l}{{Uses activity landscape to}\\{guide target cell selection}}  & &\tabincell{l}{{Sets negative activity}\\{to other robots' cells}} & & \vspace{0.5em}\\

& \tabincell{l}{\textbf{DecPPCPP}~\cite{hassan2020dec}}       & 2020     &\tabincell{l}{{Uses a predator-distance-}\\{based cost function to}\\{determine target cell}} & &\tabincell{l}{{Treat other robots}\\{as dynamic predators}} & & \vspace{0.5em}\\

& \tabincell{l}{\textbf{APFCPP}~\cite{wang2024apf}}      & 2024     &\tabincell{l}{{Prioritizes neighboring}\\{cells with smallest angle}\\{to total repulsive force}} & &\tabincell{l}{{Uses repulsive forces}\\{from other robots for}\\{spatial separation}} & & \vspace{0.5em}\\

& \tabincell{l}{\textbf{GAMRCPP}~\cite{mitra2025online}}      & 2025     &\tabincell{l}{{Centrally assigns target}\\{cells and plans collision-}\\{free transition paths}} & &\tabincell{l}{{Treats other robots}\\{as obstacles}} & & \\

\specialrule{0.05em}{2pt}{2pt}
\multirow{16}{*}{\tabincell{l}{\textbf{{Subarea-}}\\\textbf{decomposition}}}
& \tabincell{l}{\textbf{CARE}~\cite{song2020care}}     & 2020     &\tabincell{l}{{Reassigns unfinished}\\{subareas when a robot}\\{fails or becomes idle}} &\tabincell{l}{{Offline partitions}\\{the environment}\\{into regions}} &\multirow{16}{*}{\tabincell{l}{{Assigns each robot}\\{a unique set of}\\{subareas to promote}\\{spatial separation}}} &\multirow{16}{*}{\tabincell{l}{{Enables global}\\{coordination and}\\{reduces inter-}\\{robot conflicts}}} &\multirow{16}{*}{\tabincell{l}{{Performance}\\{depends on}\\{partitioning}\\{quality}}} \vspace{0.5em}\\

& \tabincell{l}{\textbf{CCIBA$^*$}~\cite{ma2022cciba}}     & 2022     &\tabincell{l}{{Reassigns local regions}\\{among nearby robots}}   &\tabincell{l}{{Offline capacity-}\\{aware}} & & &\vspace{0.5em}\\

& \tabincell{l}{\textbf{MAC}~\cite{wang2025mac}}     & 2025     &\tabincell{l}{{Assigns subareas via}\\{k-means clustering and}\\{pairwise optimization}}    &\tabincell{l}{{Online workload-}\\{aware}} & & &\vspace{0.5em}\\

& \tabincell{l}{\textbf{ABE}~\cite{cao2025complete}}     & 2025     &\tabincell{l}{{Adjusts subarea size}\\{according to robots'}\\{residual capacities}}    &\tabincell{l}{{Online residual}\\{capacity-aware}} & & &\vspace{0.5em}\\

& \tabincell{l}{\textbf{Multi-CAP}~\cite{shen2025multi}}     & 2025     &\tabincell{l}{{Incrementally partitions}\\{the space into connected}\\{subareas and updates}\\{subarea traversal tours}\\{to guide local coverage}}    &\tabincell{l}{{Online}\\{connectivity-}\\{aware}} & & &\\

\bottomrule
\specialrule{0.1em}{1pt}{1pt}
\end{tabular}
\vspace{-1.0em}
\end{table*}

Several grid-based methods introduced centralized short-term target cell assignment. Mitra and Saha~\cite{mitra2022scalable} proposed the goal assignment-based multi-robot CPP (GAMRCPP) algorithm, which centrally assigns uncovered cells to robots within a planning horizon and generates collision-free paths to cover them. This method was further improved to reduce unnecessary replanning and robot idling time~\cite{mitra2024online,mitra2025online}.

\textit{Strengths and Limitations:} Grid-based methods are computationally efficient and easy to implement. However, since they allow all robots to operate over the entire space, they are more prone to inter-robot conflicts and overlapping paths in cluttered environments. This motivates subarea-decomposition methods, which partition the environment and assign robots to distinct regions for coverage.

\subsubsection{Subarea-decomposition methods} 
These methods partition the environment into multiple non-uniform subareas and coordinate robots at the subarea level. Each robot is assigned one or more subareas and performs local coverage within them. This hierarchical structure enables more explicit workload allocation and helps reduce inter-robot interference.

One representative direction focuses on resilient and efficient task reallocations. Song and Gupta~\cite{song2020care} proposed the cooperative autonomy for resilience and efficiency (CARE) algorithm, where the concepts of team resilience and efficiency are depicted in Fig.~\ref{fig:song2020care}. Each robot covers its assigned subarea using the $\varepsilon^*$ algorithm~\cite{song2018}. When a robot fails or becomes idle, task reallocation is performed using distributed supervisors and potential games, allowing feasible robots to take over unfinished subareas. Ma et al.~\cite{ma2022cciba} proposed the collaborative coverage improved BA$^*$ (CCIBA$^*$) algorithm, where each robot performs BA$^*$-based coverage~\cite{viet2013ba} within its assigned subarea. During execution, a global coordinator trigger predefined collaborative behaviors, including area division, recall-and-transfer, and area exchange, so that obstacle-induced residual regions or inefficient local tasks can be reassigned among neighboring robots with shorter travel distance.

Another direction emphasizes workload-aware partitioning. Wang et al.~\cite{wang2025mac} proposed the multi-agent cooperation (MAC) algorithm, which constructs the smallest rectangular region enclosing all uncovered cells, recursively partitions it into subareas, and obtains an initial allocation via K-means clustering. Then, pairwise optimization among nearby robots is done to balance workload. This method was later extended to multi-robot target coverage by maintaining a Delaunay graph over robot positions and target points for workload-balanced task allocation~\cite{wang2026dmt}. Cao et al.~\cite{cao2025complete} proposed the adaptive balance evolution (ABE) algorithm, which adjusts subarea sizes according to the robots' residual capacities. 

\begin{figure}[t]
        \centering        \includegraphics[width=0.47\textwidth]{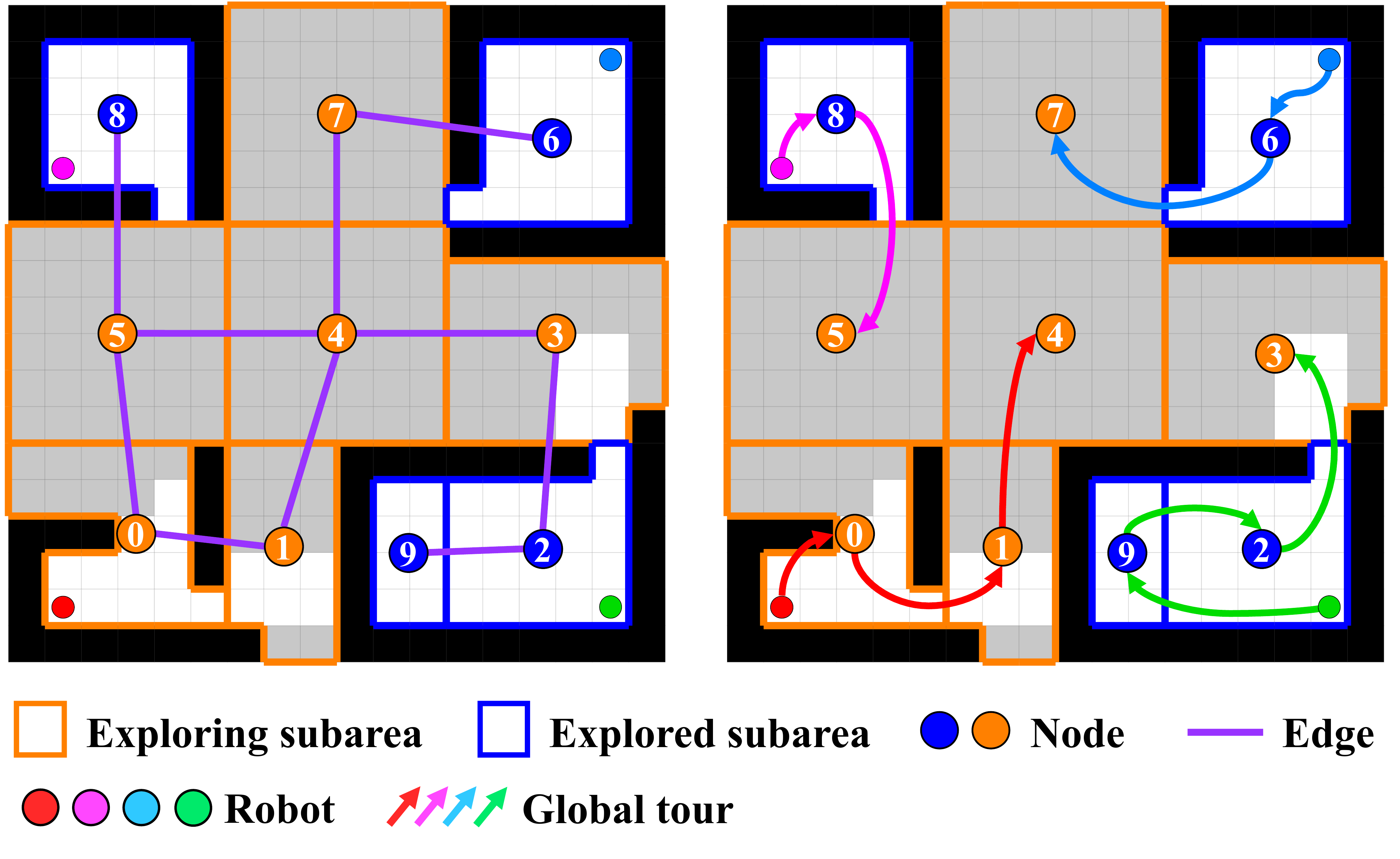}
    \caption{Illustration of the subarea-decomposition method Multi-CAP~\cite{shen2025multi}. It incrementally partitions the environment into connected subareas, maintains their adjacency relationships in a graph, and computes disjoint global tours to guide local coverage.}\label{fig:shen2025multi} 
    \vspace{-1em}
 \end{figure}

Subarea-decomposition methods can also integrate global structural guidance with local online coverage. Shen et al.~\cite{shen2025multi} proposed the multi-robot connectivity-aware hierarchical coverage path planning (Multi-CAP) algorithm, as shown in Fig.~\ref{fig:shen2025multi}. The method incrementally partitions the environment into connected subareas and maintains their adjacency graph as new obstacles are discovered. At the global level, subarea assignment is formulated as an open multi-depot vehicle routing problem to compute disjoint robot tours, thereby reducing redundant travel and inter-robot conflicts. At the local level, each robot adapts its coverage strategy according to subarea observation status, using back-and-forth coverage in exploring subareas and shortest coverage paths in explored subareas.

\textit{Strengths and Limitations:} Subarea-decomposition methods use region-level coordination to improve workload management, reduce inter-robot conflicts, and support task reassignment or global tour updates as new areas are discovered. However, their performance depends on partitioning and reallocation quality, and they usually require more computation and communication than grid-based methods.

\section{3D CPP}
\label{sec:threeDCPP}

3D CPP extends 2D CPP to workspaces or surfaces with complex 3D geometry, such as underwater terrains, mountainous environments, high-rise structures, and object surfaces. Compared with 2D CPP, it must account for surface geometry, elevation variation, sensing or viewpoint constraints, energy cost, and, in manipulator applications, kinematic feasibility. 

Existing methods can be broadly grouped by how they represent and exploit the  3D geometry. Layered-coverage methods reduce 3D CPP to a sequence of 2D coverage tasks over planar layers or subregions. Contour-aware methods exploit terrain contours or elevation structures to generate terrain-conforming paths with reduced vertical motion. Decomposition-based methods partition 3D surfaces into smaller patches according to geometric, topological, or kinematic properties, followed by patch sequencing and local coverage. Table~\ref{tab:threeD_cpp_table} summarizes the key features of representative 3D CPP methods.
 
\subsection{Layered-coverage Methods} 

Layered-coverage methods simplify 3D CPP by decomposing the environment into 2D layers (or planar regions), allowing the existing 2D CPP strategies to be used on these layers. The main challenge is to order, refine, and connect these layers while preserving inter-layer connectivity and handling disconnected regions.

Sadat et al.~\cite{sadat2014recursive} proposed an online method for aerial coverage of unknown terrains, where the environment is assumed to be obstacle-free and represented as a coverage tree. Upper-level nodes correspond to coarse coverage at higher altitudes, while lower-level nodes correspond to finer coverage at lower altitudes. The robot first covers the area coarsely and then recursively descends to cover subregions identified as regions of interest. This method was improved in~\cite{sadat2015fractal} using Hilbert space filling curves to reduce the travel cost.

Jung et al.~\cite{jung2018multi} proposed an offline method for high-rise structure coverage. The method divides the target structure into height-based layers, generates candidate viewpoints from surface normals at voxel centers, and downsamples them using a voxel grid filter. A local inspection path is computed by solving a TSP over selected viewpoints. To reduce redundant inspection, viewpoints in the next layer are updated by removing duplicated observed surfaces, and all layer-wise paths are connected into a full coverage path.

\begin{figure}[t]
        \centering        \includegraphics[width=0.45\textwidth]{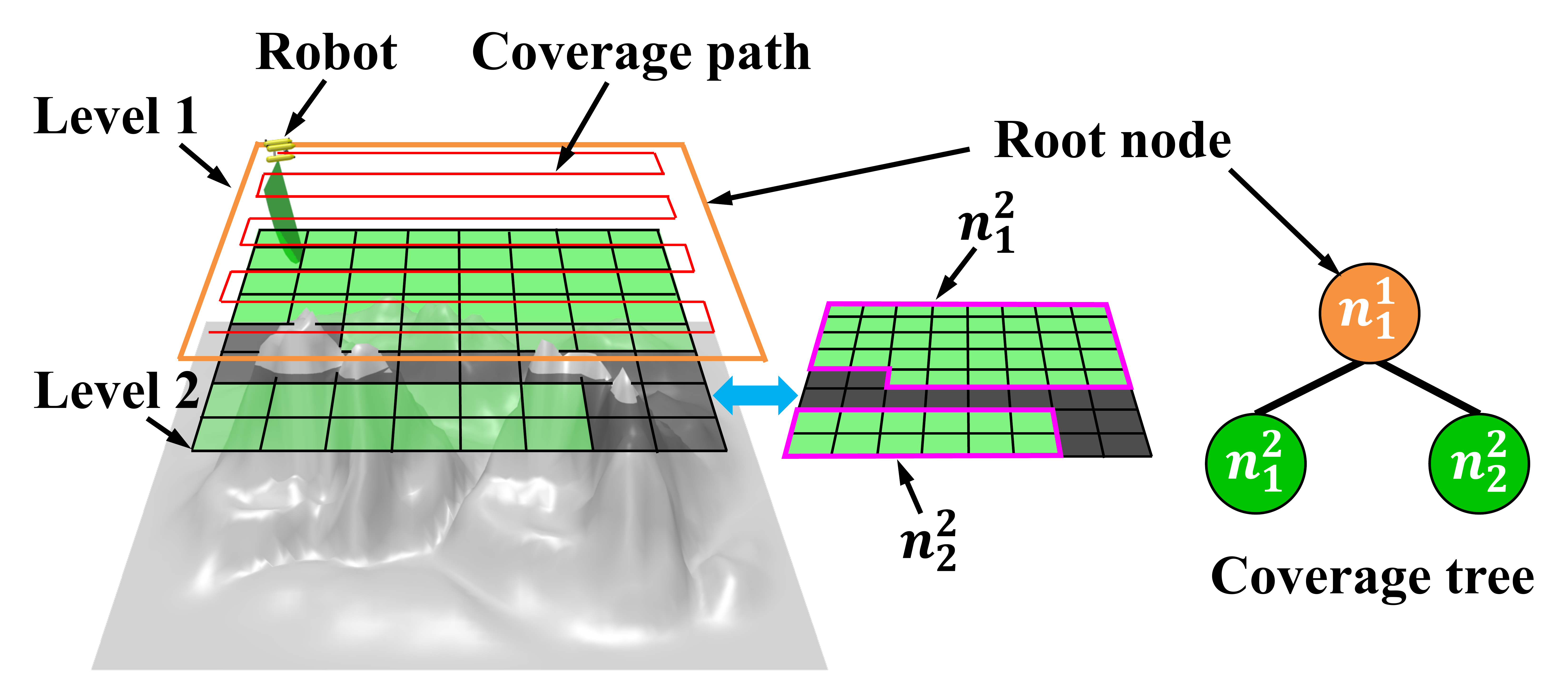}
    \caption{Illustration of the layered-coverage method CT-CPP~\cite{shen2022ct}. It performs back-and-forth coverage within each planar subregion, projects detected terrain structures onto lower levels to identify disconnected subregions, and adds them to the coverage tree.}\label{fig:shen2022ct} 
    \vspace{-1em}
 \end{figure}

Shen et al.~\cite{shen2016autonomous,shen2017,shen2022ct} proposed CT-CPP, which performs layered coverage in unknown obstacle-rich environments using a dynamic coverage tree, as shown in Fig.~\ref{fig:shen2022ct}. The robot starts from the ocean surface and incrementally builds the tree in a top-down manner, where each node represents a disconnected planar subregion discovered at a specific level. During the coverage of each node, the robot collects data from the 3D terrain structures below and projects the obstacle information onto the plane below to form a 2D occupancy map. Then, the newly detected disconnected subregions are added to that node as child nodes. The updated coverage tree is then used to determine the subregion traversal sequence by solving a TSP. 

\textit{Strengths and Limitations:} Layered-coverage methods reduce the complexity of 3D CPP by leveraging 2D CPP within individual layers; however, their performance depends on the layer construction and inter-layer transition strategy.

\subsection{Contour-aware Methods}

Contour-aware methods are mainly designed for terrains with significant elevation variations. Instead of treating the 3D workspace as independent layers, these methods exploit terrain contours, elevation gradients, or geodesic structures to generate paths that better conform to the surface and reduce energy-consuming vertical motions.

Galceran et al.~\cite{galceran2015coverage} classified the terrain into planar and steep-slope areas using a bathymetric map. The method then plans back-and-forth coverage paths for planar areas using boustrophedon decomposition~\cite{choset2000coverage} while treating steep-slope areas as obstacles. For steep-slope areas, the terrain surface is sliced by horizontal planes at successive depth levels, with spacing determined by the coverage footprint, to obtain contour lines. These contours are shifted outward to maintain a safe distance and then linked to form the final coverage path.

Wu et al.~\cite{wu2019energy} generated a contour-aware spiral coverage path for mountainous terrain using geodesic distance fields. The method defines a geodesic source set on the terrain surface, consisting of the surface boundary and the peak and valley points. An exact geodesic distance field is then constructed from these sources, from which iso-contours are extracted and connected into a spiral coverage path. By incorporating terrain height extrema into the source set, the generated path tends to follow regions of similar elevation, thereby reducing uphill motion and energy cost.

    \begin{figure}[t]
        \centering        \includegraphics[width=0.49\textwidth]{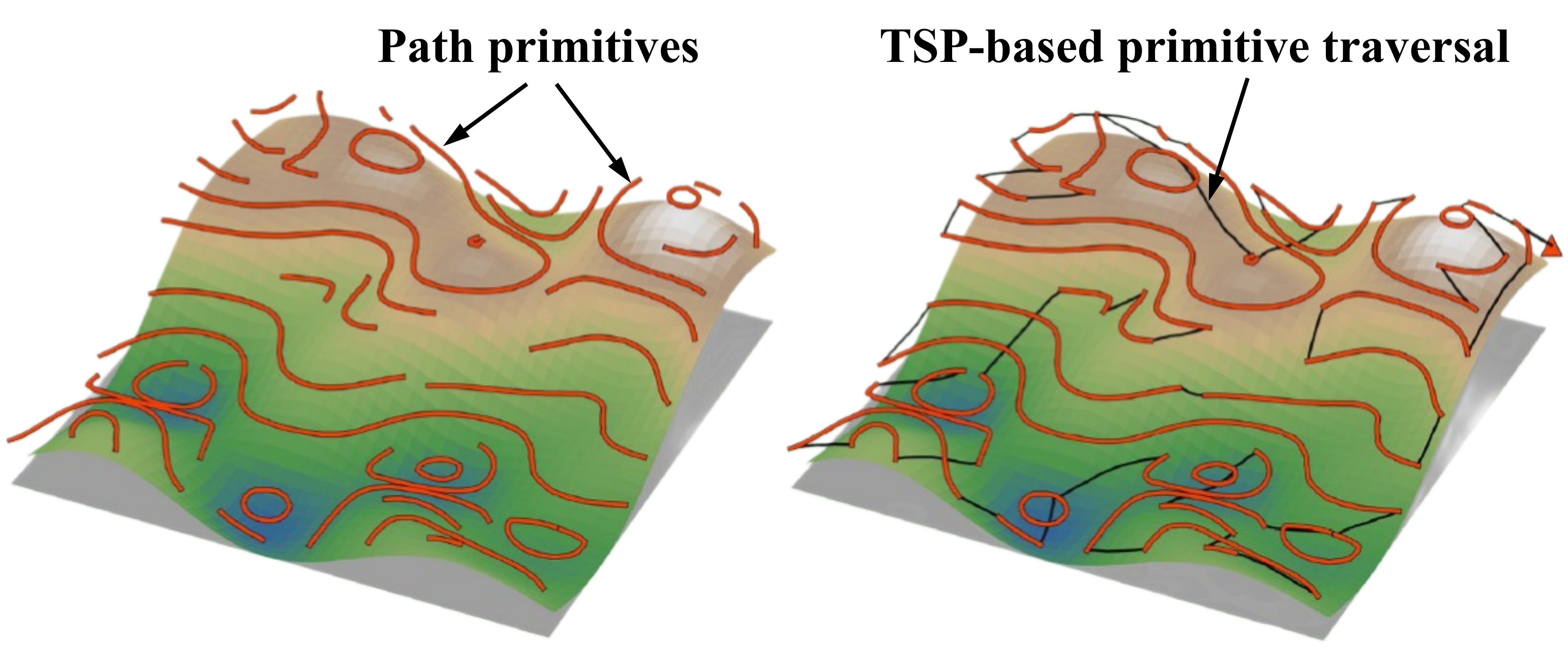}
    \caption{Illustration of the contour-aware method in~\cite{shao2025energy}. It first generates path primitives along terrain contours and then optimizes their traversal order to obtain the full coverage path.}\label{fig:shao2025energy} 
    \vspace{-1em}
 \end{figure}

Shao et al.~\cite{shao2025energy}  proposed contour-aware method via path primitives, as shown in Fig.~\ref{fig:shao2025energy}. It first extracts quantized elevation contours from the terrain surface. Seed points are then sampled from uncovered regions and grown bidirectionally along nearby contours to generate contour-aware path primitives. These primitives are connected into a complete coverage path by solving a TSP, where the transition cost accounts for the higher energy cost of climbs than descents.

\textit{Strengths and Limitations:} Contour-aware methods exploit terrain elevation structures by generating paths that conform to terrain contours. This helps reduce energy-intensive vertical motion and promotes more uniform surface coverage, which is particularly useful for mapping and terrain reconstruction. However, these methods usually require reliable terrain models, accurate contour extraction, and precise control, which limits their applicability in unknown environments.

\begin{figure}[t]
        \centering        \includegraphics[width=0.4\textwidth]{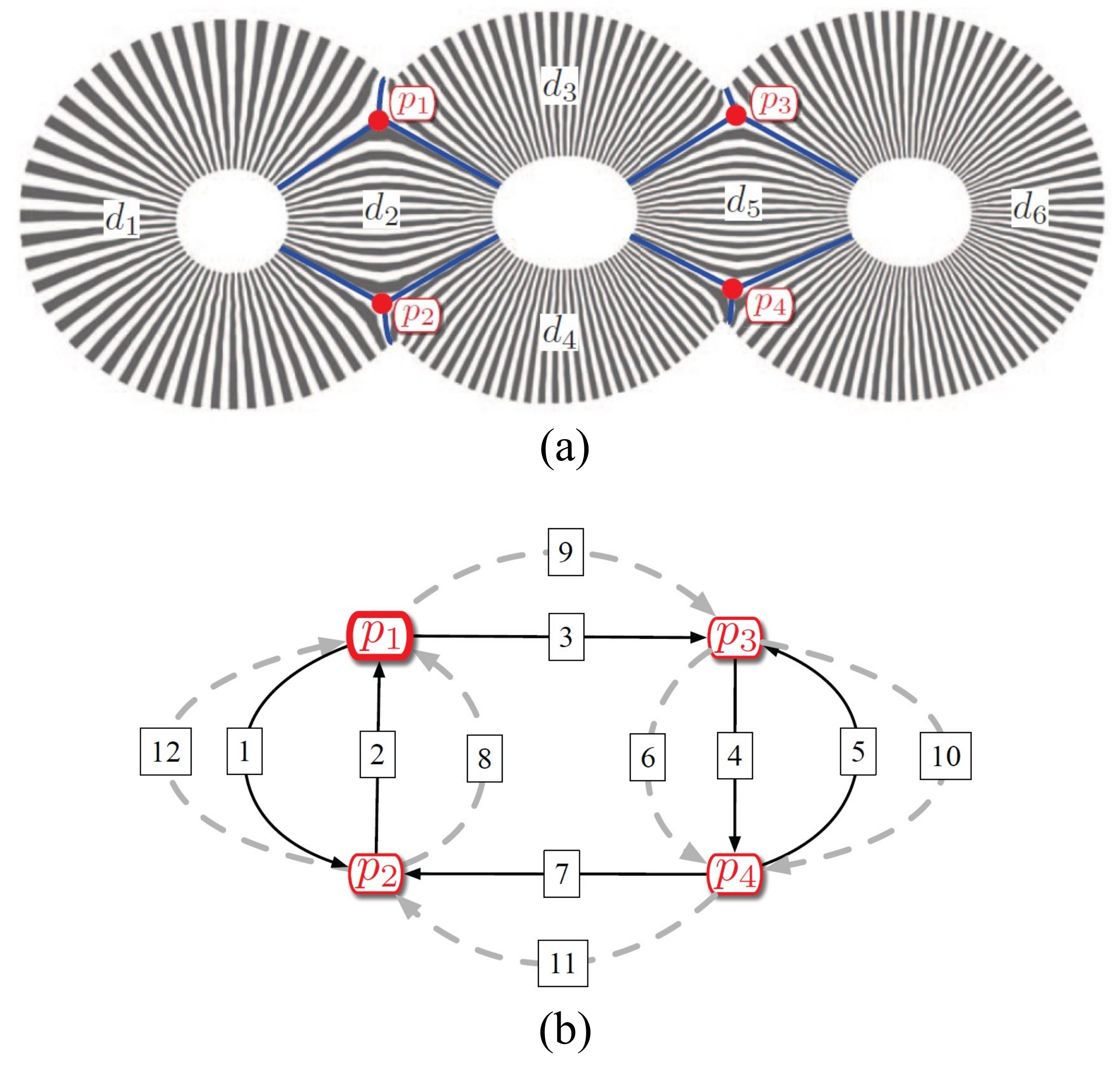}
    \caption{Illustration of the decomposition-based method in~\cite{lin2017}. (a) A three-hole surface is decomposed into connected surface patches $d_1$-$d_6$ by critical trajectories connecting zero points $p_1$-$p_4$. (b) An Euler cycle is computed over the patches, starting from $p_1$, following the numbered directed edges, and finally returning to $p_1$.}\label{fig:lin2017} 
    \vspace{-1em}
 \end{figure}

\begin{table*}[t]{}
\footnotesize
\caption {Comparison of representative 3D CPP methods. }\label{tab:threeD_cpp_table}\vspace{-3pt}
\centering
\setlength\tabcolsep{3.0pt}
\begin{tabular}{l l l l l l l l} 
 \toprule
\specialrule{0.1em}{1pt}{1pt} 
\tabincell{l}{\textbf{Category}}
&\multicolumn{1}{l}{\textbf{Method}}
&\tabincell{c}{\textbf{Year}}
&\tabincell{l}{\textbf{Main Idea}} 
&\tabincell{l}{{\textbf{Geometry}}\\{\textbf{Representation}}} 
&\tabincell{l}{{\textbf{Coverage Path}}\\{\textbf{Generation}}}
&\tabincell{l}{\textbf{{Strength}}}
&\tabincell{l}{\textbf{Limitation}}\\ 
\toprule

\specialrule{0em}{1pt}{0pt}
\multirow{7}{*}{\tabincell{l}{\textbf{{Layered-}}\\\textbf{coverage}}}
& \tabincell{l}{\textbf{Sadat et al.}~\cite{sadat2014recursive}}     & 2014     &\tabincell{l}{{Recursive coarse-}\\{to-fine ROI coverage}} &\tabincell{l}{{Multi-resolution}\\{altitude layers}} &\tabincell{l}{{Recursive descent}\\{to selected ROIs}} &\multirow{7}{*}{\tabincell{l}{{Converts}\\{complex}\\{3D CPP}\\{into simple}\\{2D CPPs}}} &\multirow{7}{*}{\tabincell{l}{{Performance}\\{depends on}\\{the layer design}}} \vspace{0.5em}\\

& \tabincell{l}{\textbf{Jung et al.}~\cite{jung2018multi}}     & 2018     &\tabincell{l}{{Layer-wise high-rise}\\{inspection}}   &\tabincell{l}{{Viewpoints created}\\{at different layers}} &\tabincell{l}{{TSP-based viewpoint}\\{sequencing}} & &\vspace{0.5em}\\

& \tabincell{l}{\textbf{CT-CPP}~\cite{shen2022ct}}     & 2022     &\tabincell{l}{{Constructs a dynamic}\\{tree to track and guide}\\{online coverage process}}    &\tabincell{l}{{Disjoint subregions}\\{at different levels}\\{organized as a tree}} &\tabincell{l}{{TSP-based subregions}\\{visiting and local back}\\{-and-forth coverage}} & &\\

\specialrule{0.05em}{2pt}{2pt}
\multirow{9}{*}{\tabincell{l}{\textbf{{Contour-}}\\\textbf{aware}}}
& \tabincell{l}{\textbf{Galceran et al.}~\cite{galceran2015coverage}}     & 2015     &\tabincell{l}{{Adapts coverage pattern}\\{by terrain type}} &\tabincell{l}{{Planar and steep}\\{areas}} &\tabincell{l}{{Back-and-forth paths in}\\{planar areas and contour}\\{following in steep areas}} &\multirow{9}{*}{\tabincell{l}{{Generates}\\{terrain-}\\{conforming}\\{paths with}\\{reduced}\\{energy-}\\{intensive}\\{motion}}} &\multirow{9}{*}{\tabincell{l}{{Requires reliable}\\{terrain models,}\\{accurate contour}\\{extraction, and}\\{precise control}}} \vspace{0.5em}\\

& \tabincell{l}{\textbf{Wu et al.}~\cite{wu2019energy}}     & 2019     &\tabincell{l}{{Uses geodesic distance}\\{fields for energy-}\\{efficient coverage}}    &\tabincell{l}{{Geodesic distance}\\{field}} &\tabincell{l}{{Extracts iso-contours}\\{and connects them into}\\{a full coverage path}} & &\vspace{0.5em}\\

& \tabincell{l}{\textbf{Shao et al.}~\cite{shao2025energy}}     & 2025     &\tabincell{l}{{Generates and sequences}\\{contour-aware path}\\{primitives for energy-}\\{efficient coverage}}    &\tabincell{l}{{Quantized elevation}\\{contours}} &\tabincell{l}{{TSP-based traversal of}\\{contour-aware path}\\{primitives}} & &\\

\specialrule{0.05em}{2pt}{2pt}
\multirow{10}{*}{\tabincell{l}{\textbf{{Decom-}}\\\textbf{position}\\\textbf{-based}}}
& \tabincell{l}{\textbf{Lin et al.}~\cite{lin2017}}     & 2017     &\tabincell{l}{{Decomposes surfaces by}\\{critical trajectories and}\\{generates patch-wise}\\{coverage paths}} &\tabincell{l}{{Intrinsic coordinate}\\{system from holo-}\\{morphic quadratic}\\{differentials}} &\tabincell{l}{{Regular-trajectory}\\{tracing with Euler}\\{-cycle sequencing}} &\multirow{10}{*}{\tabincell{l}{{Directly}\\{handles}\\{complex}\\{3D surfaces}\\{via patch-}\\{wise planning}\\{with kinematic}\\{constraints}}} &\multirow{10}{*}{\tabincell{l}{{Requires reliable}\\{surface models,}\\{and mesh}\\{representations}}} \vspace{0.5em}\\

& \tabincell{l}{\textbf{Yang et al.}~\cite{yang2020cellular}}     & 2020     &\tabincell{l}{{Decomposes surfaces into}\\{kinematically continuous}\\{patches to reduce end-}\\{effector lift-offs}}    &\tabincell{l}{{Task-space labels}\\{induced by reachable}\\{IK configuration sets}} &\tabincell{l}{{Patch-wise path}\\{generation in IK-}\\{continuous patches}} & &\vspace{0.5em}\\

& \tabincell{l}{\textbf{Wang et al.}~\cite{wang2025hierarchically}}     & 2025     &\tabincell{l}{{Hierarchically clusters}\\{surface points to reduce}\\{computational time}}    &\tabincell{l}{{Position-normal}\\{surface clusters with}\\{IK-feasible exemplars}} &\tabincell{l}{{Upper-level exemplar}\\{TSP with lower-level}\\{joint-space TSP}} & &\\

\bottomrule
\specialrule{0.1em}{1pt}{1pt}
\end{tabular}
\vspace{-1.0em}
\end{table*}

\subsection{Decomposition-based Methods}

Decomposition-based methods operate directly on 3D surfaces by partitioning them into smaller patches for sequencing and local coverage path generation.

One line of work decomposes surfaces using intrinsic geometric structures. As shown in Fig.~\ref{fig:lin2017}, Lin et al.~\cite{lin2017} built an intrinsic surface parameterization based on holomorphic quadratic differentials. This parameterization assigns natural coordinates to surface points and induce a structured trajectory field, where trajectories ending at zero points are treated as critical trajectories and used to partition the surface into patches. Within each patch, regular trajectories are traced and connected into back-and-forth coverage paths, and the patch traversal order is organized via an Euler-cycle formulation.

Another line of work incorporates manipulator kinematics into surface decomposition. A continuous task-space coverage path may correspond to disconnected inverse-kinematics configuration sets in the manipulator joint space, causing end-effector lift-offs. To reduce such lift-offs, Yang et al.~\cite{yang2020cellular} decomposed the surface according to joint-space continuity. For each surface point, the method computes non-singular inverse-kinematics configurations and groups them into disjoint sets according to their continuity in joint space. The resulting labels are mapped back to the task space, where surface points sharing the same reachable configuration labels are grouped into the same surface patch. Each patch can be covered without joint-space discontinuities. This framework was later extended to jointly optimize object placements and covered surface portions~\cite{yang2021optimal}, use valid singular configurations to connect otherwise disconnected nonsingular sets~\cite{yang2024optimal}, and develop a more efficient decomposition-refinement solver while preserving global optimality and completeness~\cite{yang2024improved}.

Surface decomposition can also be achieved through mesh discretization or geometric clustering. Yang et al.~\cite{yang2023template} discretized the surface into a uniform unstructured mesh with a resolution determined by the coverage tool size. The mesh is refined by edge subdivision, and coverage waypoints are generated over the refined patches, mapped back to the original surface, and locally optimized to reduce overlap, gaps, and path irregularity. Do and Pham~\cite{do2023geometry} partitioned the mesh into connected low-curvature clusters using constrained centroidal Voronoi tessellation~\cite{nguyen2009constrained}. Cluster centroids and proxy normals are used to define viewpoints, and the coverage path is obtained by solving a TSP over the centroids with geodesic edge costs. Wang and Gleicher~\cite{wang2025hierarchically} proposed a hierarchical method that reduces computation by decomposing the problem into smaller generalized TSPs. It first clusters surface points by positional and normal similarity and selects representative exemplars. For each exemplar, feasible inverse-kinematics solutions that can reach all points in the same cluster are retained. An upper-level TSP over the exemplars then provides a guide path, followed by a lower-level TSP that generates the complete joint-space coverage path.

\textit{Strengths and Limitations:} Decomposition-based methods directly handle complex 3D surfaces, especially when geometric continuity, surface curvature, viewpoint quality, or manipulator feasibility must be considered. Their main limitation is the reliance on accurate surface models, mesh representations, or inverse-kinematics analysis, which increases computational complexity and limits applicability in unknown environments.

\section{Constrained CPP}

\label{sec:robotconstrainedcpp}

\begin{figure}[t]
    \centering
    \subfloat[Energy-constrained robot]{
    \includegraphics[width=0.23\textwidth]{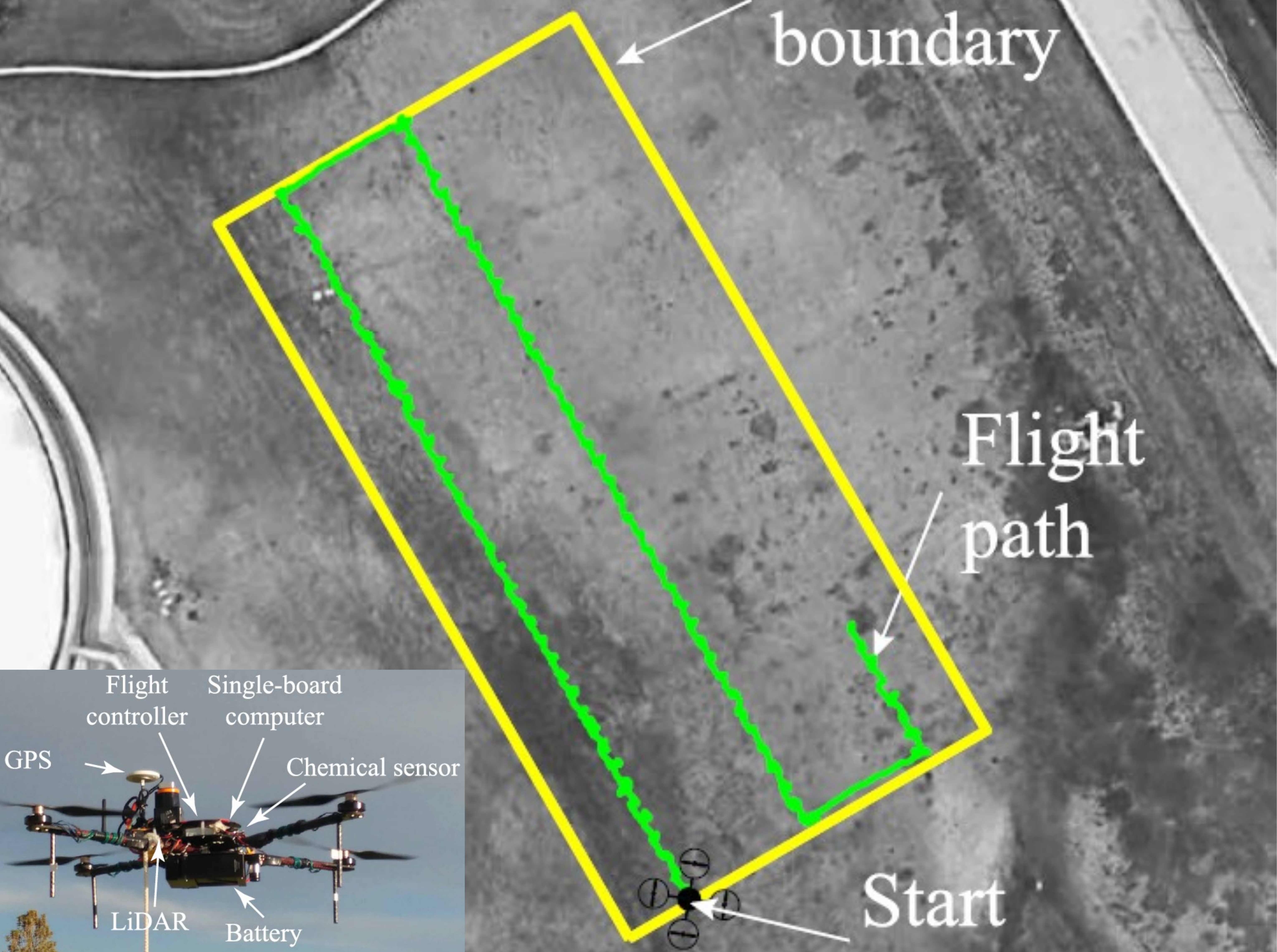}\label{fig:constrainedrobotexample_part1}}\hspace{-10pt}\quad
    \centering
    \subfloat[Curvature-constrained robot]{
    \includegraphics[width=0.23\textwidth]{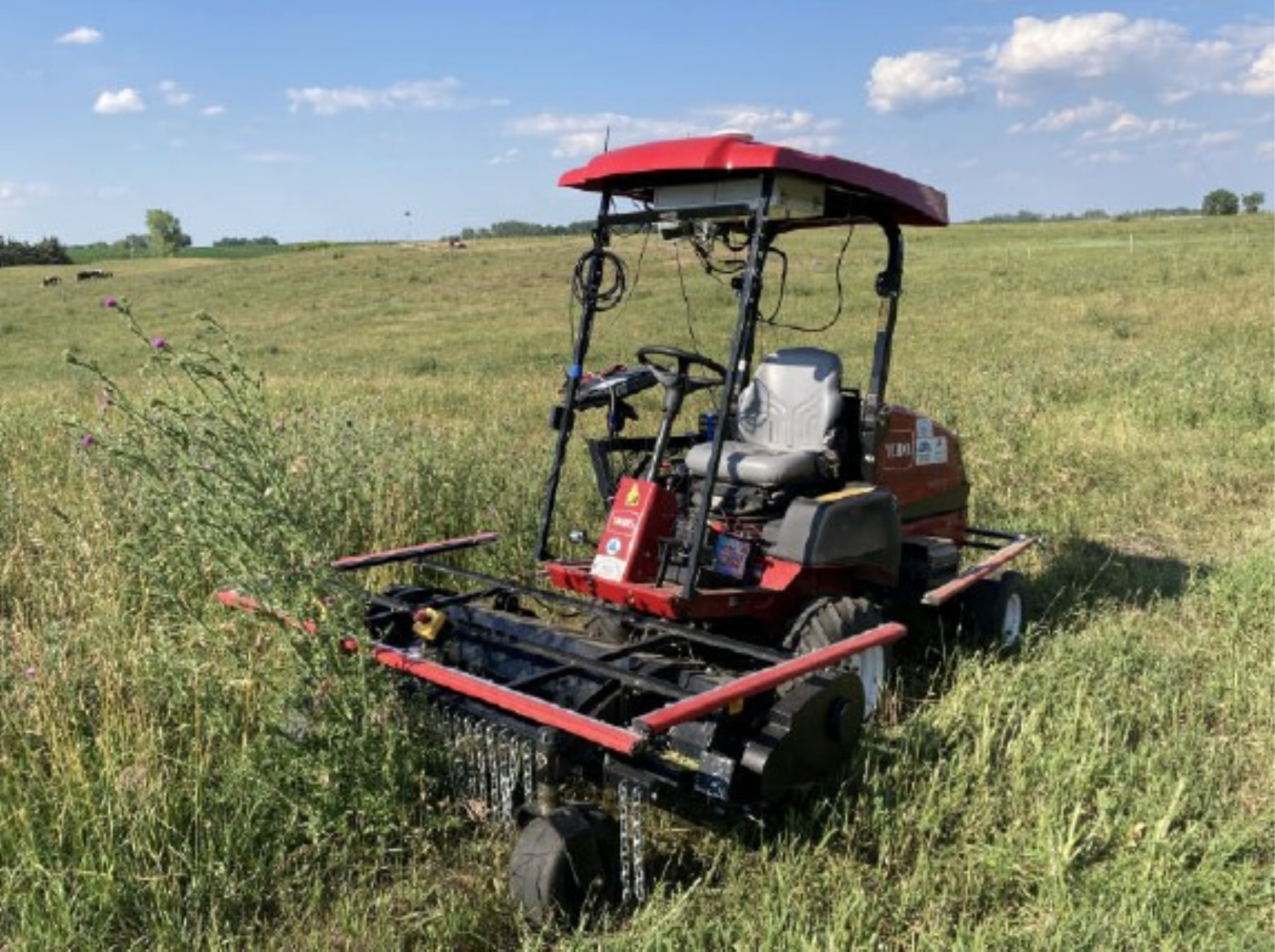}\label{fig:constrainedrobotexample_part2}}\vspace{0.5em}\\
    \centering
    \subfloat[Tethered robot]{
    \includegraphics[width=0.23\textwidth]{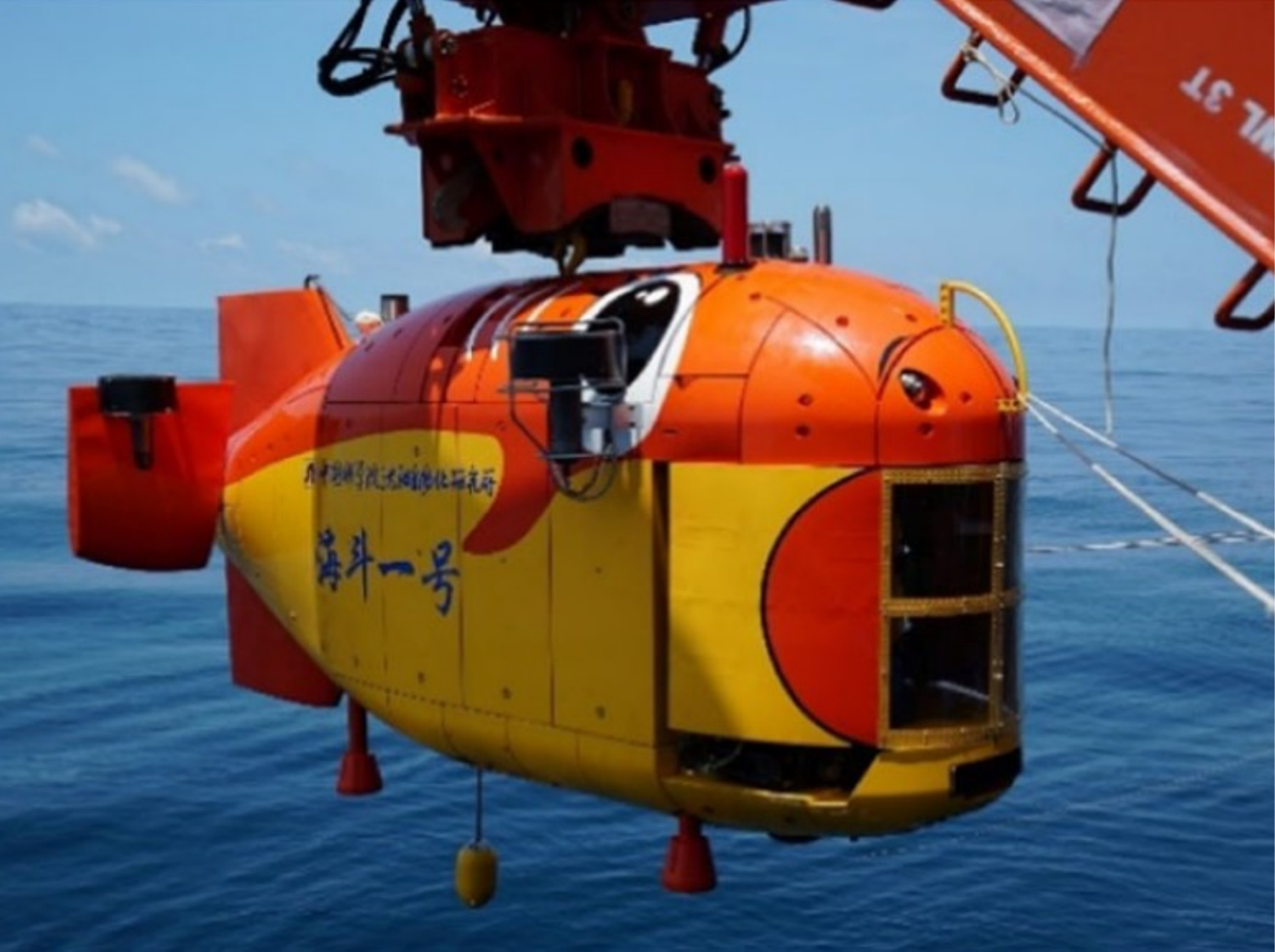}\label{fig:constrainedrobotexample_part3}}\hspace{-10pt}\quad
    \centering
    \subfloat[Reconfigurable robot]{
    \includegraphics[width=0.23\textwidth]{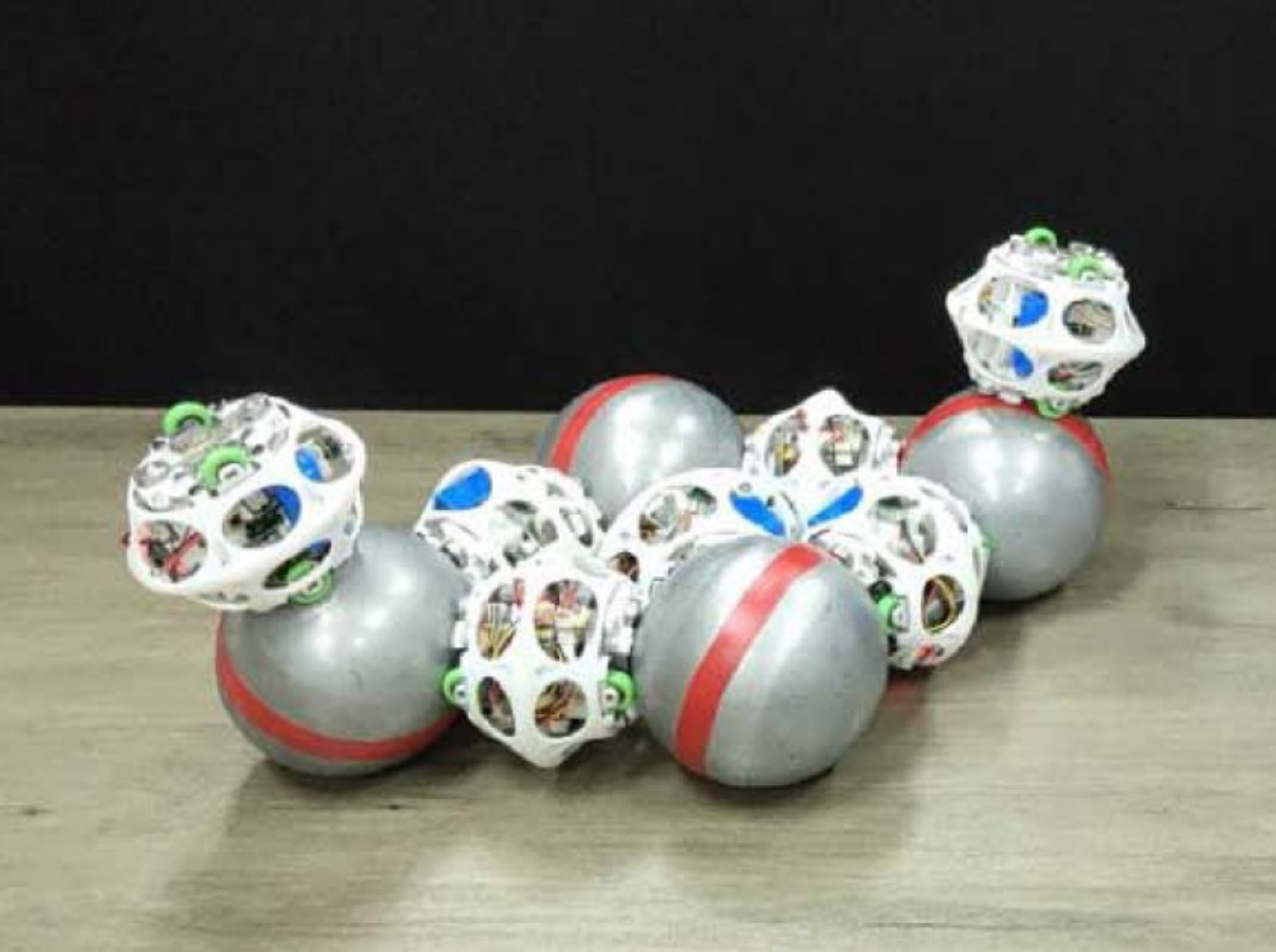}\label{fig:constrainedrobotexample_part4}}\\
          \caption{Robot examples with different platform-specific constraints: (a) aerial robot performing partial coverage under a limited battery budget~\cite{jensen2020near}, (b) autonomous weed-mowing robot with bounded-curvature motion~\cite{maini2022online}, (c) tethered underwater robot using a tethered cable for remote operation and communication~\cite{wang2024haidou}, and (d) reconfigurable robot with morphological restrictions~\cite{tu2023configuration}.}\label{fig:constrainedrobotexample}
          \vspace{-1em}
\end{figure}

Beyond environmental knowledge, workspace geometry, and robot-team size, CPP also depends on robot-specific constraints. As shown in Fig.~\ref{fig:constrainedrobotexample}, such constraints may arise from battery capacities, bounded-curvatures, tethered cable limits, and morphological restrictions. These constraints restrict feasible motions, require safe return or battery replenishment strategies, or require the robot to adapt its shape or footprint to different workspace conditions. This section reviews four classes of constrained CPP: energy-constrained CPP, curvature-constrained CPP, tethered CPP, and reconfigurable CPP. Table~\ref{tab:robotconstrainedcpp} summarizes representative constraints, planning implications, CPP adaptations, benefits, and limitations.
 
\subsection{Energy-constrained CPP}
\label{sec:energyconstrainedrobot}

Energy-constrained robots operate with limited battery capacity, which restricts the duration and spatial extent of coverage operation. Thus, energy-constrained CPP methods must not only generate efficient coverage paths, but also ensure that the robot can return to a charging location or coordinate with replenishment resources (e.g., mobile charging robots) when its energy becomes low. Existing methods mainly address three settings: complete coverage with return-and-recharge scheduling, partial coverage optimization under a fixed energy budget, and coverage coordinated with mobile replenishment.

\subsubsection{Return-and-recharge methods}

These methods achieve complete coverage by repeatedly alternating between coverage path execution and returning to the base station for recharging. The key issue is not only to decide which region to cover after each recharging, but also to ensure that every coverage decision preserves enough residual energy for a safe return to the charging station. Existing methods mainly differ in the coverage pattern and how they enforce return decision.

One line of work uses the distance-to-base information  to decide whether to return to the charge station during the coverage process. Shnaps and Rimon~\cite{shnaps2016online} adapted their shortest-path-potential framework for tethered coverage~\cite{shnaps2014online} of battery-limited robots. The robot follows equipotential contours in increasing potential order, covers higher-potential contours within each corridor, and returns to the base along the shortest known path when the battery becomes low. After recharging, it resumes from the uncovered corridor with the smallest estimated potential. Wei and Isler~\cite{wei2018coverage} similarly exploited distance-to-base information through equidistance polylines. Their from-far-and-near strategy advances the robot to the furthest uncovered cell, follows equidistance contours for coverage, and retreats to the base station when the remaining energy becomes low. After recharging, the robot resumes coverage from the previously recorded retreat point.

Shen et al.~\cite{shen2020} proposed the $\varepsilon^*$+ algorithm, which uses the $\varepsilon^*$ algorithm for coverage~\cite{song2018} and computes a retreat path from each candidate target cell to the base station using the A$^*$ search~\cite{hart1968formal}. If the remaining energy is insufficient to reach the target cell and return safely to the base station, the robot immediately retreats for recharging. After charging is complete, it resumes coverage from the nearest uncovered cell. Dogru and Marques~\cite{dogru2022eco} used boundary-following rules for both coverage and retreat. The robot follows the left contour of the covered or occupied area, stores the current path and adjacent strip-like uncovered area, and computes a retreat path to the base within this area. When energy becomes low, it retreats along this path while covering cells along the way.

\begin{table*}[t]
\centering
\caption{Summary of representative constrained CPP categories.}\vspace{-3pt}
\label{tab:robotconstrainedcpp}
\centering
\setlength\tabcolsep{4.5pt}
\begin{tabular}{l l l l l l}
\toprule
\specialrule{0.1em}{1pt}{1pt} 
\tabincell{l}{\textbf{Category}} 
&\tabincell{l}{\textbf{Robot Constraint}} 
&\tabincell{l}{\textbf{Planning Implication}} 
&\tabincell{l}{\textbf{CPP Adaptation}}
&\tabincell{l}{\textbf{{Strength}}}
&\tabincell{l}{\textbf{Limitation}}\\
\toprule

\tabincell{l}{\textbf{Energy-}\\\textbf{constrained}\\\textbf{CPP}} 
&\tabincell{l}{{Battery capacity,}\\{mission duration,}\\{recharge requirement}} 
&\tabincell{l}{{Respects return feasibility,}\\{energy budget, or partial}\\{coverage objective}} 
&\tabincell{l}{{Return-and-recharge;}\\{energy-budgeted partial}\\{coverage; mobile}\\{replenishment}} 
&\tabincell{l}{{Improves practical}\\{deployability under}\\{limited endurance}}
&\tabincell{l}{{Higher computational}\\{time caused by}\\{scheduling and routing}}\vspace{0.5em}\\

\specialrule{0em}{2pt}{2pt}
\tabincell{l}{\textbf{Curvature-}\\\textbf{constrained}\\\textbf{CPP}} 
&\tabincell{l}{{Minimum turning}\\{radius and heading}\\{continuity}} 
&\tabincell{l}{{Sharp turns and arbitrary}\\{sweep connections may}\\{be infeasible}} 
&\tabincell{l}{{Dubins/Reeds-Shepp}\\{path generation; turn-}\\{minimizing decomp-}\\{osition and sequencing}} 
&\tabincell{l}{{Produces executable}\\{paths for nonholo-}\\{nomic vehicles}}
&\tabincell{l}{{Increases path length}\\{and complicates de-}\\{composition, ordering,}\\{and boundary coverage}}\vspace{0.5em}\\

\specialrule{0em}{2pt}{2pt}
\tabincell{l}{\textbf{Tethered CPP}} 
&\tabincell{l}{{Cable length and}\\{topology}} 
&\tabincell{l}{{Remains cable-feasible}\\{and avoid entanglement}} 
&\tabincell{l}{{Distance-contour}\\{traversal, cable-}\\{feasible tree or}\\{graph search}} 
&\tabincell{l}{{Extends operation time,}\\{communication reliability}\\{and robot's safety}}
&\tabincell{l}{{Restricted by base}\\{location, obstacle}\\{topology, and cable}\\{configuration}}\vspace{0.5em}\\

\specialrule{0em}{2pt}{2pt}
\tabincell{l}{\textbf{Reconfigurable}\\\textbf{CPP}} 
&\tabincell{l}{{Variable footprint}\\{shape, or morphology}} 
&\tabincell{l}{{Considers when and how}\\{the robot changes its}\\{configuration during}\\{decision-making}} 
&\tabincell{l}{{Size/shape-aware target}\\{selection; reconfiguration-}\\{coupled coverage planning}} 
&\tabincell{l}{{Improves access to}\\{narrow regions and}\\{coverage efficiency}\\{in open areas}}
&\tabincell{l}{{Requires modeling}\\{reconfiguration cost,}\\{footprint-dependent}\\{reachability, and}\\{transition feasibility}}\\
\bottomrule
\specialrule{0.1em}{1pt}{1pt} 
\end{tabular}
\vspace{-1.0em}
\end{table*}

\subsubsection{Partial-coverage methods}

Instead of enforcing complete coverage, these methods optimize coverage quality under a fixed energy budget. Jensen-Nau et al.~\cite{jensen2020near} represented the path as a sequence of connected waypoints with the length constrained by the robot's energy budget. The waypoints are redistributed using a Voronoi-based mass-spring-damper model, where the path is optimized by minimizing the distance of points in the area to the nearest path waypoints. Unlike return-and-recharge methods, this formulation optimizes (partial) coverage quality given a path-length constraint rather than scheduling repeated recharging for complete coverage.

\subsubsection{Mobile {energy-}replenishment methods}

These methods incorporate mobile energy-replenishment resources into the coverage process. Instead of returning to a fixed base, robots coordinate with mobile stations or heterogeneous teammates for recharging. Karapetyan et al.~\cite{karapetyan2024ag} considered a heterogeneous UAV-UGV team, where both robots perform coverage and the UGV also serves as a mobile recharging station for the energy-constrained UAV. The method first computes coverage paths for both robots without considering energy constraints, partitions the UAV path into endurance-feasible coverage segments and the UGV path into corresponding rendezvous segments, and then uses bipartite graph matching to determine recharging rendezvous locations.

\textit{Strengths and Limitations:} Energy-constrained CPP supports long-duration missions. However, it introduces additional decisions on return timing, recharging location, and coverage resumption, and its performance may degrade under uncertain energy consumption, travel time, or rendezvous execution.

\subsection{Curvature-constrained CPP}
\label{sec:curvaturerobot}

Curvature-constrained robots have a minimum turning radius and their motion is often described by the Dubins model~\cite{dubins1957curves,wilson2025generalized}, which consists of straight-line segments and circular arcs with bounded curvature. Thus, the straight-line sweep patterns with sharp turns must be modified to ensure curvature feasibility. Existing methods address this problem by incorporating Dubins costs or curvature-constrained motion primitives into target selection, or by decomposing the environment into subareas and optimizing their traversal order with curvature-constrained transition costs.

Shen et al.~\cite{shen2019online} extended the $\varepsilon^*$ algorithm~\cite{song2018} to autonomous underwater vehicles for back-and-forth coverage pattern while selecting target cells based on the Dubins travel cost. The method discretizes the orientations in neighboring cells, and evaluates the Dubins paths from the current vehicle state to these candidate poses to find the path segment. Kan et al.~\cite{kan2020online} partitioned the environment into hexagonal cells whose size guarantees that the Dubins path covering each cell remains inside it. The robot then selects the neighboring uncovered cell with the largest number of covered or obstacle cells around it and follows a Dubins path to that cell. Maini et al.~\cite{maini2022online} developed two autonomous mowing planners, JUMP and SNAKE. JUMP follows a back-and-forth pattern and uses Dubins paths to visit nearby weeds before returning to the current lap. In contrast, SNAKE does not require such returns; it connects Dubins subpaths to weeds along the current motion direction, forming a more flexible serpentine trajectory.
 
Lewis et al.~\cite{lewis2017semi} partitioned the environment into rectangular subareas, each of which can be covered by a single straight line. Each subarea is represented by two directed vertices corresponding to opposite coverage directions, and the edge weights are the Dubins path lengths between subarea exit and entry poses. The coverage order and direction are then determined by solving a generalized TSP. Karapetyan et al.~\cite{karapetyan2018multi} extended this framework to multi-robot systems by computing a single-robot Dubins coverage route and splitting it into balanced subtours through min-max optimization. Li et al.~\cite{li2023collision} further extended~\cite{lewis2017semi} to variable-speed curvature-constrained robots by sampling speed pairs, constructing variable-speed Dubins paths, pruning risky paths, and selecting minimum-time feasible transitions for TSP-based subarea sequencing.

\textit{Strengths and Limitations:} Curvature-constrained CPP produces executable paths for nonholonomic robots, but often increases travel distance and complicates target selection, subarea sequencing, and boundary coverage. Its performance depends on the motion model and orientation discretization.

\subsection{Tethered CPP}
\label{sec:tetheredrobot}

A tethered robot is physically connected to a fixed base station through a cable, which can extend operation time, improve communication reliability, and enhance safety~\cite{cao2023neptune}. In comparison to classical CPP, tethered CPP must consider cable-length limits, cable entanglement, and obstacle-induced topological constraints. Shnaps and Rimon~\cite{shnaps2014online} first established an offline tethered CPP method for known environments, which later served as the basis for their online algorithm. The method assigns each free cell a potential value defined by its shortest-path distance from the base station, which represents the minimum cable length required to reach that cell. Thus, only cells whose potential values do not exceed the cable length are considered reachable. Starting from the base station, the robot follows increasing equipotential contours to cover the environment outward. When obstacles split the coverage front into different corridors, the corresponding split cells are stored in a stack, allowing the robot to cover one corridor first and later retreat along the cable path to cover the remaining corridors. The online version follows the same principle while updating potential values, split cells, and corridor partitions as new obstacles are discovered. 

Sharma et al.~\cite{sharma2019} incrementally constructed a tree map rooted at the base station, where each node corresponds to a free cell and its depth encodes contour distance from the base station. A modified depth-first search is then performed over the tree, alternating between forward traversal toward deeper contours and backtracking to the nearest node with unvisited reachable neighbors, while respecting the cable-length constraint. Peng et al.~\cite{peng2025spanning} adapted the spanning tree covering algorithm~\cite{gabriely2001spanning} to tethered CPP. The method constructs a spanning tree rooted at the base station over the coarse cells by a Dijkstra-like procedure that uses taut cable-configuration length, rather than geometric path length, as the cost-to-come measure. The tree is circumnavigated to produce a coverage cycle with the cable fully retracted at the end.

\textit{Strengths and Limitations:} Tethered CPP extends operation time, improves communication reliability, and enhances safety, but is constrained by base location, entanglement risk, obstacle topology, and cable length. Its planning complexity increases when cable configurations must be updated in cluttered, unknown, or 3D environments.

\subsection{Reconfigurable CPP}
\label{sec:reconfgrobot}

Reconfigurable CPP uses morphology adaptation as an additional planning degree of freedom. By changing shape or size, robots can improve accessibility in narrow or cluttered regions and increase coverage efficiency in open areas. Samarakoon et al.~\cite{samarakoon2022online} used the BINN algorithm~\cite{luo2008bioinspired} to cover the main workspace with a fixed shape while reconfiguring the robot to cover cluttered regions around obstacles. Muthugala et al.~\cite{muthugala2023online} maintained occupancy grid maps for different robot sizes. Target cells are selected on the map associated with the smallest size using BINN~\cite{luo2008bioinspired}, while the robot moves toward the target using the largest locally collision-free size; after arrival, cells covered by the current footprint are marked as covered. This method was later improved by considering the cost and side effects of reconfiguration. Muthugala et al.~\cite{muthugala2025improving} introduced criteria to penalize unnecessary size changes and avoid moves that create isolated uncovered regions. Yi et al.~\cite{yi2026complete} developed a BINN-based method which jointly determines target-cell selection and robot reconfiguration.

\textit{Strengths and Limitations:} Reconfigurable CPP improves coverage efficiency through morphology adaptation. However, it requires modeling reconfiguration cost, transition feasibility, and footprint-dependent reachability, and complicates planning when morphology decisions are coupled with target selection.

\begin{table*}[t]{}
\footnotesize
\caption {Comparison of representative learning-based CPP methods.}\label{tab:learnbased_cpp_table}\vspace{-3pt}
\centering
\setlength\tabcolsep{5pt}
\begin{tabular}{l l l l l} 
 \toprule
\specialrule{0.1em}{1pt}{1pt} 
\tabincell{l}{\textbf{CPP Setting}}
&\multicolumn{1}{l}{\textbf{Method}}
&\tabincell{c}{\textbf{Year}}
&\tabincell{l}{\textbf{Learning-Enhanced Component}} 
&\tabincell{l}{{\textbf{Learning Technique}}} \\ 
\toprule

\specialrule{0em}{1pt}{0pt}
\multirow{4}{*}{\tabincell{l}{\textbf{{Single-Robot CPP}}}}
& \tabincell{l}{\textbf{AD Path}~\cite{chen2019adaptive}}     & 2019     &\tabincell{l}{{Subarea traversal order}} &\tabincell{l}{{Reinforcement learning}}\vspace{0.5em}\\

& \tabincell{l}{\textbf{CPPNet}~\cite{shen2021cppnet}}     & 2021     &\tabincell{l}{{Near-optimal coverage path generation}}   &\tabincell{l}{{GNN-based supervised learning}}\vspace{0.5em}\\

& \tabincell{l}{\textbf{PaintNet}~\cite{tiboni2023paintnet}}     & 2023     &\tabincell{l}{{6D Local painting path segment generation}}    &\tabincell{l}{{Supervised learning}}\\

\specialrule{0.05em}{2pt}{2pt}
\multirow{6}{*}{\tabincell{l}{\textbf{{Multi-Robot CPP}}}}
& \tabincell{l}{\textbf{Tolstaya et al.}~\cite{tolstaya2021multi}}     & 2021     &\tabincell{l}{{Online waypoint selection}} &\tabincell{l}{{Behavior cloning and GNN}}\vspace{0.5em}\\

& \tabincell{l}{\textbf{MDOC}~\cite{hu2023multi}}     & 2023     &\tabincell{l}{{Distributed online action coordination}}    &\tabincell{l}{{Distributed cooperative deep Q-learning}}\vspace{0.5em}\\

& \tabincell{l}{\textbf{Zhao et al.}~\cite{zhao2023energy}}     & 2023     &\tabincell{l}{{Energy-aware subarea allocation}}    &\tabincell{l}{{Multi-agent reinforcement learning}}\vspace{0.5em}\\

& \tabincell{l}{\textbf{Huang et al.}~\cite{huang2025hierarchical}}     & 2025     &\tabincell{l}{{Subarea allocation with minimized travel distance}}    &\tabincell{l}{{Multi-agent reinforcement learning}}\\

\specialrule{0.05em}{2pt}{2pt}
\multirow{5}{*}{\tabincell{l}{\textbf{Constrained CPP}}}
& \tabincell{l}{\textbf{Theile et al.}~\cite{theile2020uav}}     & 2020     &\tabincell{l}{{UAV action selection under energy constraints}} &\tabincell{l}{{Double deep Q-network}}\vspace{0.5em}\\

& \tabincell{l}{\textbf{Tang et al.}~\cite{tang2022learning}}     & 2022     &\tabincell{l}{{Coordination between worker coverage and}\\{mobile-station replenishment}}    &\tabincell{l}{{Multi-agent reinforcement learning}}\vspace{0.5em}\\

& \tabincell{l}{\textbf{Pey et al.}~\cite{pey2024decentralized}}     & 2024     &\tabincell{l}{{Coverage with morphology adaptation}}    &\tabincell{l}{{Multi-agent reinforcement learning}}\\

\bottomrule
\specialrule{0.1em}{1pt}{1pt}
\end{tabular}
\vspace{-1.0em}
\end{table*}

\section{Learning-Based CPP}
\label{sec:learningcpp}
Recently, Learning-based CPP has emerged as a promising methodological direction. Existing methods integrate different learning techniques (e.g., supervised learning, graph neural networks, Q-learning, and reinforcement learning) into classical CPP frameworks to generate coverage paths, determine subarea traversal orders, allocate tasks among multiple robots, and coordinate coverage under robot-specific constraints. This section reviews representative learning-based CPP methods according to the planning components enhanced by learning. Table~\ref{tab:learnbased_cpp_table} summarizes their key features in terms of CPP setting, learning-enhanced component, and learning technique.

One line of work uses learning techniques in single-robot CPP to generate subarea traversal policies or coverage paths from known environment representations. Chen et al.~\cite{chen2019adaptive} proposed the adaptive deep path (AD Path) algorithm, which uses reinforcement learning for subarea-level sequencing. It first applies Boustrophedon decomposition~\cite{choset2000coverage} to partition the environment into subareas, evaluates intra-subarea coverage and inter-subarea transition costs based on the entry and exit points of each subarea, and then learns the subarea traversal order. Shen et al.~\cite{shen2021cppnet} proposed CPPNet, which employs supervised learning with a graph neural network (GNN) for near-optimal coverage path generation. For training phase, random environments are converted into adjacency-graph representations, and TSP tours are used as reference solutions to supervise edge-probability prediction. CPPNet predicts an edge-probability heat graph from an occupancy-grid representation, where each edge value indicates its likelihood of belonging to the optimal TSP tour. A greedy search is then performed on this graph to generate the coverage path. Beyond 2D environments, Tiboni et al.~\cite{tiboni2023paintnet} proposed PaintNet for spray painting on 3D objects. Given an object point cloud, the supervised learning generates 6D local path segments, where each point contains the spray gun position and orientation. These segments are concatenated into full painting paths. 

Another line of work uses learning techniques for multi-robot coordination and task allocation. Tolstaya et al.~\cite{tolstaya2021multi} represented the environment as a spatial graph, with waypoints and robots as nodes and feasible movements as edges. A GNN is trained by behavior cloning to imitate an optimization-based vehicle-routing expert and select each robot's next waypoint online. Hu et al.~\cite{hu2023multi} proposed a multi-UAV distributed online cooperation (MDOC) method for coverage in unknown environments. Robots share coverage records and detected obstacles via an environmental map fusion mechanism. A distributed cooperative deep Q-learning algorithm selects actions to reduce completion time while avoiding overlap, missed coverage, and collisions. In energy-constrained problems, Zhao et al.~\cite{zhao2023energy} used multi-agent reinforcement learning (MARL) to dynamically allocate target subareas while accounting for energy consumption. Huang et al.~\cite{huang2025hierarchical} formulated multi-robot CPP as a two-level TSP problem. A MARL framework determines high-level subarea allocation and visitation order, while preset TSP-based paths are used for local coverage.

Learning techniques have also been used for constrained CPP. Theile et al.~\cite{theile2020uav} studied UAV CPP under energy constraints, random start positions, multiple landing positions, and no-fly zones. They formulated the problem as a Markov decision process and trained a double deep Q-network with map-like input channels to learn a control policy that balances target coverage, limited movement budget, and safe landing. Tang et al.~\cite{tang2022learning} considered a worker-station multi-robot coverage system, where energy-constrained worker robots coordinate with a mobile station for replenishment. This problem is formulated as a MARL problem, where robots decide whether to continue coverage or coordinate with the station based on local observations. Pey et al.~\cite{pey2024decentralized} formulated reconfigurable multi-robot CPP as a decentralized partially observable Markov decision process, where each robot selects motion and reconfiguration actions based on local observations and its current state. MARL is then used to coordinate coverage, obstacle avoidance, and morphology adaptation.

\textit{Strengths and Limitations:} Learning-based CPP is a relatively new direction. It shows potential in improving coverage efficiency by learning heuristics, traversal policies, or coordination strategies from data or interactions. However, these methods still provide weaker guarantees on completeness, safety, generalization, and robustness than classical CPP methods. This motivates hybrid frameworks that combine learned decision-making with formal planning mechanisms.

\begin{figure*}[t]
        \centering        
        \includegraphics[width=0.95\textwidth]{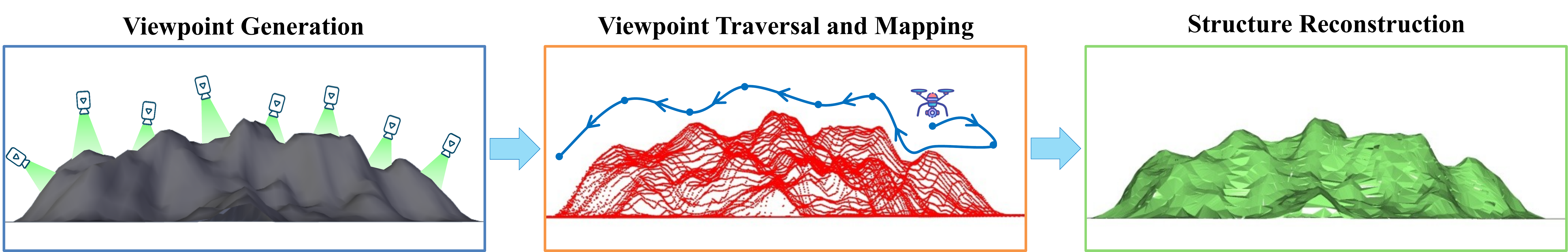}
    \caption{Illustration of a typical visual CPP pipeline. Candidate viewpoints are generated around the target surface, connected into an inspection path, and used to collect observations for reconstruction or inspection.}\label{fig:visualcppexample} 
    \vspace{-1em}
 \end{figure*}
 
\section{Visual CPP}
\label{sec:visualcpp}

Visual CPP is closely related to classical CPP, but differs in how coverage is achieved. Classical CPP requires the robot to physically traverse or sweep the target region, while visual CPP aims to inspect an object, structure, or scene by planning viewpoints that sufficiently observe its surface or geometry, as shown in Fig.~\ref{fig:visualcppexample}. In literature, visual CPP is often discussed under related terms such as inspection path planning or view path planning~\cite{song2020online}. Since the primary focus of this survey is classical CPP, this section briefly summarizes representative Visual CPP paradigms that are most relevant from the coverage-planning perspective.

\subsection{Classical Methods}

Given a target object, the classical Visual CPP methods first generate viewpoints to observe the surface, then select viewpoints based on coverage completeness, sensing quality, and travel cost, and finally compute a tour over the selected viewpoints using TSP formulations. Early methods solve a set-covering problem to obtain a compact viewpoint subset that satisfies complete coverage, while later extensions improve viewpoint quality by refining placement, optimizing viewpoint connections, or incorporating measurement uncertainty.

Englot and Hover~\cite{englot2013three} sampled random viewpoints around the target object and built a roadmap where each surface primitive is observable from multiple viewpoints. Then, a compact subset is selected via set covering, followed by TSP-based tour generation for collision-free inspection. Jing et al.~\cite{jing2016sampling} generated random viewpoints, determined viewing directions from nearby surface patches, and selected a compact subset through set covering. This method was later extended to manipulator inspection~\cite{jing2017sampling}, local path primitives instead of discrete viewpoints~\cite{jing2019coverage}, and multi-robot visual CPP~\cite{jing2020multi}.

Later methods used geometric structure to generate more informative viewpoint candidates. Zheng et al.~\cite{zheng2024new} clustered triangular meshes using spatial proximity and normal-vector similarity, generated viewpoints from the clusters, and refined viewpoint positions using a local potential field. Zhang et al.~\cite{zhang2026novel} further improved this pipeline by clustering mesh facets according to normal directions and spatial density, projecting each cluster onto a 2D plane, generating viewpoints through projection-based rectangular coverage, and jointly optimizing viewpoint selection and sequencing with a multi-objective evolutionary framework~\cite{gupta2015multifactorial}. Bircher et al.~\cite{bircher2016three} initialized one feasible viewpoint for each mesh triangle and alternated between viewpoint resampling and tour optimization to preserve visibility while reducing travel cost. Almadhoun et al.~\cite{almadhoun2018coverage} used adaptive viewpoint sampling to add viewpoints in regions with insufficient coverage or low reconstruction accuracy, and selected the next waypoint by balancing travel distance and expected information gain.

Beyond geometric visibility, some methods incorporate reconstruction or measurement uncertainty into Visual CPP. Lindner et al.~\cite{lindner2019optimization} addressed multi-view coverage for structure-from-motion reconstruction by estimating an upper bound on the expected reconstruction error. The method models image-plane projection errors, propagates the corresponding uncertainty into a 3D covariance estimate for each surface point, and updates this covariance as new views are added. The next viewpoint is selected to maximize reconstruction-uncertainty reduction under geometric and dynamic constraints. Liu et al.~\cite{liu2022coverage} considered measurement uncertainty in robotic quality inspection by deriving admissible uncertainty ranges from tolerance specifications, generating feasible viewpoints, and selecting viewpoints that balance the number of views and average measurement uncertainty.

\textit{Strengths and Limitations:} Classical methods typically optimize viewpoints over the entire space and sequence them using TSP or related tour-optimization formulations. They are effective for simple or moderate-scale environments, but become computationally expensive in large or complex scenes due to dense viewpoints, visibility and collision checks, occlusion reasoning, and large-scale tour optimization. This motivates hierarchical methods that break the space into smaller planning units and coordinate local plans via a higher-level structure.

\subsection{Hierarchical planning Methods}

Hierarchical methods decompose the environment into subareas, compute a global visiting order among them, and solve smaller local coverage problems within each subarea. This reduces planning complexity, improves scalability, and allows complex regions to be planned at finer resolution.

Cao et al.~\cite{cao2020hierarchical} proposed a multi-resolution Octree-based hierarchy for complex 3D environments. Feasible viewpoints are sampled and stored in an Octree, where denser or more complex regions are subdivided into finer subareas. A global TSP determines the visiting order of subareas, while local set-covering and TSP problems are solved in each leaf subspace to generate local coverage paths. Song et al.~\cite{song2020online} addressed online 3D modeling in unknown environments by combining global coverage planning with local inspection. The environment is incrementally decomposed into sectors to form a topological map, where a global coverage path guides the robot toward unexplored sectors, and local inspection paths scan frontier regions around the current path. Morilla et al.~\cite{morilla2022sweep} focused on large-scale outdoor visual coverage with multiple UAVs. A high-altitude reconnaissance flight first builds a coarse prior map, after which poorly observed or unobserved regions are detected, clustered, covered by sweep trajectories, and assigned to multiple drones through a vehicle-routing-based planner. Feng et al.~\cite{feng2024fc} proposed a skeleton-guided hierarchy for aerial coverage of complex 3D scenes. A scene skeleton is extracted from the point cloud and decomposed into branches, each defining a subarea. The skeleton guides free-space identification and viewpoint sampling, while a hierarchical planner determines the global visiting order of subspaces and optimizes local coverage paths.

%% file: sections/conclusions.tex
\section{Conclusion and Future Directions} \label{sec:conclusions}

\textit{Conclusion:} This survey presents a comprehensive and structured review of 125 representative works, primarily published between 2015 and 2026, while linking recent advances to classical CPP foundations. These reviewed methods are organized into six categories: single-robot CPP, multi-robot CPP, 3D CPP, constrained CPP, learning-based CPP, and visual CPP. For each category, we summarize the main problem formulations, representative algorithms, key ideas, strengths, and limitations. Several methodological trends are observed. 

\begin{itemize}
\item Subarea decomposition remains a central principle in CPP to reduce the complexity of offline coverage path generation in single-robot, multi-robot, and 3D CPP.  It also provides higher-level decision structures for non-local waypoint selection in single-robot CPP, and reduces inter-robot conflicts by assigning robots to distinct subareas in multi-robot CPP.

\item Online single-robot CPP has long relied on local methods for computational efficiency and ease of implementation. Recently, non-local methods have attracted increasing attention by incorporating higher-level decision structures to enable non-myopic waypoint selection and reduce dead ends and fragmentation-induced redundant travel.

\item Compared with single-robot CPP, multi-robot CPP can accelerate large-scale coverage, improve resilience to robot failures, and support parallel execution. Recent methods have moved from grid-cell-level coordination toward subarea-level coordination to reduce allocation complexity, improve workload balancing, lower coordination overhead, and mitigate inter-robot conflicts.

\item Learning-based CPP has emerged as a
novel methodology that improves coverage efficiency, adaptability, and scalability by using data-driven models to guide path generation, subarea sequencing, task allocation, and coordination, often in combination with classical CPP.

\item Constrained and visual CPP extend the application scope of CPP. Constrained CPP adapts coverage planning to robot-specific feasibility requirements, such as tether length, energy budget, curvature limits, and morphology constraints, while visual CPP extends coverage from physical area traversal to sensing-driven tasks such as inspection, reconstruction, and monitoring by explicitly considering visibility, viewpoints, and sensing quality.

\end{itemize}

\textit{Future Works:} Several research directions deserve further attention in future:
\begin{itemize}
\item  Recently, non-local methods in online single-robot CPP have shown the value of higher-level decision structures. Extending such methods to multi-robot, 3D, and visual CPP remains a promising but underexplored direction.

\item Multi-robot CPP is important for large-scale coverage, but many methods still assume homogeneous robots and reliable communication. In real applications, robots may differ in coverage footprint, speed, energy capacity, and motion constraints, while communication can be intermittent or locally restricted. Future work should develop multi-robot CPP methods for heterogeneous teams, limited communication, and robust task allocation.

\item Existing 3D and visual CPP methods often rely on accurate terrain models, surface meshes, visibility models, or prior geometric information. However, in many real-world tasks, geometry must be reconstructed online from noisy and incomplete sensor data. Future research should further integrate mapping, surface representation, uncertainty estimation, viewpoint selection, and path planning, especially for online 3D coverage and inspection tasks.

\item Learning-based methods have shown potential for improving coverage efficiency. However, these methods still lack guarantees on coverage completeness, safety, generalization, and robustness. Future research should explore hybrid frameworks that combine the adaptability of learning with the interpretability and guarantees of classical CPP methods. Learning may be used to guide sampling, predict difficult regions, estimate coverage cost, or accelerate high-level decisions, while classical planners provide feasibility and completeness mechanisms.

\item Embodied AI may extend CPP from geometry-driven coverage to task- and semantics-aware coverage by integrating semantic mapping, vision-language models, active perception, and embodied decision-making. This could help robots adapt coverage strategies to semantic regions, human instructions, and changing mission objectives.

\item Recent developments on motion planning in dynamic environments have highlighted the importance of real-time replanning, uncertainty-aware prediction, and risk-aware decision-making~\cite{shen2023smart, shen2026motion}. These techniques may provide useful foundations for extending online CPP from static environments to highly dynamic environments with moving obstacles, humans, or other robots.

\end{itemize}

%% file: reference.bib
@IEEEtranBSTCTL{IEEEexample:BSTcontrol,
  CTLdash_repeated_names = "no"
}

@article{acar2002sensor,
  title={Sensor-based coverage of unknown environments: Incremental construction of morse decompositions},
  author={Acar, Ercan U and Choset, Howie},
  journal={International Journal of Robotics Research},
  volume={21},
  number={4},
  pages={345--366},
  year={2002},
  publisher={SAGE Publications Sage UK: London, England}
}

@article{wilson2025generalized,
  title={Generalized multi-speed dubins motion model},
  author={Wilson, James P and Gupta, Shalabh and Wettergren, Thomas A},
  journal={IEEE Transactions on Robotics},
    year={2025},
  volume={41},
  number={},
  pages={2861-2878},
  publisher={IEEE}
}

@article{shen2023smart,
  title={\uppercase{SMART}: Self-Morphing Adaptive Replanning Tree},
  author={Shen, Zongyuan and Wilson, James P and Gupta, Shalabh and Harvey, Ryan},
  journal={IEEE Robotics and Automation Letters},
  year={2023},
  volume={8},
  number={11},
  pages={7312-7319}
}

@article{shen2026motion,
  title={Motion Planning in Dynamic Environments: A Survey from Classical to Modern Methods},
  author={Shen, Zongyuan and Ou, Yaming and Gupta, Shalabh and Zhao, Shancheng and Zhou, Dehua and Wang, Gao and Ren, Zhongqiang and Fan, Junfeng and Cheng, Long},
  journal={arXiv preprint arXiv:2606.02677},
  year={2026}
}

@article{lloyd1982least,
  title={Least squares quantization in PCM},
  author={Lloyd, Stuart},
  journal={IEEE Transactions on Information Theory},
  volume={28},
  number={2},
  pages={129--137},
  year={1982},
  publisher={IEEE}
}

@article{gabriely2001spanning,
  title={Spanning-tree based coverage of continuous areas by a mobile robot},
  author={Gabriely, Yoav and Rimon, Elon},
  journal={Annals of Mathematics and Artificial Intelligence},
  volume={31},
  number={1-4},
  pages={77--98},
  year={2001},
  publisher={Springer}
}

@article{gabriely2003competitive,
  title={Competitive on-line coverage of grid environments by a mobile robot},
  author={Gabriely, Yoav and Rimon, Elon},
  journal={Computational Geometry},
  volume={24},
  number={3},
  pages={197--224},
  year={2003},
  publisher={Elsevier}
}

@inproceedings{gonzalez2005bsa,
  title={\uppercase{BSA}: a complete coverage algorithm},
  author={Gonzalez, Enrique and Alvarez, Oscar and Diaz, Yul and Parra, Carlos and Bustacara, Cesar},
    booktitle={IEEE International Conference on Robotics and Automation},
  pages={2040--2044},
  year={2005},    
}

@inproceedings{ferranti2007brick,
  title={Brick\&{M}ortar: {A}n on-line multi-agent exploration algorithm},
  author={Ferranti, Ettore and Trigoni, Niki and Levene, Mark},
booktitle={IEEE International Conference on Robotics and Automation},
  pages={761--767},
  year={2007},  
}

@article{luo2008bioinspired,
  title={A bioinspired neural network for real-time concurrent map building and complete coverage robot navigation in unknown environments},
  author={Luo, Chaomin and Yang, Simon X},
  journal={IEEE Transactions on Neural Networks and Learning Systems},
  volume={19},
  number={7},
  pages={1279--1298},
  year={2008},
  publisher={IEEE}
}

@article{viet2013ba,
  title={\uppercase{BA}*: an online complete coverage algorithm for cleaning robots},
  author={Viet, Hoang Huu and Dang, Viet-Hung and Laskar, Md Nasir Uddin and Chung, TaeChoong},
  journal={Applied Intelligence},
  volume={39},
  number={2},
  pages={217--235},
  year={2013},
  publisher={Springer}
}

@article{song2018,
 title={$\epsilon^\star$: An Online Coverage Path Planning Algorithm},
 author={J. Song and S. Gupta},
 journal={IEEE Transactions on Robotics},
  volume={34},
  number={2},
  pages={526--533},
  year={2018},
  publisher={IEEE}
}

@article{hassan2019ppcpp,
  title={\uppercase{PPCPP}: A Predator--Prey-Based Approach to Adaptive Coverage Path Planning},
  author={Hassan, Mahdi and Liu, Dikai},
  journal={IEEE Transactions on Robotics},
  volume={36},
  number={1},
  pages={284--301},
  year={2019},
  publisher={IEEE}
}

@article{jin2011coverage,
  title={Coverage path planning on three-dimensional terrain for arable farming},
  author={Jin, Jian and Tang, Lie},
  journal={Journal of Field Robotics},
  volume={28},
  number={3},
  pages={424--440},
  year={2011},
  publisher={Wiley Online Library}
}

@INPROCEEDINGS{Song_sose2015,
  author={Song, Junnan and Gupta, Shalabh},
  booktitle={2015 10th System of Systems Engineering Conference}, 
  title={\uppercase{SLAM} based shape adaptive coverage control using autonomous vehicles}, 
  year={2015},
  volume={},
  number={},
  pages={268-273},
  doi={10.1109/SYSOSE.2015.7151959}}

@article{cai2023,
  author={Cai, Wenyu and Zhang, Shuai and Zhang, Meiyan and Wang, Chengcai},
  journal={IEEE Internet of Things Journal}, 
  title={Improved \uppercase{BINN}-Based Underwater Topography Scanning Coverage Path Planning for AUV in Internet of Underwater Things}, 
  year={2023},
  volume={10},
  number={20},
  pages={18375-18386}}

@ARTICLE{han2023,
  author={Han, Linhui and Tan, Xiangquan and Wu, Qingwen and Deng, Xu},
  journal={IEEE Transactions on Cognitive and Developmental Systems}, 
  title={An Improved Algorithm for Complete Coverage Path Planning Based on Biologically Inspired Neural Network}, 
  year={2023},
  volume={15},
  number={3},
  pages={1605-1617}}

@article{li2023sp2e,
  title={\uppercase{SP2E}: Online spiral coverage with proactive prevention extremum for unknown environments},
  author={Li, Lin and Shi, Dianxi and Jin, Songchang and Yang, Shaowu and Lian, Yaoning and Liu, Hengzhu},
  journal={Journal of Intelligent \& Robotic Systems},
  volume={108},
  number={2},
  pages={30},
  year={2023},
  publisher={Springer}
}

@ARTICLE{ramesh2024,
  author={Ramesh, Megnath and Imeson, Frank and Fidan, Baris and Smith, Stephen L.},
  journal={IEEE Transactions on Robotics}, 
  title={Anytime Replanning of Robot Coverage Paths for Partially Unknown Environments}, 
  year={2024},
  volume={40},
  number={},
  pages={4190-4206}}

@ARTICLE{huo2025,
  author={Huo, Lintao and Liu, Ying and Chen, Zengtao and Yang, Yutu and Yan, Xiaoan and Xia, Haifei and Sun, Qi},
  journal={IEEE Robotics and Automation Letters}, 
  title={Complete Coverage Path Planning Algorithm Based on Improved Biologically Inspired Neural Networks in Spray Painting}, 
  year={2025},
  volume={10},
  number={6},
  pages={5697-5704}}

@ARTICLE{li2025,
  author={Li, Lin and Shi, Dianxi and Jin, Songchang and Zhou, Xing and Li, Yahui and Bai, Bin},
  journal={IEEE Robotics and Automation Letters}, 
  title={Hierarchy Coverage Path Planning With Proactive Extremum Prevention in Unknown Environments}, 
  year={2025},
  volume={10},
  number={4},
  pages={3358-3365}}

@INPROCEEDINGS{shen2025_cap,
  author={Shen, Zongyuan and Shirose, Burhanuddin and Sriganesh, Prasanna and Travers, Matthew},
  booktitle={IEEE/RSJ International Conference on Intelligent Robots and Systems}, 
  title={\uppercase{CAP}: A Connectivity-Aware Hierarchical Coverage Path Planning Algorithm for Unknown Environments using Coverage Guidance Graph}, 
  year={2025},
  volume={},
  number={},
  pages={13244-13249}}

@article{palomeras2018autonomous,
  title={Autonomous mapping of underwater 3-\uppercase{D} structures: From view planning to execution},
  author={Palomeras, Narc{\'\i}s and Hurt{\'o}s, Natalia and Carreras, Marc and Ridao, Pere},
  journal={IEEE Robotics and Automation Letters},
  volume={3},
  number={3},
  pages={1965--1971},
  year={2018},
  publisher={IEEE}
}

@article{vidal2017,
  title={Online View Planning for Inspecting Unexplored Underwater Structures.},
  author={Vidal, Eduard and Hern{\'a}ndez, Juan David and Istenic, Klemen and Carreras, Marc},
  journal={IEEE Robotics and Automation Letters},
  volume={2},
  number={3},
  pages={1436--1443},
  year={2017}
}

@ARTICLE{shen2026,
  author={Shen, Zongyuan and Wilson, James P. and Gupta, Shalabh},
  journal={IEEE Transactions on Robotics}, 
  title={C$^{*}$: A Coverage Path Planning Algorithm for Unknown Environments Using Rapidly Covering Graphs}, 
  year={2026},
  volume={42},
  number={},
  pages={1233-1253}}

@article{hart1968formal,
  title={A formal basis for the heuristic determination of minimum cost paths},
  author={Hart, Peter E and Nilsson, Nils J and Raphael, Bertram},
  journal={IEEE Transactions on Systems, Man, and Cybernetics},
  volume={4},
  number={2},
  pages={100--107},
  year={1968},
  publisher={IEEE}
}

@inproceedings{bochkarev2016minimizing,
  title={On minimizing turns in robot coverage path planning},
  author={Bochkarev, Stanislav and Smith, Stephen L},
  booktitle={IEEE International Conference on Automation Science and Engineering},
  pages={1237--1242},
  year={2016}
}

@inproceedings{brown2016constriction,
  title={The constriction decomposition method for coverage path planning},
  author={Brown, Stanley and Waslander, Steven L},
  booktitle={IEEE/RSJ International Conference on Intelligent Robots and Systems},
  pages={3233--3238},
  year={2016}
}

@article{xu2014efficient,
  title={Efficient complete coverage of a known arbitrary environment with applications to aerial operations},
  author={Xu, Anqi and Viriyasuthee, Chatavut and Rekleitis, Ioannis},
  journal={Autonomous Robots},
  volume={36},
  number={4},
  pages={365--381},
  year={2014},
  publisher={Springer}
}

@article{ramesh2022optimal,
  title={Optimal partitioning of non-convex environments for minimum turn coverage planning},
  author={Ramesh, Megnath and Imeson, Frank and Fidan, Baris and Smith, Stephen L},
  journal={IEEE Robotics and Automation Letters},
  volume={7},
  number={4},
  pages={9731--9738},
  year={2022},
  publisher={IEEE}
}

@ARTICLE{ramesh2025,
  author={Ramesh, Megnath and Imeson, Frank and Fidan, Baris and Smith, Stephen L.},
  journal={IEEE Robotics and Automation Letters}, 
  title={Minimum-Length Coverage Path Planning for Grid Environments With Approximation Guarantees}, 
  year={2025},
  volume={10},
  number={10},
  pages={10674-10681}}

@article{choset2000coverage,
  title={Coverage of known spaces: The boustrophedon cellular decomposition},
  author={Choset, Howie},
  journal={Autonomous Robots},
  volume={9},
  number={3},
  pages={247--253},
  year={2000},
  publisher={Springer}
}

@article{acar2002morse,
  title={Morse decompositions for coverage tasks},
  author={Acar, Ercan U and Choset, Howie and Rizzi, Alfred A and Atkar, Prasad N and Hull, Douglas},
  journal={International Journal of Robotics Research},
  volume={21},
  number={4},
  pages={331--344},
  year={2002},
  publisher={SAGE Publications Sage UK: London, England}
}

@incollection{latombe1991exact,
  title={Exact cell decomposition},
  author={Latombe, Jean-Claude},
  booktitle={Robot motion planning},
  pages={200--247},
  year={1991},
  publisher={Springer}
}

@book{milnor1963morse,
  title={Morse theory},
  author={Milnor, John Willard},
  number={51},
  year={1963},
  publisher={Princeton university press}
}

@article{shnaps2014online,
  title={Online coverage by a tethered autonomous mobile robot in planar unknown environments},
  author={Shnaps, Iddo and Rimon, Elon},
  journal={IEEE Transactions on Robotics},
  volume={30},
  number={4},
  pages={966--974},
  year={2014},
  publisher={IEEE}
}

@inproceedings{sharma2019,
  title={A 2-Approximation Algorithm for the Online Tethered Coverage Problem.},
  author={Sharma, Gokarna and Poudel, Pavan and Dutta, Ayan and Zeinali, Vala and Khoei, Tala Talaei and Kim, Jong-Hoon},
  booktitle={Robotics: Science and systems},
  year={2019}
}

@article{peng2025spanning,
  title={Spanning-tree based coverage for a tethered robot},
  author={Peng, Xiao and Schwarzentruber, Fran{\c{c}}ois and Simonin, Olivier and Solnon, Christine},
  journal={IEEE Robotics and Automation Letters},
  volume={10},
  number={2},
  pages={1888--1895},
  year={2025},
  publisher={IEEE}
}

@article{cao2023neptune,
  title={\uppercase{Neptune}: Nonentangling trajectory planning for multiple tethered unmanned vehicles},
  author={Cao, Muqing and Cao, Kun and Yuan, Shenghai and Nguyen, Thien-Minh and Xie, Lihua},
  journal={IEEE Transactions on Robotics},
  volume={39},
  number={4},
  pages={2786--2804},
  year={2023},
  publisher={IEEE}
}

@article{shnaps2016online,
  title={Online coverage of planar environments by a battery powered autonomous mobile robot},
  author={Shnaps, Iddo and Rimon, Elon},
  journal={IEEE Transactions on Automation Science and Engineering},
  volume={13},
  number={2},
  pages={425--436},
  year={2016},
  publisher={IEEE}
}

@inproceedings{wei2018coverage,
  title={Coverage path planning under the energy constraint},
  author={Wei, Minghan and Isler, Volkan},
  booktitle={IEEE International Conference on Robotics and Automation},
  pages={368--373},
  year={2018}
}

@INPROCEEDINGS{shen2020,
  author={Z. {Shen} and J. P. {Wilson} and S. {Gupta}},
  booktitle={Global Oceans 2020: Singapore – U.S. Gulf Coast}, 
  title={$\epsilon^{\star}+$: An Online Coverage Path Planning Algorithm for Energy-constrained Autonomous Vehicles}, 
  year={2020},
  volume={},
  number={},
  pages={1-6}}

@article{dogru2022eco,
  title={\uppercase{ECO-CPP}: Energy constrained online coverage path planning},
  author={Dogru, Sedat and Marques, Lino},
  journal={Robotics and Autonomous Systems},
  volume={157},
  pages={104242},
  year={2022},
  publisher={Elsevier}
}

@article{kan2020online,
  title={Online exploration and coverage planning in unknown obstacle-cluttered environments},
  author={Kan, Xinyue and Teng, Hanzhe and Karydis, Konstantinos},
  journal={IEEE Robotics and Automation Letters},
  volume={5},
  number={4},
  pages={5969--5976},
  year={2020},
  publisher={IEEE}
}

@inproceedings{shen2019online,
  title={An online coverage path planning algorithm for curvature-constrained \uppercase{AUV}s},
  author={Shen, Zongyuan and Wilson, James P and Gupta, Shalabh},
  booktitle={OCEANS 2019 MTS/IEEE SEATTLE},
  pages={1--5},
  year={2019},
  organization={IEEE}
}

@Inproceedings{SGH13,
  author = "J. Song and S. Gupta and J. Hare and S. Zhou",
  title = "Adaptive Cleaning of Oil Spills by Autonomous Vehicles under Partial Information",
  year = "2013",
  pages = "1--5",
  booktitle={OCEANS 2019 MTS/IEEE San Diego}
}

@article{mukherjee2011symbolic,
  title={Symbolic analysis of sonar data for underwater target detection},
  author={Mukherjee, Kushal and Gupta, Shalabh and Ray, Asok and Phoha, Shashi},
  journal={IEEE Journal of Oceanic Engineering},
  volume={36},
  number={2},
  pages={219--230},
  year={2011},
  publisher={IEEE}
}

@article{maini2022online,
  title={Online coverage planning for an autonomous weed mowing robot with curvature constraints},
  author={Maini, Parikshit and Gonultas, Burak M and Isler, Volkan},
  journal={IEEE Robotics and Automation Letters},
  volume={7},
  number={2},
  pages={5445--5452},
  year={2022},
  publisher={IEEE}
}

@inproceedings{li2023collision,
  title={Collision-free coverage path planning for the variable-speed curvature-constrained robot},
  author={Li, Lin and Shi, Dianxi and Jin, Songchang and Sun, Yixuan and Zhou, Xing and Yang, Shaowu and Liu, Hengzhu},
  booktitle={IEEE International Conference on Robotics and Automation},
  pages={3600--3606},
  year={2023}
}

@inproceedings{samarakoon2022online,
  title={Online complete coverage path planning of a reconfigurable robot using glasius bio-inspired neural network and genetic algorithm},
  author={Samarakoon, SM Bhagya P and Muthugala, MA Viraj J and Elara, Mohan Rajesh},
  booktitle={IEEE/RSJ International Conference on Intelligent Robots and Systems},
  pages={5744--5751},
  year={2022}
}

@inproceedings{theile2020uav,
  title={\uppercase{UAV} coverage path planning under varying power constraints using deep reinforcement learning},
  author={Theile, Mirco and Bayerlein, Harald and Nai, Richard and Gesbert, David and Caccamo, Marco},
  booktitle={IEEE/RSJ International Conference on Intelligent Robots and Systems},
  pages={1444--1449},
  year={2020}
}

@inproceedings{tolstaya2021multi,
  title={Multi-robot coverage and exploration using spatial graph neural networks},
  author={Tolstaya, Ekaterina and Paulos, James and Kumar, Vijay and Ribeiro, Alejandro},
  booktitle={IEEE/RSJ International Conference on Intelligent Robots and Systems},
  pages={8944--8950},
  year={2021}
}

@inproceedings{pey2024decentralized,
  title={A Decentralized Partially Observable Markov Decision Process for Dynamic Obstacle Avoidance and Complete Area Coverage using Multiple Reconfigurable Robots},
  author={Pey, Javier Jia Jie and Samarakoon, SM Bhagya P and Muthugala, MA Viraj J and Elara, Mohan Rajesh},
  booktitle={IEEE/RSJ International Conference on Intelligent Robots and Systems},
  pages={8023--8030},
  year={2024}
}

@inproceedings{muthugala2023online,
  title={Online coverage path planning scheme for a size-variable robot},
  author={Muthugala, MA Viraj J and Samarakoon, SM Bhagya P and Elara, Mohan Rajesh},
  booktitle={IEEE International Conference on Robotics and Automation},
  pages={5688--5694},
  year={2023}
}

@inproceedings{muthugala2025improving,
  title={Improving Coverage Performance of a Size-Reconfigurable Robot Based on Overlapping and Reconfiguration Reduction Criteria},
  author={Muthugala, MA Viraj J and Samarakoon, SM Bhagya P and Wijegunawardana, ID and Elara, Mohan Rajesh},
  booktitle={IEEE International Conference on Robotics and Automation},
  pages={7902--7908},
  year={2025}
}

@article{yi2026complete,
  title={Complete Coverage Path Planning for Omnidirectional Self-Reconfigurable Cleaning Robot Using $ a $ \uppercase{GBNN}},
  author={Yi, Lim and Hayat, Abdullah Aamir and Sang, Ash Wan Yaw and Le, Anh Vu and Qinrui, Tang and Elara, Mohan Rajesh},
  journal={IEEE Transactions on Automation Science and Engineering},
  volume={23},
  pages={2212--2230},
  year={2026},
  publisher={IEEE}
}

@article{hu2023multi,
  title={Multi-\uppercase{UAV} coverage path planning: A distributed online cooperation method},
  author={Hu, Wenjian and Yu, Yao and Liu, Shumei and She, Changyang and Guo, Lei and Vucetic, Branka and Li, Yonghui},
  journal={IEEE Transactions on Vehicular Technology},
  volume={72},
  number={9},
  pages={11727--11740},
  year={2023},
  publisher={IEEE}
}

@inproceedings{tiboni2023paintnet,
  title={Paint\uppercase{n}et: Unstructured multi-path learning from 3\uppercase{d} point clouds for robotic spray painting},
  author={Tiboni, Gabriele and Camoriano, Raffaello and Tommasi, Tatiana},
  booktitle={IEEE/RSJ International Conference on Intelligent Robots and Systems},
  pages={3857--3864},
  year={2023}
}

@article{dubins1957curves,
  title={On curves of minimal length with a constraint on average curvature, and with prescribed initial and terminal positions and tangents},
  author={Dubins, Lester E},
  journal={American Journal of Mathematics},
  volume={79},
  number={3},
  pages={497--516},
  year={1957},
  publisher={JSTOR}
}

@inproceedings{lewis2017semi,
  title={Semi-boustrophedon coverage with a dubins vehicle},
  author={Lewis, Jeremy S and Edwards, William and Benson, Kelly and Rekleitis, Ioannis and O'Kane, Jason M},
  booktitle={IEEE/RSJ International Conference on Intelligent Robots and Systems},
  pages={5630--5637},
  year={2017}
}

@inproceedings{karapetyan2018multi,
  title={Multi-robot dubins coverage with autonomous surface vehicles},
  author={Karapetyan, Nare and Moulton, Jason and Lewis, Jeremy S and Li, Alberto Quattrini and O'Kane, Jason M and Rekleitis, Ioannis},
  booktitle={IEEE International Conference on Robotics and Automation},
  pages={2373--2379},
  year={2018}
}

@inproceedings{shen2021cppnet,
  title={\uppercase{CPPN}et: A coverage path planning network},
  author={Shen, Zongyuan and Agrawal, Palash and Wilson, James P and Harvey, Ryan and Gupta, Shalabh},
  booktitle={OCEANS 2021: San Diego--Porto},
  pages={1--5},
  year={2021},
  organization={IEEE}
}

@article{huang2025hierarchical,
  title={A hierarchical multi robot coverage strategy for large maps with reinforcement learning and dense segmented siamese network},
  author={Huang, Yihang and Wang, Yanwei and Li, Zeyi and Zhang, Haitao and Zhang, Chong},
  journal={IEEE Robotics and Automation Letters},
  volume={10},
  number={1},
  pages={444--451},
  year={2025},
  publisher={IEEE}
}

@article{viet2015bob,
  title={Bo\uppercase{B}: an online coverage approach for multi-robot systems},
  author={Viet, Hoang Huu and Dang, Viet-Hung and Choi, SeungYoon and Chung, Tae Choong},
  journal={Applied Intelligence},
  volume={42},
  number={2},
  pages={157--173},
  year={2015},
  publisher={Springer}
}

@article{luo2017neural,
  title={Neural-dynamics-driven complete area coverage navigation through cooperation of multiple mobile robots},
  author={Luo, Chaomin and Yang, Simon X and Li, Xinde and Meng, Max Q-H},
  journal={IEEE Transactions on Industrial Electronics},
  volume={64},
  number={1},
  pages={750--760},
  year={2017},
  publisher={IEEE}
}

@article{sun2019complete,
  title={Complete coverage autonomous underwater vehicles path planning based on glasius bio-inspired neural network algorithm for discrete and centralized programming},
  author={Sun, Bing and Zhu, Daqi and Tian, Chen and Luo, Chaomin},
  journal={IEEE Transactions on Cognitive and Developmental Systems},
  volume={11},
  number={1},
  pages={73--84},
  year={2019},
  publisher={IEEE}
}

@inproceedings{hassan2020dec,
  title={Dec-\uppercase{PPCPP}: A decentralized predator--prey-based approach to adaptive coverage path planning amid moving obstacles},
  author={Hassan, Mahdi and Mustafic, Daut and Liu, Dikai},
  booktitle={IEEE/RSJ International Conference on Intelligent Robots and Systems},
  pages={11732--11739},
  year={2020}
}

@article{ma2022cciba,
  title={\uppercase{CCIBA}*: An improved \uppercase{BA}* based collaborative coverage path planning method for multiple unmanned surface mapping vehicles},
  author={Ma, Yong and Zhao, Yujiao and Li, Zhixiong and Bi, Huaxiong and Wang, Jing and Malekian, Reza and Sotelo, Miguel Angel},
  journal={IEEE Transactions on Intelligent Transportation Systems},
  volume={23},
  number={10},
  pages={19578--19588},
  year={2022},
  publisher={IEEE}
}

@article{zhang2024herd,
  title={A herd-foraging-based approach to adaptive coverage path planning in dual environments},
  author={Zhang, Junqi and Zu, Peng and Liu, Kun and Zhou, Mengchu},
  journal={IEEE Transactions on Cybernetics},
  volume={54},
  number={3},
  pages={1882--1893},
  year={2024},
  publisher={IEEE}
}

@article{wang2024apf,
  title={\uppercase{APF-CPP}: An artificial potential field based multi-robot online coverage path planning approach},
  author={Wang, Zikai and Zhao, Xiaoqi and Zhang, Jiekai and Yang, Nachuan and Wang, Pengyu and Tang, Jiawei and Zhang, Jiuzhou and Shi, Ling},
  journal={IEEE Robotics and Automation Letters},
  volume={9},
  number={11},
  pages={9199--9206},
  year={2024},
  publisher={IEEE}
}

@article{song2020care,
  title={\uppercase{Care}: Cooperative autonomy for resilience and efficiency of robot teams for complete coverage of unknown environments under robot failures},
  author={Song, Junnan and Gupta, Shalabh},
  journal={Autonomous Robots},
  volume={44},
  pages={647--671},
  year={2020},
  publisher={Springer}
}

@article{wang2025mac,
  title={\uppercase{MAC-P}lanner: A Novel Task Allocation and Path Planning Framework for Multi-Robot Online Coverage Processes},
  author={Wang, Zikai and Lyu, Xiaoxu and Zhang, Jiekai and Wang, Pengyu and Zhong, Yuxing and Shi, Ling},
  journal={IEEE Robotics and Automation Letters},
  volume={10},
  number={5},
  pages={4404-4411},
  year={2025},
  publisher={IEEE}
}

@inproceedings{tang2021mstc,
  title={\uppercase{MSTC}*: Multi-robot coverage path planning under physical constrain},
  author={Tang, Jingtao and Sun, Chun and Zhang, Xinyu},
  booktitle={IEEE International Conference on Robotics and Automation},
  pages={2518--2524},
  year={2021}
}

@article{lu2023tmstc,
  title={\uppercase{TMSTC}*: A path planning algorithm for minimizing turns in multi-robot coverage},
  author={Lu, Junjie and Zeng, Bi and Tang, Jingtao and Lam, Tin Lun and Wen, Junbin},
  journal={IEEE Robotics and Automation Letters},
  volume={8},
  number={8},
  pages={5275--5282},
  year={2023},
  publisher={IEEE}
}

@article{tang2023mixed,
  title={Mixed integer programming for time-optimal multi-robot coverage path planning with efficient heuristics},
  author={Tang, Jingtao and Ma, Hang},
  journal={IEEE Robotics and Automation Letters},
  volume={8},
  number={10},
  pages={6491--6498},
  year={2023},
  publisher={IEEE}
}

@ARTICLE{tang2025,
  author={Tang, Jingtao and Mao, Zining and Ma, Hang},
  journal={IEEE Transactions on Robotics}, 
  title={Large-Scale Multirobot Coverage Path Planning on Grids With Path Deconfliction}, 
  year={2025},
  volume={41},
  number={},
  pages={3348-3367}}

@ARTICLE{lee2026,
  author={Lee, Kanghoon and Kim, Hyeonjun and Li, Jiachen and Park, Jinkyoo},
  journal={IEEE Robotics and Automation Letters}, 
  title={Priority-Aware Multi-Robot Coverage Path Planning}, 
  year={2026},
  volume={11},
  number={3},
  pages={3534-3541}}

@article{zheng2010multirobot,
  title={Multirobot forest coverage for weighted and unweighted terrain},
  author={Zheng, Xiaoming and Koenig, Sven and Kempe, David and Jain, Sonal},
  journal={IEEE Transactions on Robotics},
  volume={26},
  number={6},
  pages={1018--1031},
  year={2010},
  publisher={IEEE}
}

@inproceedings{hazon2005redundancy,
  title={Redundancy, efficiency and robustness in multi-robot coverage},
  author={Hazon, Noam and Kaminka, Gal A},
  booktitle={IEEE International Conference on Robotics and Automation},
  pages={735--741},
  year={2005}
}

@article{dong2020artificially,
  title={An artificially weighted spanning tree coverage algorithm for decentralized flying robots},
  author={Dong, Wei and Liu, Sensen and Ding, Ye and Sheng, Xinjun and Zhu, Xiangyang},
  journal={IEEE Transactions on Automation Science and Engineering},
  volume={17},
  number={4},
  pages={1689--1698},
  year={2020},
  publisher={IEEE}
}

@ARTICLE{chen2026dynamic,
  author={Chen, Yanjie and Zeng, Zhaoguo and Lai, Zhennan and Jiang, Wensheng and Lu, Huimin and Zhang, Hui and Wang, Yaonan},
  journal={IEEE Transactions on Industrial Electronics}, 
  title={Dynamic Task-Priority Coverage Planning for Efficient Multi-Robot Collaboration}, 
  year={2026},
  volume={73},
  number={2},
  pages={2521-2532}}

@inproceedings{karapetyan2017efficient,
  title={Efficient multi-robot coverage of a known environment},
  author={Karapetyan, Nare and Benson, Kelly and McKinney, Chris and Taslakian, Perouz and Rekleitis, Ioannis},
  booktitle={IEEE/RSJ International Conference on Intelligent Robots and Systems},
  pages={1846--1852},
  year={2017}
}

@inproceedings{vandermeulen2019turn,
  title={Turn-minimizing multirobot coverage},
  author={Vandermeulen, Isaac and Gro{\ss}, Roderich and Kolling, Andreas},
  booktitle={IEEE International Conference on Robotics and Automation},
  pages={1014--1020},
  year={2019}
}

@article{wang2017memetic,
  title={Memetic algorithm based on sequential variable neighborhood descent for the minmax multiple traveling salesman problem},
  author={Wang, Yongzhen and Chen, Yan and Lin, Yan},
  journal={Computers \& Industrial Engineering},
  volume={106},
  pages={105--122},
  year={2017},
  publisher={Elsevier}
}

@article{palacios2019equitable,
  title={Equitable persistent coverage of non-convex environments with graph-based planning},
  author={Palacios-Gas{\'o}s, Jos{\'e} Manuel and Tardioli, Danilo and Montijano, Eduardo and Sag{\"u}{\'e}s, Carlos},
  journal={International Journal of Robotics Research},
  volume={38},
  number={14},
  pages={1674--1694},
  year={2019},
  publisher={SAGE Publications Sage UK: London, England}
}

@ARTICLE{tangyuan2020,
  author={Tang, Yuan and Zhou, Rui and Sun, Guibin and Di, Bin and Xiong, Rongling},
  journal={IEEE Sensors Journal}, 
  title={A Novel Cooperative Path Planning for Multirobot Persistent Coverage in Complex Environments}, 
  year={2020},
  volume={20},
  number={8},
  pages={4485-4495}}

@inproceedings{collins2021scalable,
  title={Scalable coverage path planning of multi-robot teams for monitoring non-convex areas},
  author={Collins, Leighton and Ghassemi, Payam and Esfahani, Ehsan T and Doermann, David and Dantu, Karthik and Chowdhury, Souma},
  booktitle={IEEE International Conference on Robotics and Automation},
  pages={7393--7399},
  year={2021}
}

@article{agarwal2022area,
  title={Area coverage with multiple capacity-constrained robots},
  author={Agarwal, Saurav and Akella, Srinivas},
  journal={IEEE Robotics and Automation Letters},
  volume={7},
  number={2},
  pages={3734--3741},
  year={2022},
  publisher={IEEE}
}

@inproceedings{agarwal2020line,
  title={Line coverage with multiple robots},
  author={Agarwal, Saurav and Akella, Srinivas},
  booktitle={IEEE International Conference on Robotics and Automation},
  pages={3248--3254},
  year={2020}
}

@article{datsko2024energy,
  title={Energy-aware multi-uav coverage mission planning with optimal speed of flight},
  author={Datsko, Denys and Nekovar, Frantisek and Penicka, Robert and Saska, Martin},
  journal={IEEE Robotics and Automation Letters},
  volume={9},
  number={3},
  pages={2893--2900},
  year={2024},
  publisher={IEEE}
}

@ARTICLE{cao2026multi,
  author={Cao, Mengfan and Yang, Zaiyue and Miao, Haoyu},
  journal={IEEE Robotics and Automation Letters}, 
  title={Multi-\uppercase{UAV} Coverage Path Planning Based on Balanced Graph Partitioning}, 
  year={2026},
  volume={11},
  number={4},
  pages={4409-4416}}

@article{huang2026comaea,
  title={\uppercase{C}o\uppercase{MAEA}: A Collision-Avoiding Multi-Agent Evolutionary Algorithm for Coverage Path Planning},
  author={Huang, Ting and Ma, Xiu-Wen and Liu, Xiao-Tao and Gong, Yue-Jiao},
  journal={IEEE Transactions on Evolutionary Computation},
  volume={},
  pages={},
  year={2026}
}

@article{xie2022multiregional,
  title={Multiregional coverage path planning for multiple energy constrained \uppercase{UAV}s},
  author={Xie, Junfei and Chen, Jun},
  journal={IEEE Transactions on Intelligent Transportation Systems},
  volume={23},
  number={10},
  pages={17366--17381},
  year={2022},
  publisher={IEEE}
}

@article{zhang2025multi,
  title={Multi-region joint coverage for environmental monitoring using energy-constrained \uppercase{UAV}s},
  author={Zhang, Changming and Xu, Caiyue and Cheng, Xu and Li, Xin and Li, Gang and He, Bin},
  journal={IEEE Transactions on Instrumentation and Measurement},
  volume={74},
  number={},
  pages={1-13},
  year={2025},
  publisher={IEEE}
}

@article{luo2026real,
  title={Real-Time Path-Reconfigurable Coverage Planning for Multi-\uppercase{UAV} Missions Over Disjoint Areas},
  author={Luo, Cai and Guo, Jintao and Liu, Zixuan and Liu, Lei and Luo, Chunbo},
  journal={IEEE Robotics and Automation Letters},
  volume={11},
  number={2},
  pages={1226--1233},
  year={2026},
  publisher={IEEE}
}

@inproceedings{zhao2023energy,
  title={Energy constrained multi-agent reinforcement learning for coverage path planning},
  author={Zhao, Chenyang and Liu, Juan and Yoon, Suk-Un and Li, Xinde and Li, Heqing and Zhang, Zhentong},
  booktitle={IEEE/RSJ International Conference on Intelligent Robots and Systems},
  pages={5590--5597},
  year={2023}
}

@inproceedings{shen2017,
  title={Autonomous 3-\uppercase{D} mapping and safe-path planning for underwater terrain reconstruction using multi-level coverage trees},
  author={Shen, Zongyuan and Song, Junnan and Mittal, Khushboo and Gupta, Shalabh},
  year = "2017",
  pages = "1--6",
  booktitle = "Proceedings OCEANS'17 MTS/IEEE"
}

@inproceedings{shen2016autonomous,
  title={An autonomous integrated system for 3-\uppercase{D} underwater terrain map reconstruction},
  author={Shen, Zongyuan and Song, Junnan and Mittal, Khushboo and Gupta, Shalabh},
  booktitle={Proceedings OCEANS'16 MTS/IEEE},
  pages={1--6},
  year={2016}
}

@article{shen2022ct,
  title={\uppercase{CT-CPP}: Coverage Path Planning for 3\uppercase{D} Terrain Reconstruction Using Dynamic Coverage Trees},
  author={Shen, Zongyuan and Song, Junnan and Mittal, Khushboo and Gupta, Shalabh},
  journal={IEEE Robotics and Automation Letters},
  volume={7},
  number={1},
  pages={135--142},
  year={2022},
  publisher={IEEE}
}

@article{galceran2015coverage,
  title={Coverage path planning with real-time replanning and surface reconstruction for inspection of three-dimensional underwater structures using autonomous underwater vehicles},
  author={Galceran, Enric and Campos, Ricard and Palomeras, Narc{\'\i}s and Ribas, David and Carreras, Marc and Ridao, Pere},
  journal={Journal of Field Robotics},
  volume={32},
  number={7},
  pages={952--983},
  year={2015},
  publisher={Wiley Online Library}
}

@INPROCEEDINGS{lin2017,
  author={Lin, Yu-Yao and Ni, Chien-Chun and Lei, Na and David Gu, Xianfeng and Gao, Jie},
  booktitle={IEEE International Conference on Robotics and Automation}, 
  title={Robot Coverage Path planning for general surfaces using quadratic differentials}, 
  year={2017},
  volume={},
  number={},
  pages={5005-5011}}

@article{wu2019energy,
  title={Energy-efficient coverage path planning for general terrain surfaces},
  author={Wu, Chenming and Dai, Chengkai and Gong, Xiaoxi and Liu, Yong-Jin and Wang, Jun and Gu, Xianfeng David and Wang, Charlie CL},
  journal={IEEE Robotics and Automation Letters},
  volume={4},
  number={3},
  pages={2584--2591},
  year={2019},
  publisher={IEEE}
}

@article{yordanova2020coverage,
  title={Coverage path planning with track spacing adaptation for autonomous underwater vehicles},
  author={Yordanova, Veronika and Gips, Bart},
  journal={IEEE Robotics and Automation Letters},
  volume={5},
  number={3},
  pages={4774--4780},
  year={2020},
  publisher={IEEE}
}

@ARTICLE{shao2025energy,
  author={Shao, Qi and Mao, Xuefei and Xu, Wenbin},
  journal={IEEE Robotics and Automation Letters}, 
  title={Energy-Aware \uppercase{UAV} Coverage Planning in Mountainous Terrain via Contour-Aligned Path Generation}, 
  year={2025},
  volume={10},
  number={12},
  pages={12373-12380}}

@article{do2023geometry,
  title={Geometry-aware coverage path planning for depowdering on complex 3\uppercase{D} surfaces},
  author={Do, Van-Thach and Pham, Quang-Cuong},
  journal={IEEE Robotics and Automation Letters},
  volume={8},
  number={9},
  pages={5552--5559},
  year={2023},
  publisher={IEEE}
}

@article{nguyen2009constrained,
  title={Constrained \uppercase{CVT} meshes and a comparison of triangular mesh generators},
  author={Nguyen, Hoa and Burkardt, John and Gunzburger, Max and Ju, Lili and Saka, Yuki},
  journal={Computational Geometry},
  volume={42},
  number={1},
  pages={1--19},
  year={2009},
  publisher={Elsevier}
}

@article{yang2020cellular,
  title={Cellular decomposition for nonrepetitive coverage task with minimum discontinuities},
  author={Yang, Tong and Miro, Jaime Valls and Lai, Qianen and Wang, Yue and Xiong, Rong},
  journal={IEEE/ASME Transactions on Mechatronics},
  volume={25},
  number={4},
  pages={1698--1708},
  year={2020},
  publisher={IEEE}
}

@inproceedings{yang2021optimal,
  title={Optimal object placement for minimum discontinuity non-revisiting coverage task},
  author={Yang, Tong and Miro, Jaime Valls and Wang, Yue and Xiong, Rong},
  booktitle={IEEE International Conference on Robotics and Automation},
  pages={8422--8428},
  year={2021}
}

@article{yang2023template,
  title={Template-free nonrevisiting uniform coverage path planning on curved surfaces},
  author={Yang, Tong and Miro, Jaime Valls and Nguyen, Minh and Wang, Yue and Xiong, Rong},
  journal={IEEE/ASME Transactions on Mechatronics},
  volume={28},
  number={4},
  pages={1853--1861},
  year={2023},
  publisher={IEEE}
}

@inproceedings{yang2024optimal,
  title={Optimal Non-Redundant Manipulator Surface Coverage with Rank-Deficient Manipulability Constraints.},
  author={Yang, Tong and Huang, Li and Mir{\'o}, Jaime Valls and Wang, Yue and Xiong, Rong},
  booktitle={Robotics: Science and Systems},
  year={2024}
}

@article{yang2024improved,
  title={An improved maximal continuity graph solver for non-redundant manipulator non-revisiting coverage},
  author={Yang, Tong and Miro, Jaime Valls and Wang, Yue and Xiong, Rong},
  journal={IEEE Transactions on Automation Science and Engineering},
  volume={22},
  pages={3822--3834},
  year={2024},
  publisher={IEEE}
}

@inproceedings{wang2025hierarchically,
  title={Hierarchically Accelerated Coverage Path Planning for Redundant Manipulators},
  author={Wang, Yeping and Gleicher, Michael},
  booktitle={IEEE International Conference on Robotics and Automation},
  pages={12098--12104},
  year={2025}
}

@article{bircher2016three,
  title={Three-dimensional coverage path planning via viewpoint resampling and tour optimization for aerial robots},
  author={Bircher, Andreas and Kamel, Mina and Alexis, Kostas and Burri, Michael and Oettershagen, Philipp and Omari, Sammy and Mantel, Thomas and Siegwart, Roland},
  journal={Autonomous Robots},
  volume={40},
  number={6},
  pages={1059--1078},
  year={2016},
  publisher={Springer}
}

@inproceedings{jing2016sampling,
  title={Sampling-based view planning for 3\uppercase{d} visual coverage task with unmanned aerial vehicle},
  author={Jing, Wei and Polden, Joseph and Lin, Wei and Shimada, Kenji},
  booktitle={IEEE/RSJ International Conference on Intelligent Robots and Systems},
  pages={1808--1815},
  year={2016}
}

@inproceedings{jing2017sampling,
  title={Sampling-based coverage motion planning for industrial inspection application with redundant robotic system},
  author={Jing, Wei and Polden, Joseph and Goh, Chun Fan and Rajaraman, Mabaran and Lin, Wei and Shimada, Kenji},
  booktitle={IEEE/RSJ International Conference on Intelligent Robots and Systems},
  pages={5211--5218},
  year={2017}
}

@inproceedings{jing2019coverage,
  title={Coverage path planning using path primitive sampling and primitive coverage graph for visual inspection},
  author={Jing, Wei and Deng, Di and Xiao, Zhe and Liu, Yong and Shimada, Kenji},
  booktitle={IEEE/RSJ International Conference on Intelligent Robots and Systems},
  pages={1472--1479},
  year={2019}
}

@inproceedings{jing2020multi,
  title={Multi-\uppercase{UAV} coverage path planning for the inspection of large and complex structures},
  author={Jing, Wei and Deng, Di and Wu, Yan and Shimada, Kenji},
  booktitle={IEEE/RSJ International Conference on Intelligent Robots and Systems},
  pages={1480--1486},
  year={2020}
}

@inproceedings{almadhoun2018coverage,
  title={Coverage path planning with adaptive viewpoint sampling to construct 3\uppercase{D} models of complex structures for the purpose of inspection},
  author={Almadhoun, Randa and Taha, Tarek and Gan, Dongming and Dias, Jorge and Zweiri, Yahya and Seneviratne, Lakmal},
  booktitle={IEEE/RSJ International Conference on Intelligent Robots and Systems},
  pages={7047--7054},
  year={2018}
}

@article{lindner2019optimization,
  title={Optimization based multi-view coverage path planning for autonomous structure from motion recordings},
  author={Lindner, Silvan and Garbe, Christoph and Mombaur, Katja},
  journal={IEEE Robotics and Automation Letters},
  volume={4},
  number={4},
  pages={3278--3285},
  year={2019},
  publisher={IEEE}
}

@inproceedings{cao2020hierarchical,
  title={Hierarchical coverage path planning in complex 3\uppercase{D} environments},
  author={Cao, Chao and Zhang, Ji and Travers, Matt and Choset, Howie},
  booktitle={IEEE International Conference on Robotics and Automation},
  pages={3206--3212},
  year={2020}
}

@article{song2020online,
  title={Online coverage and inspection planning for 3\uppercase{D} modeling},
  author={Song, Soohwan and Kim, Daekyum and Jo, Sungho},
  journal={Autonomous Robots},
  volume={44},
  number={8},
  pages={1431--1450},
  year={2020},
  publisher={Springer}
}

@article{liu2022coverage,
  title={Coverage path planning for robotic quality inspection with control on measurement uncertainty},
  author={Liu, Yinhua and Zhao, Wenzheng and Liu, Hongpeng and Wang, Yinan and Yue, Xiaowei},
  journal={IEEE/ASME Transactions on Mechatronics},
  volume={27},
  number={5},
  pages={3482--3493},
  year={2022},
  publisher={IEEE}
}

@inproceedings{feng2024fc,
  title={\uppercase{Fc-p}lanner: A skeleton-guided planning framework for fast aerial coverage of complex 3\uppercase{d} scenes},
  author={Feng, Chen and Li, Haojia and Zhang, Mingjie and Chen, Xinyi and Zhou, Boyu and Shen, Shaojie},
  booktitle={IEEE International Conference on Robotics and Automation},
  pages={8686--8692},
  year={2024}
}

@article{zhang2026novel,
  title={A Novel View Planning With Joint Optimization for Efficient 3\uppercase{D} Building Inspection},
  author={Zhang, Tenglong and Liu, Guoliang and Tian, Guohui},
  journal={IEEE Robotics and Automation Letters},
  volume={11},
  number={2},
  pages={1162--1169},
  year={2026},
  publisher={IEEE}
}

@article{gupta2015multifactorial,
  title={Multifactorial evolution: Toward evolutionary multitasking},
  author={Gupta, Abhishek and Ong, Yew-Soon and Feng, Liang},
  journal={IEEE Transactions on Evolutionary Computation},
  volume={20},
  number={3},
  pages={343--357},
  year={2015},
  publisher={IEEE}
}

@article{zheng2024new,
  title={A new clustering-based view planning method for building inspection with drone},
  author={Zheng, Yongshuai and Liu, Guoliang and Ding, Yan and Tian, Guohui},
  journal={IEEE Robotics and Automation Letters},
  volume={9},
  number={11},
  pages={9781--9788},
  year={2024},
  publisher={IEEE}
}

@article{cabreira2019survey,
  title={Survey on coverage path planning with unmanned aerial vehicles},
  author={Cabreira, Tau{\~a} M and Brisolara, Lisane B and Paulo R, Ferreira Jr},
  journal={Drones},
  volume={3},
  number={1},
  pages={4},
  year={2019},
  publisher={MDPI}
}

@article{vempati2018paintcopter,
  title={Paint\uppercase{c}opter: An autonomous \uppercase{uav} for spray painting on three-dimensional surfaces},
  author={Vempati, Anurag Sai and Kamel, Mina and Stilinovic, Nikola and Zhang, Qixuan and Reusser, Dorothea and Sa, Inkyu and Nieto, Juan and Siegwart, Roland and Beardsley, Paul},
  journal={IEEE Robotics and Automation Letters},
  volume={3},
  number={4},
  pages={2862--2869},
  year={2018},
  publisher={IEEE}
}

@article{englot2013three,
  title={Three-dimensional coverage planning for an underwater inspection robot},
  author={Englot, Brendan and Hover, Franz S},
  journal={International Journal of Robotics Research},
  volume={32},
  number={9-10},
  pages={1048--1073},
  year={2013},
  publisher={SAGE Publications Sage UK: London, England}
}

@article{atkar2005,
  title={Uniform coverage of automotive surface patches},
  author={Atkar, Prasad N and Greenfield, Aaron and Conner, David C and Choset, Howie and Rizzi, Alfred A},
  journal={International Journal of Robotics Research},
  volume={24},
  number={11},
  pages={883--898},
  year={2005},
  publisher={SAGE Publications}
}

@article{morilla2022sweep,
  title={Sweep-\uppercase{y}our-\uppercase{m}ap: Efficient coverage planning for aerial teams in large-scale environments},
  author={Morilla-Cabello, David and Bartolomei, Luca and Teixeira, Lucas and Montijano, Eduardo and Chli, Margarita},
  journal={IEEE Robotics and Automation Letters},
  volume={7},
  number={4},
  pages={10810--10817},
  year={2022},
  publisher={IEEE}
}

@article{cao2025complete,
  title={Complete coverage search for multiple autonomous underwater vehicles based on neuronal activity reassignment},
  author={Cao, Zhe and Fan, Huili and Hu, Xinyu and Chen, Yanli and Kang, Shuai},
  journal={IEEE Transactions on Intelligent Transportation Systems},
  year={2025},
  volume={26},
  number={7},
  pages={9693-9710},
  publisher={IEEE}
}

@inproceedings{sadat2014recursive,
  title={Recursive non-uniform coverage of unknown terrains for \uppercase{uav}s},
  author={Sadat, Seyed Abbas and Wawerla, Jens and Vaughan, Richard T},
  booktitle={IEEE/RSJ international conference on intelligent robots and systems},
  pages={1742--1747},
  year={2014}
}

@inproceedings{sadat2015fractal,
  title={Fractal trajectories for online non-uniform aerial coverage},
  author={Sadat, Seyed Abbas and Wawerla, Jens and Vaughan, Richard},
  booktitle={IEEE International Conference on Robotics and Automation},
  pages={2971--2976},
  year={2015}
}

@article{jensen2020near,
  title={Near-optimal area-coverage path planning of energy-constrained aerial robots with application in autonomous environmental monitoring},
  author={Jensen-Nau, Katharin R and Hermans, Tucker and Leang, Kam K},
  journal={IEEE Transactions on Automation Science and Engineering},
  volume={18},
  number={3},
  pages={1453--1468},
  year={2020},
  publisher={IEEE}
}

@inproceedings{chen2019adaptive,
  title={Adaptive deep path: Efficient coverage of a known environment under various configurations},
  author={Chen, Xin and Tucker, Thomas M and Kurfess, Thomas R and Vuduc, Richard},
  booktitle={IEEE/RSJ International Conference on Intelligent Robots and Systems},
  pages={3549--3556},
  year={2019}
}

@article{fu2024full,
  title={Full coverage path planning recombination framework for unmanned vehicles with multi-objective constraints},
  author={Fu, Jinyu and Yao, Weiran and Sun, Guanghui and Liu, Jianxing and Wu, Ligang},
  journal={IEEE Transactions on Industrial Electronics},
  volume={71},
  number={8},
  pages={9276--9286},
  year={2024},
  publisher={IEEE}
}

@article{fu2026online,
  title={Online Exploratory Coverage Path Planning of Incremental SLAM for Autonomous Vehicles},
  author={Fu, Jinyu and Zhu, Hong and Zhang, Kai and Ma, Teng and Liu, Jingcheng and Li, Ye},
  journal={IEEE Transactions on Industrial Informatics},
  year={2026},
  volume={22},
  number={3},
  pages={1861-1870},
  publisher={IEEE}
}

@article{tang2022learning,
  title={Learning to coordinate for a worker-station multi-robot system in planar coverage tasks},
  author={Tang, Jingtao and Gao, Yuan and Lam, Tin Lun},
  journal={IEEE Robotics and Automation Letters},
  volume={7},
  number={4},
  pages={12315--12322},
  year={2022},
  publisher={IEEE}
}

@article{galceran2013survey,
  title={A survey on coverage path planning for robotics},
  author={Galceran, Enric and Carreras, Marc},
  journal={Robotics and Autonomous systems},
  volume={61},
  number={12},
  pages={1258--1276},
  year={2013},
  publisher={Elsevier}
}

@inproceedings{bormann2018indoor,
  title={Indoor coverage path planning: Survey, implementation, analysis},
  author={Bormann, Richard and Jordan, Florian and Hampp, Joshua and H{\"a}gele, Martin},
  booktitle={IEEE International Conference on Robotics and Automation},
  pages={1718--1725},
  year={2018}
}

@article{hoffmann2024optimal,
  title={Optimal guidance track generation for precision agriculture: A review of coverage path planning techniques},
  author={H{\"o}ffmann, Maria and Patel, Shruti and B{\"u}skens, Christof},
  journal={Journal of Field Robotics},
  volume={41},
  number={3},
  pages={823--844},
  year={2024},
  publisher={Wiley Online Library}
}

@article{choset2001coverage,
  title={Coverage for robotics--a survey of recent results},
  author={Choset, Howie},
  journal={Annals of mathematics and artificial intelligence},
  volume={31},
  number={1},
  pages={113--126},
  year={2001},
  publisher={Springer}
}

@inproceedings{modares2017ub,
  title={\uppercase{UB-ANC} planner: Energy efficient coverage path planning with multiple drones},
  author={Modares, Jalil and Ghanei, Farshad and Mastronarde, Nicholas and Dantu, Karthik},
  booktitle={IEEE International Conference on Robotics and Automation},
  pages={6182--6189},
  year={2017}
}

@inproceedings{jung2018multi,
  title={Multi-layer coverage path planner for autonomous structural inspection of high-rise structures},
  author={Jung, Sungwook and Song, Seungwon and Youn, Pillip and Myung, Hyun},
  booktitle={IEEE/RSJ International Conference on Intelligent Robots and Systems},
  pages={1--9},
  year={2018}
}

@inproceedings{schirmer2019coverage,
  title={Coverage path planning in belief space},
  author={Schirmer, Robert and Biber, Peter and Stachniss, Cyrill},
  booktitle={IEEE International Conference on Robotics and Automation},
  pages={7604--7610},
  year={2019}
}

@inproceedings{yu2019complete,
  title={Complete and near-optimal path planning for simultaneous sensor-based inspection and footprint coverage in robotic crack filling},
  author={Yu, Kaiyan and Guo, Chaoke and Yi, Jingang},
  booktitle={IEEE International Conference on Robotics and Automation},
  pages={8812--8818},
  year={2019}
}

@article{Ou2025,
  author={Ou, Yaming and Fan, Junfeng and Zhou, Chao and Kang, Song and Zhang, Zhuoliang and Hou, Zeng-Guang and Tan, Min},
  journal={IEEE Transactions on Systems, Man, and Cybernetics}, 
  title={Structured Light-Based Underwater Collision-Free Navigation and Dense Mapping System for Refined Exploration in Unknown Dark Environments}, 
  year={2025},
  volume={55},
  number={1},
  pages={110-123}}

@article{veeraraghavan2024complete,
  title={Complete and near-optimal robotic crack coverage and filling in civil infrastructure},
  author={Veeraraghavan, Vishnu and Hunte, Kyle and Yi, Jingang and Yu, Kaiyan},
  journal={IEEE Transactions on Robotics},
  volume={40},
  pages={2850--2867},
  year={2024},
  publisher={IEEE}
}

@inproceedings{choton2023optimal,
  title={Optimal multi-robot coverage path planning for agricultural fields using motion dynamics},
  author={Choton, Jahid Chowdhury and Prabhakar, Pavithra},
  booktitle={IEEE International Conference on Robotics and Automation},
  pages={11817--11823},
  year={2023}
}

@inproceedings{karapetyan2024ag,
  title={\uppercase{Ag-cvg}: Coverage planning with a mobile recharging ugv and an energy-constrained \uppercase{uav}},
  author={Karapetyan, Nare and Asghar, Ahmad Bilal and Bhaskar, Amisha and Shi, Guangyao and Manocha, Dinesh and Tokekar, Pratap},
  booktitle={IEEE International Conference on Robotics and Automation},
  pages={2617--2623},
  year={2024}
}

@inproceedings{mitra2022scalable,
  title={Scalable online coverage path planning for multi-robot systems},
  author={Mitra, Ratijit and Saha, Indranil},
  booktitle={IEEE/RSJ International Conference on Intelligent Robots and Systems},
  pages={10102--10109},
  year={2022}
}

@inproceedings{mitra2024online,
  title={Online on-demand multi-robot coverage path planning},
  author={Mitra, Ratijit and Saha, Indranil},
  booktitle={IEEE International Conference on Robotics and Automation},
  pages={14583--14589},
  year={2024}
}

@inproceedings{mitra2025online,
  title={Online concurrent multi-robot coverage path planning},
  author={Mitra, Ratijit and Saha, Indranil},
  booktitle={IEEE/RSJ International Conference on Intelligent Robots and Systems},
  pages={8691--8698},
  year={2025}
}

@article{shen2025multi,
  title={Multi-\uppercase{CAP}: A Multi-Robot Connectivity-Aware Hierarchical Coverage Path Planning Algorithm for Unknown Environments},
  author={Shen, Zongyuan and Shirose, Burhanuddin and Sriganesh, Prasanna and Vundurthy, Bhaskar and Choset, Howie and Travers, Matthew},
  journal={arXiv preprint arXiv:2509.14941},
  year={2025}
}

@article{wang2026dmt,
  title={\uppercase{DMT-CPP}: A Delaunay-Graph-Based Framework for Real-Time Multi-Robot Online Target Coverage},
  author={Wang, Pengyu and Wang, Zikai and Shi, Ling and Meng, Max Q-H},
  journal={IEEE Robotics and Automation Letters},
  year={2026},
  publisher={IEEE}
}

@article{tu2023configuration,
  title={Configuration identification for a freeform modular self-reconfigurable robot-\uppercase{f}ree\uppercase{sn}},
  author={Tu, Yuxiao and Lam, Tin Lun},
  journal={IEEE Transactions on Robotics},
  volume={39},
  number={6},
  pages={4636--4652},
  year={2023},
  publisher={IEEE}
}

@article{wang2024haidou,
  title={The Haidou-1 hybrid underwater vehicle for the Mariana Trench science exploration to 10,908 m depth},
  author={Wang, Jian and Tang, Yuangui and Li, Shuo and Lu, Yang and Li, Jixu and Liu, Tiejun and Jiang, Zhibin and Chen, Cong and Cheng, Yu and Yu, Deyong and others},
  journal={Journal of Field Robotics},
  volume={41},
  number={4},
  pages={1054--1079},
  year={2024},
  publisher={Wiley Online Library}
}
